\documentclass[]{fairmeta}

\ifPDFTeX
  \usepackage[utf8]{inputenc}
\fi
\usepackage{latexsym}
\usepackage{textcomp}
\usepackage{tabularx}
\usepackage{array}
\usepackage{longtable}
\usepackage{calc}
\usepackage{amsmath}
\usepackage{amssymb}
\usepackage{colortbl}
\usepackage{float}

\ifPDFTeX
  \makeatletter
  \@ifl@t@r\fmtversion{2020/01/01}{}{}
  \makeatother
\fi

\ifPDFTeX
\DeclareUnicodeCharacter{00B0}{\ensuremath{^\circ}}
\DeclareUnicodeCharacter{00B1}{\ensuremath{\pm}}
\DeclareUnicodeCharacter{00B2}{\textsuperscript{2}}
\DeclareUnicodeCharacter{00B3}{\textsuperscript{3}}
\DeclareUnicodeCharacter{00B7}{\ensuremath{\cdot}}
\DeclareUnicodeCharacter{00B9}{\textsuperscript{1}}
\DeclareUnicodeCharacter{00D7}{\ensuremath{\times}}
\DeclareUnicodeCharacter{0394}{\ensuremath{\Delta}}
\DeclareUnicodeCharacter{03B1}{\ensuremath{\alpha}}
\DeclareUnicodeCharacter{03B2}{\ensuremath{\beta}}
\DeclareUnicodeCharacter{03B8}{\ensuremath{\theta}}
\DeclareUnicodeCharacter{03BB}{\ensuremath{\lambda}}
\DeclareUnicodeCharacter{03C0}{\ensuremath{\pi}}
\DeclareUnicodeCharacter{03C9}{\ensuremath{\omega}}
\DeclareUnicodeCharacter{2076}{\textsuperscript{6}}
\DeclareUnicodeCharacter{2079}{\textsuperscript{9}}
\DeclareUnicodeCharacter{207B}{\textsuperscript{-}}
\DeclareUnicodeCharacter{211D}{\ensuremath{\mathbb{R}}}
\DeclareUnicodeCharacter{2191}{\ensuremath{\uparrow}}
\DeclareUnicodeCharacter{2192}{\ensuremath{\rightarrow}}
\DeclareUnicodeCharacter{2193}{\ensuremath{\downarrow}}
\DeclareUnicodeCharacter{2208}{\ensuremath{\in}}
\DeclareUnicodeCharacter{2212}{\ensuremath{-}}
\DeclareUnicodeCharacter{2216}{\ensuremath{\setminus}}
\DeclareUnicodeCharacter{221A}{\ensuremath{\sqrt{}}}
\DeclareUnicodeCharacter{2248}{\ensuremath{\approx}}
\DeclareUnicodeCharacter{2261}{\ensuremath{\equiv}}
\DeclareUnicodeCharacter{2264}{\ensuremath{\leq}}
\DeclareUnicodeCharacter{2265}{\ensuremath{\geq}}
\DeclareUnicodeCharacter{22EF}{\ensuremath{\cdots}}
\fi

\newcolumntype{Y}{>{\raggedright\arraybackslash}X}
\newcolumntype{C}{>{\centering\arraybackslash}X}

\definecolor{takeawaybg}{HTML}{F2F2F2}
\definecolor{bestrow}{HTML}{FFF3B8}
\definecolor{baserow}{HTML}{E6E6E6}
\definecolor{pairedrow}{HTML}{DCEBFA}
\newtcolorbox{takeawaypanel}{
  colback=takeawaybg,
  colframe=takeawaybg,
  boxrule=0pt,
  arc=2mm,
  outer arc=2mm,
  left=3pt,
  right=3pt,
  top=2pt,
  bottom=2pt,
  before skip=3pt,
  after skip=3pt
}
\newtcolorbox{promptbox}{
  colback=white,
  colframe=black!45,
  boxrule=0.4pt,
  arc=1mm,
  left=4pt,
  right=4pt,
  top=3pt,
  bottom=3pt,
  before skip=4pt,
  after skip=6pt,
  fontupper=\small\ttfamily
}
\newcommand{\takeawaybox}[2]{%
  \begin{takeawaypanel}
   \textbf{#1 Takeaway.} #2%
  \end{takeawaypanel}%
}
\title{Privileged Information Does Not Guarantee Privileged Guidance: Auditing On-Policy Self-Distillation for Reasoning Models}
\author[1,2,*]{Haoyi Qiu}
\author[1]{Akshat Agarwal}
\affiliation[1]{Meta Superintelligence Labs}
\affiliation[2]{University of California, Los Angeles}
\contribution[*]{Work done during internship at Meta}
\correspondence{Haoyi Qiu (\email{haoyiqiu@cs.ucla.edu})}
\date{October 15, 2026}
\abstract{
On-policy self-distillation (OPSD) trains a student on next-token targets from a teacher, a frozen copy of the student that additionally receives privileged information, such as a worked solution, hidden from the student. Whether student gains reflect that information remains unclear. Taking the worked solution as privileged information, we audit four stages across three Qwen3 scales on mathematics: \textbf{Source}, whether the teacher can use the solution; \textbf{Encoding}, whether the resulting token targets are useful; \textbf{Transmission}, whether the student learns from the targets; and \textbf{Expression}, whether the student applies the learned behavior without the solution. With the solution, the teacher solves hard problems and recovers from student errors, yet the teacher does not reliably recognize errors, and its token targets do not favor correct trajectories. Repeated training makes successes within the student's reach more frequent rather than unlocking new ones, and corrections the student learns to predict are not reliably expressed in free generation. To isolate the solution's contribution, we vary its availability separately from teacher and student thinking modes. The largest gains occur when a teacher in thinking mode supervises a student trained and evaluated without an explicit thinking trace, and problem-only teachers often match or exceed privileged-solution teachers: the gain tracks the teacher's thinking mode rather than the solution, a pattern we call reasoning arbitrage. Accuracy gains alone therefore do not establish transfer of reasoning from privileged solutions: bottlenecks lie after Source, across Encoding, Transmission, and Expression. Future evaluations must measure what the solution adds to otherwise identical training.
}

\begin{document}
\maketitle

\section{Introduction}
\label{sec:introduction}

Privileged supervision offers a route to self-improving reasoning models. On-policy self-distillation (OPSD) extends on-policy distillation \citep{agarwal2024onpolicy}: after the student generates a response, a frozen copy of the same model conditions on privileged information, here a worked solution (the privileged solution), and scores the rollout at each prefix, providing dense next-token supervision \citep{zhao2026selfdistilled,shenfeld2026selfdistillation,hubotter2026reinforcement}. Unlike outcome-based reinforcement learning with verifiable rewards \citep{shao2024deepseekmath,guo2025deepseekr1}, this objective supplies a training signal at every response position. OPSD asks one privileged model to act as \emph{solver}, \emph{evaluator}, and \emph{teacher}. But solving with the solution does not imply recognizing student errors, nor does recognition imply producing useful targets.

\begin{figure*}[t]
  \centering
  \includegraphics[width=\textwidth]{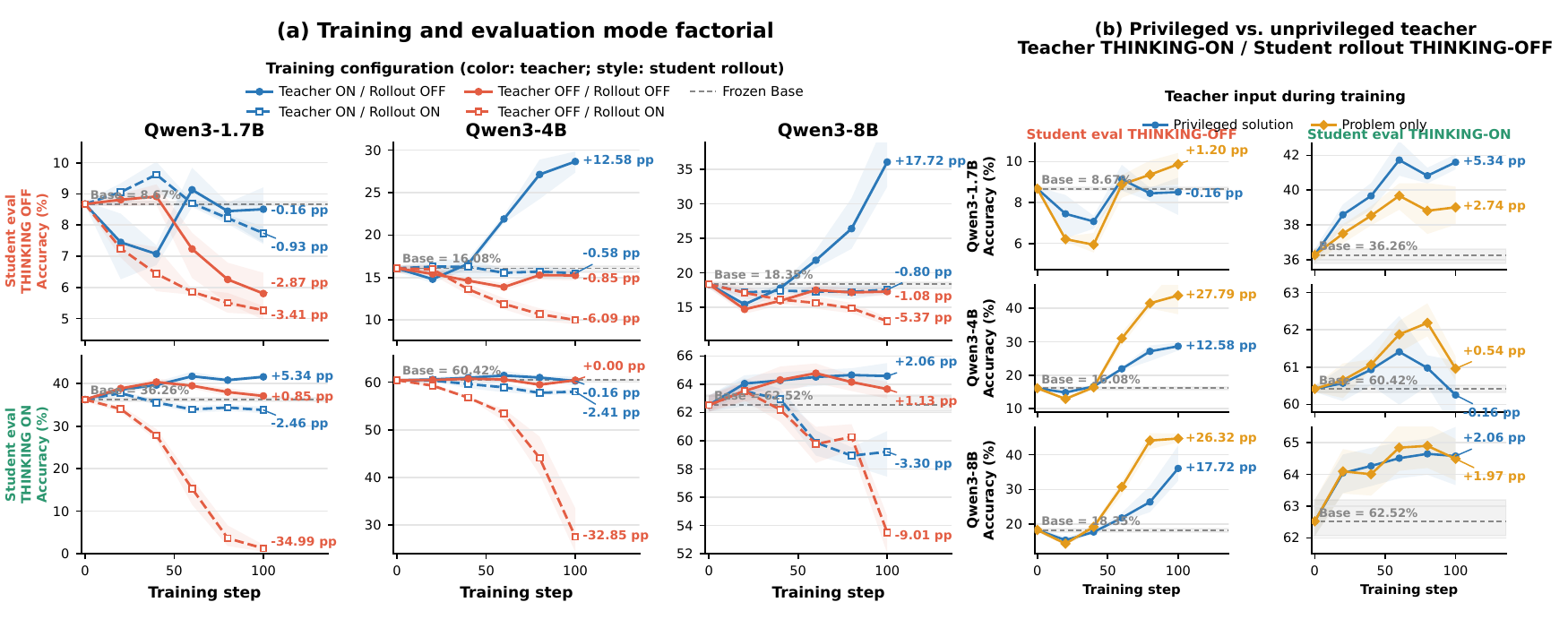}
  \vspace{-8mm}
  \caption{\textbf{Thinking-ON teachers improve Thinking-OFF generation at 4B and 8B with or without the privileged solution.} (a) Avg@12 by teacher and rollout modes, averaged over AIME 2024, AIME 2025, AIME 2026, HMMT Nov 2025, and HMMT Feb 2026. Rows give evaluation modes. (b) Privileged-solution versus problem-only teachers with a Thinking-ON teacher and Thinking-OFF rollouts. All teachers in (a) receive the solution, and (b) varies only teacher access to it. Base is the starting checkpoint in each evaluation mode. Trained curves average three training seeds, while Base averages three decoding seeds. Labels show Step-100 changes from Base (pp). Shading shows descriptive 95\% bootstrap intervals.}
  \label{fig:interface-factorial}
\end{figure*}

The privileged solution is not the teacher's only advantage over the student. The teacher's thinking mode can also affect student gains, so attributing those gains to the solution requires a control. Qwen3's chat template enables or disables a dedicated thinking block (Thinking ON or OFF) \citep{qwen2025qwen3}, and we set this mode independently for teacher scoring, student rollouts, and evaluation. Across Qwen3-1.7B, 4B, and 8B and five competition-math benchmarks (AIME 2024, 2025, 2026 and HMMT Nov 2025 and Feb 2026), Figure~\ref{fig:interface-factorial} compares these choices. When teacher scoring, student rollouts, and evaluation use the same mode, Standard OPSD \citep{zhao2026selfdistilled} gives no sustained gains over Base, the starting checkpoint. The largest sustained gains at 4B and 8B occur with a Thinking-ON teacher, Thinking-OFF rollouts, and Thinking-OFF evaluation. At Step 100 under Thinking-OFF evaluation, students trained with a problem-only teacher have higher mean Avg@12 than those trained with a privileged-solution teacher at all three scales (+1.36pp, +15.21pp, and +8.60pp at 1.7B, 4B, and 8B). The 1.7B interval includes zero. Under Thinking-ON evaluation, differences are small and mixed. We call this pattern \textbf{reasoning arbitrage}: using a Thinking-ON teacher to improve Thinking-OFF generation from the same model without a supplied solution.

Concurrent work questions both the source of OPSD gains and whether privileged teacher targets improve student accuracy. OPSD gains can survive replacing or removing the privileged solution \citep{ichihara2026privileged,shrestha2026rethinking,zhang2026privileged}. Teacher targets can be pulled toward the supplied solution rather than toward correctness \citep{harne2026privileged}. Token-level scores need not track task success \citep{nguyen2026privileged}. Thinking models change behavior as privileged context is withheld \citep{kaur2026rethinking}. Together, these studies leave open whether a teacher that solves with the solution can teach the student to solve without it. We test the steps required for transfer from the solution: the teacher must use it (\textbf{Source}), the resulting token targets must be useful (\textbf{Encoding}), the student must learn from those targets (\textbf{Transmission}), and the learned behavior must appear without the solution (\textbf{Expression}). We examine these steps across Qwen3 models using matched comparisons tailored to each stage.

The Standard OPSD objective shows where the solution enters teacher supervision while the student remains solution-free. For a minibatch $\{(x_i,y_i^\star)\}_{i\in\mathcal B}$, the student $p_S$ (parameters $\theta$) samples $\hat y_i\sim p_S(\cdot\mid x_i)$. A frozen initial copy $p_T$ alone sees $y_i^\star$ and supplies next-token targets at the same prefixes. With $D$ denoting clipped forward KL, the implemented loss averages over all valid response positions in the minibatch:

\begingroup\small
\vspace{-4mm}
\begin{equation*}
\begin{split}
\mathcal{L}_{\mathrm{OPSD}}(\theta;\mathcal B)
={}& \frac{1}{\sum_{i\in\mathcal B}|\hat y_i|}
\sum_{i\in\mathcal B}\sum_{n=1}^{|\hat y_i|}
D\big( p_T(\cdot\mid x_i,y_i^\star,\hat y_{i,<n})
\,\big\|\,
p_S(\cdot\mid x_i,\hat y_{i,<n})
\big).
\end{split}
\end{equation*}
\endgroup

We organize the audit around \textit{four} research questions. \textbf{RQ1: Source} (\S\ref{sec:source}) Can the frozen teacher use the solution to solve hard problems, recover after student errors, and recognize those errors? \textbf{RQ2: Encoding} (\S\ref{sec:encoding}) Does adding the solution change token targets in ways that distinguish correct from incorrect trajectories and help select continuations with higher completion accuracy? \textbf{RQ3: Transmission} (\S\ref{sec:transmission}) Does repeated OPSD make correct answers more frequent, including on questions with no correct Base response in screening? Do gains require the solution, and do they persist during longer training? \textbf{RQ4: Expression} (\S\ref{sec:expression}) Does training on verified corrections improve error recovery and solving from the original question without the privileged solution?\looseness=-1

To answer these questions, we audit privileged supervision from teacher behavior to student generation. Across three Qwen3 scales on mathematics, we test teacher solving and error recognition, token targets, and student learning. We also test verified corrections at 4B and 8B and run matched retention checks in OLMo and on coding tasks. The results show that a teacher's ability to use a solution does not guarantee useful student supervision. The solution improves teacher solving and recovery and shifts token targets, but does not reliably improve error recognition, correctness ranking, or branch selection. Repeated training raises pass@8 on the same questions, but Thinking-OFF solving remains unreliable on audited questions with no correct Base response in screening. Large Thinking-OFF gains occur with problem-only teachers, and Purified OPSD \citep{shen2026purified} improves retention during longer training. In the correction pilot, some variants score slightly above Standard OPSD when generation starts from the original question, but none improves error-prefix recovery and higher correction doses can reduce broader accuracy. Together, these findings support the reasoning-arbitrage interpretation of the largest gains.

\section{Related Work}
\label{sec:related-work}

\noindent\textbf{Privileged information and OPSD.} Learning using privileged information \citep{vapnik2009new,lopezpaz2016unifying} studies training-only supervision, while on-policy distillation \citep{agarwal2024onpolicy} evaluates teacher targets on student-sampled states. OPSD combines them by conditioning a frozen self-teacher on a privileged solution \citep{zhao2026selfdistilled}. Related methods \citep{shenfeld2026selfdistillation,hubotter2026reinforcement,ye2026onpolicy} use demonstrations, feedback, or other training-only context. Variants alter targets or credit assignment \citep{shen2026purified,xu2026bopsd,peng2026diagnosing}. Concurrent work varies the teacher's context using problem-space guidance, pseudo-solutions, or abstractions of full solutions \citep{zhao2026problemspace,li2026unsupervised,tian2026selfteacher}, while Privileged Hidden Flow adds hidden-state transition matching to output supervision \citep{li2026hiddenflow}.

\vspace{1mm}
\noindent\textbf{Empirical analyses of OPSD.} At the target level, richer teacher context can suppress epistemic verbalization \citep{kim2026selfdistillation}, privileged targets can penalize reconsideration tokens \citep{kaur2026rethinking}, and teacher targets can be pulled toward the supplied solution \citep{harne2026privileged}. Privileged-likelihood scores need not track correctness \citep{nguyen2026privileged}. At the training level, gains can survive replacing or removing the privileged solution \citep{ichihara2026privileged,shrestha2026rethinking}. Matched reference-free distillation finds that the supplied reference explains only part of the improvement in some settings \citep{zhang2026privileged}. Full solutions need not outperform abstracted teacher context \citep{zhao2026problemspace,tian2026selfteacher}. Separate analyses find that OPSD can compress correct Thinking-ON rollouts more reliably than it corrects incorrect ones \citep{kim2026opsd} and that a student can learn an average of instance-specific privileged teachers that lacks needed information at test time \citep{zhu2026manyfaces}. Our audit combines a teacher, rollout, and evaluation thinking-mode factorial with matched problem-only controls. It then tests teacher error recognition, correctness ranking and sibling-branch selection, acquisition on questions with no correct Base response in screening, and unaided generation after training on verified corrections (Appendix~\ref{app:related}). These tests distinguish the solution's contribution from the teacher's thinking-mode advantage and examine whether corrections learned with privileged support appear in generation without it. Prior work on self-correction without external feedback provides context for the recognition test \citep{huang2024cannot,kamoi2024when}.

\section{Experimental Setup}
\label{sec:setup}

\noindent\textbf{Models and data.} Our primary audit uses Qwen3-1.7B, 4B, and 8B \citep{qwen2025qwen3}. The main training runs use the mathematical reasoning subset of OpenThoughts (OT) \citep{guha2025openthoughts}. We also use Hendrycks MATH Level 5 (HM) \citep{hendrycks2021measuring} to audit hard problems and repeat selected analyses on a second dataset. GPT-5.6-Sol checks whether each mathematical problem is well posed and whether its answer and worked solution are valid and consistent, without using model performance (Appendix~\ref{app:data}). Qwen3's chat template controls whether a dedicated thinking block is enabled (Thinking ON) or disabled (Thinking OFF). Disabling the block does not prevent visible step-by-step reasoning in the response. We set the mode independently for teacher scoring, student rollouts, and evaluation. Each teacher prompt uses the teacher's selected mode (Appendix~\ref{app:figure1-interfaces}). Retention analyses also train OLMo-3-7B-Think \citep{olmo2025olmo3} on Hendrycks MATH and Qwen3-4B on execution-filtered CodeContests \citep{li2022competition}.

\vspace{1mm}
\noindent\textbf{Training and controls.} Our Standard OPSD baseline reimplements the method of \citet{zhao2026selfdistilled} in veRL (Appendix~\ref{app:figure1-training}). Unless noted, the Qwen3 student and its frozen teacher start from the same Base checkpoint. Only the student's LoRA parameters are updated. Problem-only teachers receive the problem, while privileged teachers also receive the worked solution (Appendix~\ref{app:figure1-interfaces}). Teacher and student thinking modes may differ. Each primary run processes the same 3,200 OT examples once, with 100 updates at global batch size 32. For each problem, the student samples one response of up to 1,024 tokens. The frozen teacher and current student score the same sampled response prefixes under their respective inputs. The teacher-to-student forward-KL (FKL) loss caps each vocabulary term at 0.05 before summing over vocabulary and averages over valid response tokens across the global minibatch. We use this cap at every scale and in both thinking modes. Unless noted, training runs use three seeds matched across conditions.

\begin{table}[t]
  \vspace{-2mm}
  \centering
  \footnotesize
  \newcommand{\celldelta}[1]{%
    {\scriptsize\ifdim #1pt>0pt
      \textcolor{green!45!black}{(#1)}%
    \else\ifdim #1pt<0pt
      \textcolor{red!70!black}{(#1)}%
    \else
      (#1)%
    \fi\fi}%
  }
  \setlength{\tabcolsep}{1pt}
  \renewcommand{\arraystretch}{1.0}
  \begin{minipage}[t]{0.48\textwidth}
  \vspace{0pt}
  \centering
  \begin{tabularx}{\linewidth}{@{}c@{\hspace{2pt}}|@{\hspace{2pt}}*{3}{>{\centering\arraybackslash}X}@{}}
    \toprule
    \textbf{Model} & \textbf{Problem only} & \textbf{Shuffled} & \textbf{Privileged} \\
    \midrule
    1.7B & 2.93 & 1.13\,\celldelta{-1.80} & 40.54\,\celldelta{+37.61} \\
    4B & 1.14 & 1.70\,\celldelta{+0.57} & 64.20\,\celldelta{+63.07} \\
    8B & 0.00 & 1.43\,\celldelta{+1.43} & 58.57\,\celldelta{+58.57} \\
    \bottomrule
  \end{tabularx}
  \vspace{-2mm}
  \caption{\textbf{Privileged solutions improve hard-problem solving.} Avg@4 (\%) on HM problems with no correct screening rollout. Parentheses give changes from problem only (pp). Appendix Table~\hyperref[app:tab-d6]{D6} reports paired 95\% question-bootstrap intervals.}
  \label{tab:source-solving}
  \end{minipage}%
  \hfill
  \begin{minipage}[t]{0.48\textwidth}
  \vspace{0pt}
  \centering
\begin{tabularx}{\linewidth}{@{}c@{\hspace{2pt}}|@{\hspace{2pt}}*{3}{>{\centering\arraybackslash}X}@{}}
  \toprule
  \textbf{Model} & \textbf{Problem only} & \textbf{Shuffled} & \textbf{Privileged} \\
  \midrule
  1.7B & 48.50 & 41.50\,\celldelta{-7.00} & 74.50\,\celldelta{+26.00} \\
  4B & 66.00 & 51.00\,\celldelta{-15.00} & 88.50\,\celldelta{+22.50} \\
  8B & 63.00 & 53.50\,\celldelta{-9.50} & 88.50\,\celldelta{+25.50} \\
  \bottomrule
\end{tabularx}
\vspace{-2mm}
\caption{\textbf{Privileged solutions improve post-error recovery.} Avg@4 (\%) on 50 verified OT error prefixes per scale. Parentheses give changes from problem only (pp). Appendix Table~\hyperref[app:tab-d2]{D2} reports paired 95\% cluster-bootstrap intervals.}
\label{tab:source-recovery}

  \end{minipage}
\end{table}

\vspace{1mm}
\noindent\textbf{Evaluation.} On the five mathematics benchmarks, we evaluate without the privileged solution. We draw 12 responses per problem with a 38,912-token cap and score extracted final answers with an automatic mathematical-equivalence verifier. We report mean Avg@12 across AIME 2024--2026 and HMMT Nov 2025 and Feb 2026 \citep{mathai2024aime,matharena2025aime,matharena2026aime,matharena2025hmmt,matharena2026hmmt}. Coding evaluations also omit the privileged solution and use four execution-scored responses per problem on LiveCodeBench v6 \citep{jain2024livecodebench} and held-out CodeContests \citep{li2022competition}. Appendix~\ref{app:setup-eval} gives further details.

\section{RQ1: Source}
\label{sec:source}

RQ1 tests whether the teacher can use the privileged solution to solve hard problems, recover from student errors, and recognize those errors. We run the teacher with Thinking ON, the teacher mode used in our main Standard OPSD comparisons and in every configuration with sustained gains in Figure~\ref{fig:interface-factorial}. We keep the checkpoint fixed and vary only its context, isolating how privileged information affects solving, recovery, and error recognition in these audits.

\subsection{Solution improves hard-problem solving}
\label{sec:source-solving}

For each model scale, we select hard problems from the HM dataset with no correct response among 12 screening rollouts, as confirmed by GPT-5.6-Sol and Claude Opus 4.8 after an automatic answer-equivalence screen. From the 2,242 cleaned questions (\S\ref{sec:setup}), this yields 111 (5.0\%), 44 (2.0\%), and 35 (1.6\%) questions at 1.7B, 4B, and 8B. On each question, the teacher generates four fresh rollouts under each of three prompt configurations: problem only, a worked solution from another problem (shuffled reference), or the solution to the current problem (privileged solution). Because these are new samples, problem-only accuracy need not be zero. The two judges also assess these fresh responses at the task level, with disagreements resolved by a separate blinded pass. Appendix~\ref{app:source-hm} gives the protocol.

Privileged-solution input substantially improves the teacher's task-level accuracy, while a shuffled reference offers almost no advantage (Table~\ref{tab:source-solving}). A blind audit of the privileged-solution responses finds that 33.3\% of confirmed improvements closely reproduce the supplied solution, and only 45.7\% are judged self-contained (Appendix~\ref{app:source-hm-audit}). In the Thinking-ON OT audit, the corresponding rates are 10.0\% and 87.5\% (Appendix~\ref{app:source}). In this HM audit, more than half of confirmed improvements depend on the supplied solution rather than giving a self-contained derivation.

\subsection{Solution improves recovery after errors}
\label{sec:source-recovery}

To test recovery after a student error, we first segment OT Base trajectories into 128-token chunks. GPT-5.6-Sol and Claude Opus 4.8 each mark the earliest chunk containing a mathematical error. We retain 50 trajectories per model scale on which both judges identify an error. Each is cut after the later marked chunk, so that the final chunk of each prompt contains the error. The teacher generates four completions per prompt in each of the same three context conditions. The two judges independently assess each completion, with disagreements resolved by a separate blinded pass. We count recovery only if it repairs or abandons the error and reaches a valid solution (Appendix~\ref{app:source-ot-recovery}).

Table~\ref{tab:source-recovery} shows that the privileged solution raises Avg@4 by between 22.5 and 26.0 percentage points over problem only and by between 33.0 and 37.5 points over shuffled references. Every paired 95\% interval excludes zero (Appendix Table~\hyperref[app:tab-d2]{D2}). Because shuffled references underperform problem-only inputs, extra solution-like text does not explain the privileged gain. The added recovery benefit depends on task-matched information. The solution therefore remains actionable after a verified error, but recovery does not show that the teacher identified the error rather than re-solving from the supplied solution. Section~\ref{sec:source-recognition} tests this distinction directly.

\subsection{Error recognition remains limited}
\label{sec:source-recognition}

\begin{wrapfigure}{r}{0.52\textwidth}
  \vspace{-3mm}
  \centering
  \includegraphics[width=\linewidth]{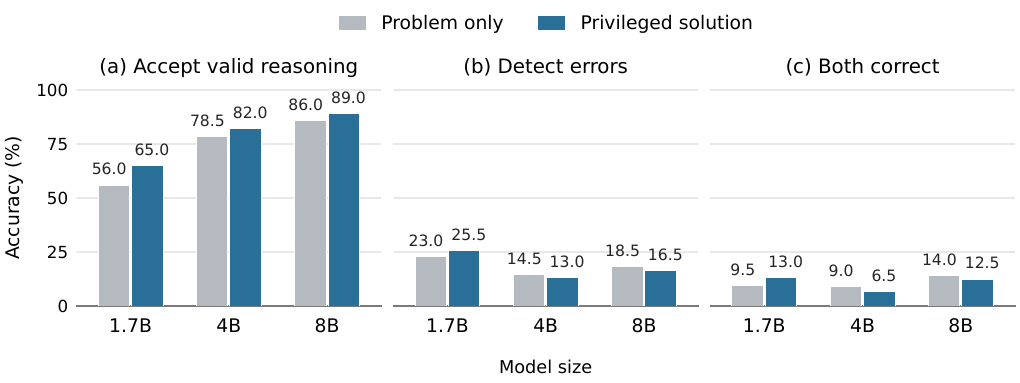}
  \vspace{-5mm}
  \caption{\textbf{Privileged solutions do not reliably improve error recognition.} Accuracy (\%) on 200 OT pairs of valid and erroneous prefixes per scale, with and without the privileged solution. ``Both correct'' requires both judgments in a pair to be correct. Paired 95\% intervals are in Appendix Table~\hyperref[app:tab-d3]{D3}.}
  \vspace{-2mm}
  \label{fig:error-recognition}
\end{wrapfigure}

Recovery alone does not show that the teacher identifies the error: it could solve the problem again using the supplied solution. To test explicit verbal recognition directly, as distinct from the implicit token-level signal that OPSD trains on, we use a separate set of 200 audited OT prefix pairs per scale. Each pair comes from the same trajectory and contains a valid prefix ending before either judge's marked error chunk and an erroneous prefix ending after both. The teacher judges whether each prefix is correct so far, with and without the privileged solution. We report accuracy on valid and erroneous prefixes, plus pair accuracy, which requires both judgments to be correct (Appendix~\ref{app:source-ot-recognition}).

With the privileged solution, the teacher correctly flags only between 13.0\% and 25.5\% of erroneous prefixes, lower than with problem only at 4B and 8B (Figure~\ref{fig:error-recognition}). Pair accuracy ranges from 6.5\% to 13.0\%, and the paired 95\% confidence interval for the difference from problem only includes zero at every scale. Overall prefix accuracy, averaged over valid and erroneous prefixes, shows a gain whose paired 95\% interval excludes zero only at 1.7B (+5.75pp). Even there, accuracy remains below the 50\% baseline of accepting every prefix (Appendix~\ref{app:source-ot-recognition}). The response parse rate ranges from 78.8\% to 97.0\%, and unparsed responses count as incorrect.

\vspace{4mm}
\takeawaybox{RQ1}{The privileged solution is actionable: across model scales, it improves hard-problem solving and recovery from verified errors, while shuffled references show that the benefit depends on task-matched information. Yet it does not reliably improve explicit error recognition. The teacher can therefore use the solution to produce a correct continuation without necessarily diagnosing where the student's reasoning failed. This reveals a gap between outcome guidance and error-sensitive supervision.}

\section{RQ2: Encoding}
\label{sec:encoding}

RQ2 tests whether the privileged solution changes the teacher's token-level supervision and whether that change tracks downstream gains, favors correct trajectories, or helps select better continuations.

\subsection{The solution changes the teacher's targets}
\label{sec:encoding-shift}

\begin{table}[t]
  \vspace{-2mm}
  \centering
  \footnotesize
  \setlength{\tabcolsep}{1pt}
  \renewcommand{\arraystretch}{1.0}
  \begin{minipage}[t]{0.48\textwidth}
  \vspace{0pt}
  \centering
  \begin{tabularx}{\linewidth}{@{}*{5}{C}@{}}
    \toprule
    \textbf{Trajectory} & \textbf{Teacher} &
    \textbf{1.7B} & \textbf{4B} & \textbf{8B} \\
    \midrule
    OFF & OFF & 0.051 & 0.037 & 0.033 \\
    OFF & ON & 0.042 & 0.035 & 0.031 \\
    ON & OFF & 0.138 & 0.104 & 0.099 \\
    ON & ON & 0.076 & 0.069 & 0.058 \\
    \bottomrule
  \end{tabularx}
  \vspace{-2mm}
  \caption{\textbf{Privileged solutions increase FKL across scales and thinking modes.} $\Delta\text{FKL}_{\text{ref}}$ on fixed Base trajectories. OFF/ON denote trajectory and teacher thinking. Appendix Table~\hyperref[app:tab-e1a]{E1a} reports paired 95\% confidence intervals.}
  \label{tab:reference-fkl}
  \end{minipage}%
  \hfill
  \begin{minipage}[t]{0.48\textwidth}
  \vspace{0pt}
  \centering
\newcommand{\celldelta}[1]{%
  {\scriptsize\ifdim #1pt>0pt
    \textcolor{green!45!black}{(#1)}%
  \else\ifdim #1pt<0pt
    \textcolor{red!70!black}{(#1)}%
  \else
    (#1)%
  \fi\fi}%
}
\begin{tabularx}{\linewidth}{@{}c@{\hspace{2pt}}|@{\hspace{2pt}}c*{2}{>{\centering\arraybackslash}X}@{}}
  \toprule
  \textbf{Model} & \textbf{States} & \textbf{\shortstack{Student selected}} & \textbf{\shortstack{Privileged selected}} \\
  \midrule
  1.7B & 149 & 57.38 & 63.09\,\celldelta{+5.70} \\
  4B & 150 & 73.67 & 74.33\,\celldelta{+0.67} \\
  8B & 150 & 80.50 & 79.33\,\celldelta{-1.17} \\
  \bottomrule
\end{tabularx}
\vspace{-2mm}
\caption{\textbf{Privileged selection improves success reliably only at 1.7B.} Avg@4 (\%) over four completions per selected branch from verified OT error states. Parentheses show changes from student selection (pp). Appendix Table~\hyperref[app:tab-e5]{E5} reports paired 95\% state-bootstrap intervals.}
\label{tab:branch-selection}

  \end{minipage}
\end{table}

We score fixed correct-incorrect Base trajectory pairs per question using teachers with and without the privileged solution. We define the change in forward KL as $\Delta\text{FKL}_{\text{ref}} = \text{FKL}(\text{privileged}) - \text{FKL}(\text{problem only})$, where each FKL measures teacher-to-student divergence over the same fixed prefixes. This quantity captures how privileged conditioning changes the divergence, not the distance between teacher targets. Thinking-OFF trajectories come from OT and Thinking-ON trajectories from HM. Each has at least 1,024 clean response tokens, and we score a 512-token suffix. Because the modes come from different corpora, their difference should be interpreted descriptively rather than as a controlled estimate of thinking-mode effects.

As shown in Table~\ref{tab:reference-fkl}, adding the solution increases FKL in all 12 settings. We focus on the Thinking-ON teacher used during training. Adding the solution produces nearly twice the FKL increase on trajectories generated by a Thinking-ON student compared with those generated by a Thinking-OFF student (1.80--1.97 times across model scales). The same pattern holds when both trajectory modes use the same HM questions (Appendix Table~\hyperref[app:tab-e2]{E2}). Yet the largest sustained benchmark gains come from training on Thinking-OFF rollouts, which have the smaller FKL changes (Figure~\ref{fig:interface-factorial}). A larger FKL change is therefore not a reliable indicator of a larger performance gain.

\subsection{Correctness ranking remains inconsistent}
\label{sec:encoding-ranking}

\begin{wrapfigure}{r}{0.49\textwidth}
  \vspace{-12mm}
  \centering
  \includegraphics[width=\linewidth]{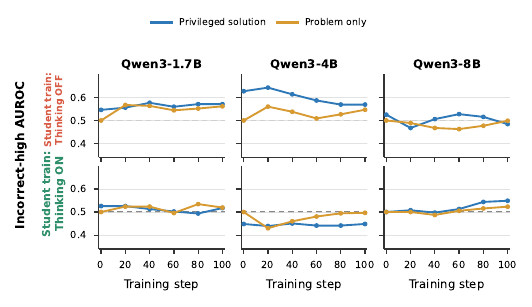}
  \vspace{-7mm}
  \caption{\textbf{The solution does not consistently improve correctness ranking.} Exact-FKL AUROC for ranking incorrect above correct trajectories. Trained checkpoints average three seeds. Problem-only AUROC is 0.5 at Step 0. Appendix~\ref{app:encoding-ranking} gives cohorts and clipped-loss values.}
  \vspace{-4mm}
  \label{fig:correctness-auroc}
\end{wrapfigure}

We test whether incorrect trajectories exhibit larger teacher-to-student divergence than correct ones, using exact FKL as a diagnostic of correctness sensitivity. We treat the incorrect trajectory as the positive class, so an AUROC above 0.5 indicates larger divergence on incorrect trajectories. This score measures distributional disagreement. It does not directly measure gradient strength or the usefulness of an update.

We reuse the correct-incorrect trajectory pairs from \S\ref{sec:encoding-shift}, keeping only those whose labels two judges (GPT-5.6-Sol and Claude Opus 4.8) confirm. For each trajectory and each checkpoint, we average the per-token exact FKL over a fixed 512-token suffix, computed between the frozen Thinking-ON teacher and the current student checkpoint. Comparing AUROC with and without the privileged solution isolates what the solution adds (Appendix~\ref{app:encoding-ranking}).

Results are shown in Figure~\ref{fig:correctness-auroc}. Overall, AUROC is close to 0.5 in most settings, with a temporary advantage for Qwen3-4B on Thinking-OFF trajectories. Hence, the gains shown in Figure~\ref{fig:interface-factorial} do not correspond to a persistent privilege-specific correctness ranking. Trajectory-level ranking is a coarse diagnostic rather than a necessary condition for useful token-local learning, so we conclude only that this FKL readout does not consistently separate correct from incorrect trajectories.

\WFclear
\subsection{Branch selection improves only at 1.7B}
\label{sec:encoding-branch}

Finally, we test whether the privileged solution helps select a better continuation at a state just before a student error. At each OT prefix ending before an error verified by GPT-5.6-Sol and Claude Opus 4.8, the student generates four 128-token branches. We compare branches selected by the student's mean log likelihood with those selected by the difference between the privileged teacher's and student's mean log likelihoods, estimating each branch's success as Avg@4 over four completions (Appendix~\ref{app:encoding-branch}). In a stratified audit of continuations marked incorrect by the automatic verifier, judges deem between 16.4\% and 64.6\% of non-ambiguous cases correct across model scales (Appendix Table~\hyperref[app:tab-e7]{E7}). We therefore use blinded task-level adjudication for all selector outcomes (Appendix~\ref{app:encoding-adjudication}).

Table~\ref{tab:branch-selection} shows a reliable advantage for the privileged selector only at 1.7B on OT (+5.70pp). At 4B and 8B on OT, the differences are +0.67pp and -1.17pp, and both paired intervals include zero. HM shows no robust effect at any model scale (Appendix Tables~\hyperref[app:tab-e5]{E5} and~\hyperref[app:tab-e6]{E6}). The tested privileged selector therefore shows no consistent advantage across scales and corpora.

\vspace{2mm}
\takeawaybox{RQ2}{The privileged solution clearly changes the teacher's token-level targets, increasing FKL across model scales and thinking modes. Yet the magnitude of this shift does not track the observed downstream gains: correctness ranking remains near chance in most settings, and branch selection improves reliably only at 1.7B. Privileged information therefore changes supervision more consistently than it aligns that supervision with correct reasoning.}


\section{RQ3: Transmission}
\label{sec:transmission}
\vspace{-1mm}

RQ3 examines what the student learns from OPSD. We test repeated training on the same questions, the contribution of privileged solutions, and whether gains persist during longer training.

\vspace{-1mm}
\subsection{Repeated training raises pass@8}
\label{sec:transmission-learning}
\vspace{-1mm}

\begin{figure}[t]
  \vspace{-2mm}
  \centering
  \begin{minipage}[t]{0.46\textwidth}
  \vspace{0pt}
  \centering
  \includegraphics[width=\linewidth]{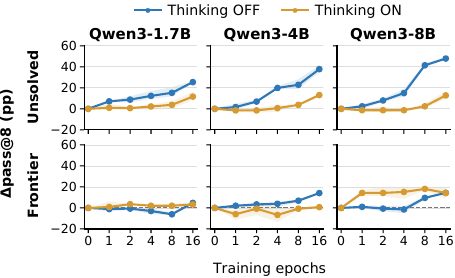}
  \vspace{-5mm}
  \caption{\textbf{Pass@8 gains from repeated training.} Change from each curve's Base on its own OT set. Thinking OFF and ON indicate matched student training and evaluation modes. Curves are three-seed means with seed-range shading.}
  \label{fig:training-pass8}
  \end{minipage}%
  \hfill
  \begin{minipage}[t]{0.48\textwidth}
  \vspace{0pt}
  \centering
  \includegraphics[width=\linewidth]{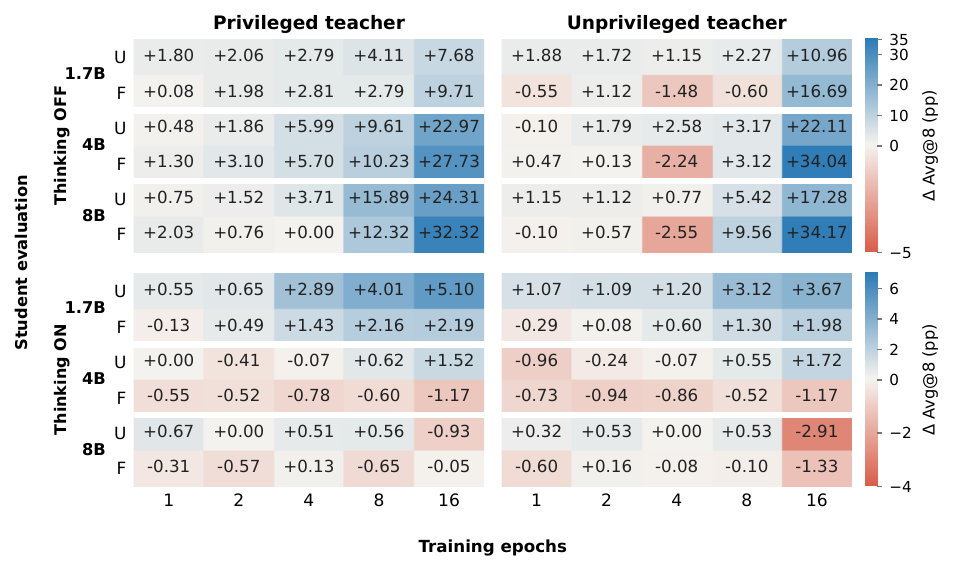}
  \vspace{-5mm}
  \caption{\textbf{Large Thinking-OFF gains occur with both teacher inputs.} A Thinking-ON teacher supervises Thinking-OFF rollouts. Columns vary teacher input. Rows vary evaluation mode and scale. U/F denote Unsolved/Frontier. Cells show three-seed $\Delta$Avg@8 (pp) from Base.}
  \label{fig:privilege-interface}
  \end{minipage}
\end{figure}

We train separately on OT \textbf{Unsolved} (0/12 correct Base responses) and \textbf{Frontier} (1--5/12) questions for about 16 epochs, testing whether repeated training helps on questions with no observed Base success or mainly makes occasional successes more frequent. These runs use a Thinking-ON privileged-solution teacher and report pass@8 (Figure~\ref{fig:training-pass8}). Figures~\ref{fig:training-pass8} and~\ref{fig:privilege-interface} evaluate the training questions, so they measure question-level fitting. Base pass@8 is 63--83\% on Frontier but only 7--24\% on Unsolved, and the two modes use different question sets. Because a 0/12 label can reflect missed successes or grading errors, we also train on a separate 0/16 set whose 16 rollouts both judges confirm incorrect: 60, 24, and 11 questions at 1.7B, 4B, and 8B, respectively (Appendix~\ref{app:verified-unsolved}).

Repeated training increases pass@8 on both question sets (Figure~\ref{fig:training-pass8}). The largest gains occur on Unsolved questions under Thinking OFF: 25.4--47.9pp after 16 epochs, versus 11.6--13.1pp under Thinking ON, so the student benefits from a Thinking-ON teacher with its own thinking block disabled. However, Thinking-OFF solving stays unreliable on the 0/16 set: at the final checkpoints, Avg@16 is only 3.2\% to 7.2\%, and no question reaches 14/16 correct responses with either teacher input, while the same checkpoints reach 61.7\% to 78.2\% under Thinking ON (Appendix Table~\hyperref[app:tab-f1g]{F1g}). These questions are unsolved under Thinking OFF, not globally, and 16 epochs on these small sets are only about 30, 12, and 6 optimizer updates. OPSD thus sharpens the output distribution toward correct samples, but within so few updates we do not observe transfer of Thinking-ON success to Thinking OFF.

\vspace{-1mm}
\subsection{Gains do not require the solution}
\label{sec:transmission-interface}
\vspace{-1mm}

To separate the privileged solution from the thinking mode, we fix teacher and rollout modes, question sets, schedule, and seeds and vary only whether the teacher receives the solution.

As shown in Figure~\ref{fig:privilege-interface}, after 16 epochs, teachers improve Thinking-OFF Avg@8 with or without the privileged solution. In fact, the problem-only teacher gives the larger Thinking-OFF gain in four of six scale and question-set comparisons. The pattern holds on held-out benchmarks: at 4B and 8B, problem-only teachers raise Thinking-OFF Avg@12 by 15.98--22.70pp, versus 5.68--10.66pp for privileged-solution teachers (Appendix Table~\hyperref[app:tab-f1f]{F1f}). In the one-pass runs of Figure~\ref{fig:interface-factorial}b, the privileged-solution teacher leads by 1.02pp to 1.91pp at Step 20 under Thinking-OFF evaluation, but this relative advantage reverses by Step 100 (Appendix Table~\hyperref[app:tab-b3]{C3}). For Thinking-ON student training, the problem-only teacher and student receive identical inputs and start with identical weights, so KL is zero and no update occurs. Adding the solution improves training-set Avg@8 by 2.8--7.4pp after 16 epochs (Appendix Figure~\hyperref[app:fig-f1]{F1}). A Thinking-ON teacher can therefore improve Thinking-OFF generation without a supplied solution.

\subsection{Purified OPSD limits late decline}
\label{sec:transmission-retention}

\begin{wrapfigure}{r}{0.5\textwidth}
  \vspace{-10mm}
  \centering
  \includegraphics[width=\linewidth]{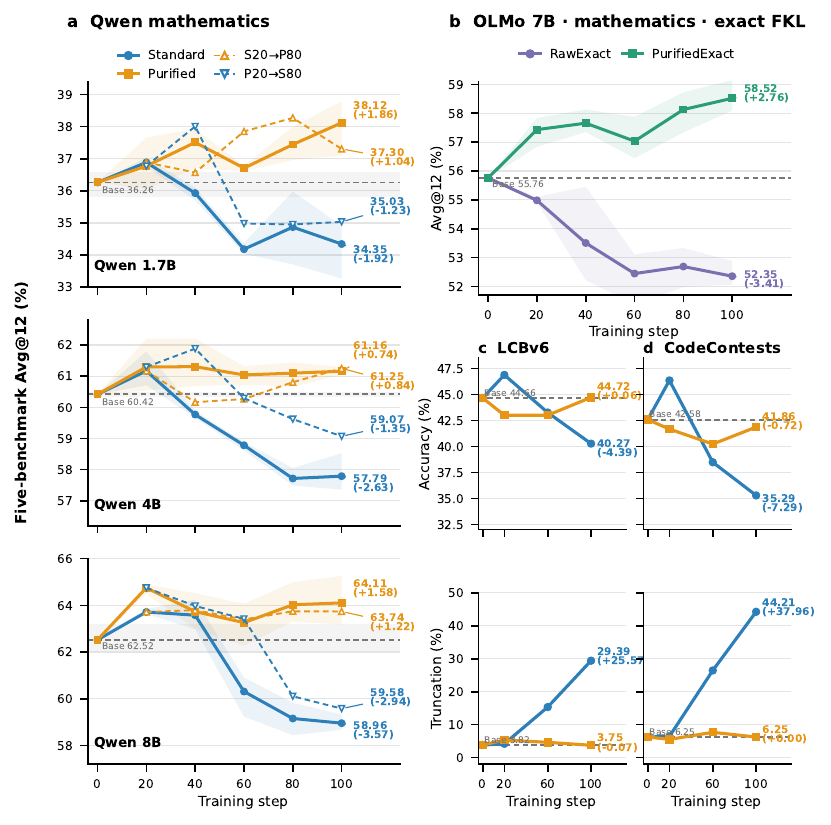}
  \vspace{-5mm}
  \caption{\textbf{Purified OPSD limits the late decline seen with Standard OPSD.} (a) Qwen3 mathematics, including runs that switch between methods. (b) OLMo with exact FKL fixed and only teacher targets changed. (c--d) Qwen3-4B coding accuracy and truncation.}
  \vspace{-4mm}
  \label{fig:retention}
\end{wrapfigure}

To test whether gains survive longer training, we compare Standard OPSD with Purified OPSD \citep{shen2026purified}. Purified discounts solution-only teacher signal and anchors its target to the problem-only Base model using three frozen-model passes and a Jensen--Shannon loss (Appendix~\ref{app:purified-target}). Teacher and student rollouts both use Thinking ON here, a setting in which a problem-only teacher yields no update. At Step 100, Purified scores 0.74--1.86pp above Base across Qwen3 scales, while Standard scores 1.92--3.57pp below it. Purified's gain is therefore evidence that solution-conditioned targets can improve held-out accuracy. Standard OPSD does not reliably deliver that signal. Switching from Standard to Purified recovers accuracy. The reverse switch ends below Base (Figure~\ref{fig:retention}a and Appendix~\ref{app:retention}).

The OLMo comparison holds exact FKL fixed for both raw and purified targets, isolating target construction (Figure~\ref{fig:retention}b). At Step 100, the purified target reaches 58.52\% Avg@12, above both the raw target at 52.35\% and Base at 55.76\%. The purified target outperforms the raw target in all three seeds, so this advantage does not require replacing FKL with the Jensen--Shannon loss used in the main Purified runs.

On Qwen3-4B coding, Standard improves early but then loses accuracy as truncation rises. At Step 100, Purified exceeds Standard by 4.45pp on LiveCodeBench v6 and 6.57pp on CodeContests, while its truncation rates remain near Base (Figure~\ref{fig:retention}c--d). The ordering persists with a longer coding rollout budget and in a single-seed Qwen3-4B mathematics continuation to 200 updates (Appendix~\ref{app:retention}).

\vspace{2mm}
\takeawaybox{RQ3}{Repeated OPSD raises pass@8, but its largest gains often need no solution and pair Thinking-ON teachers with Thinking-OFF rollouts, favoring reasoning arbitrage over new reasoning \mbox{transfer}. Purified OPSD better preserves accuracy, so target construction matters for retention.}

\section{RQ4: Expression}
\label{sec:expression}
\vspace{-1mm}

RQ4 asks whether training on verified corrections helps the student repair errors without the privileged solution. From mixed-outcome OT questions, where the Thinking-ON Base produces at least one correct and one incorrect trajectory, GPT-5.6-Sol marks a local error in each incorrect trajectory and writes a repair instruction. From that error prefix, the teacher generates corrections given the privileged solution and the instruction, and we keep verifier-correct, self-contained ones. 

\begin{wrapfigure}{r}{0.55\textwidth}
  \vspace{-3mm}
  \centering
  \includegraphics[width=\linewidth]{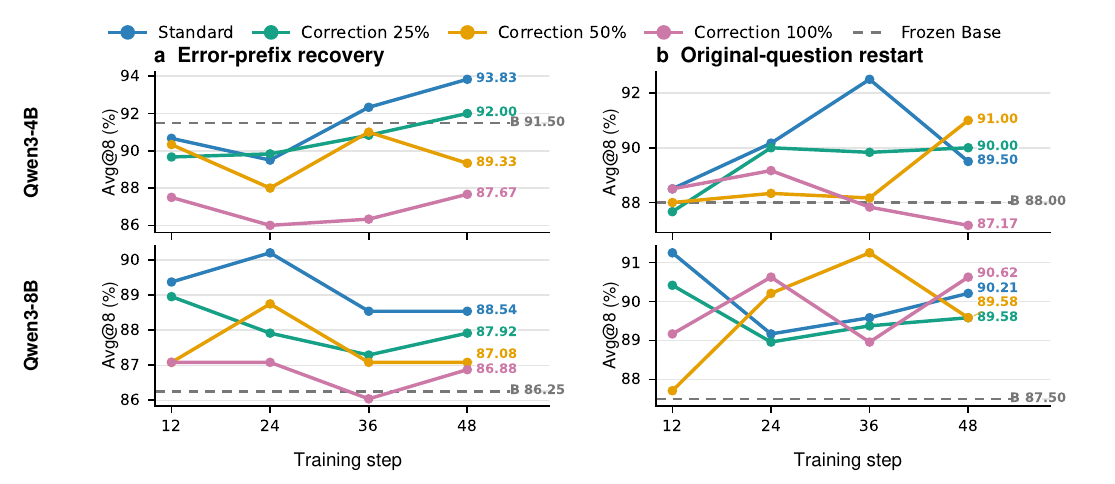}
  \vspace{-7mm}
  \caption{\textbf{Training on verified corrections does not consistently improve generation accuracy.} Accuracy on held-out questions starting (a) from the audited error prefix or (b) from the original problem, without the solution. Three-seed means. Dashed lines mark Base.}
  \vspace{-2mm}
  \label{fig:correction-expression}
\end{wrapfigure}

At Qwen3-4B and 8B, we split the correction dataset (125 and 102 questions) into training and held-out sets by question, holding out 25 and 20 questions. Frozen-Base accuracy on the held-out questions is already 86--92\% and each question is worth 4--5pp, so this is a small, near-ceiling pilot in which correction batches replace, rather than add to, Standard OPSD updates. \textbf{Correction25}, \textbf{Correction50}, and \textbf{Correction100} replace 25\%, 50\%, or 100\% of the Standard OPSD updates with supervised fine-tuning (SFT) on correction tokens. We report held-out Avg@8 when continuing from the audited error prefix (Figure~\ref{fig:correction-expression}a) or solving the original question (Figure~\ref{fig:correction-expression}b).

Held-out NLL on the correction targets falls monotonically with correction dose (Appendix Tables~\hyperref[app:tab-g4a]{G4a}--\hyperref[app:tab-g4b]{G4b}), so the student does learn to predict the corrections. Yet after 48 updates, all three correction variants have lower Avg@8 than Standard OPSD when generation starts from the error prefix. Starting from the original question, Correction25 and Correction50 at 4B and Correction100 at 8B are slightly above Standard OPSD, but no variant consistently outperforms it, and Correction100 lowers five-benchmark Avg@12 by 7.25pp at 4B and 2.42pp at 8B relative to Base. A same-prefix control at 4B attributes this decline to the correction continuations (Appendix~\ref{app:expression-controls}).

\WFclear
\vspace{2mm}
\takeawaybox{RQ4}{In this pilot, training on verified corrections lowers held-out correction NLL, but does not reliably improve error recovery and can reduce broader accuracy. Better correction prediction therefore does not imply that the student will apply the repair during generation.}

\section{Discussion}
\label{sec:discussion}

Together, the four audits show that the solution's effect on the frozen teacher does not map directly onto student performance. It improves teacher solving and recovery, but its token-level changes do not reliably favor correct trajectories. Student gains can occur without the solution, while learning to predict verified corrections does not consistently improve unaided repair. We now use additional controls to assess what privileged training contributes and why gains sometimes decline with continued training.

\begin{figure}[t]
  \centering
  \includegraphics[width=0.7\linewidth]{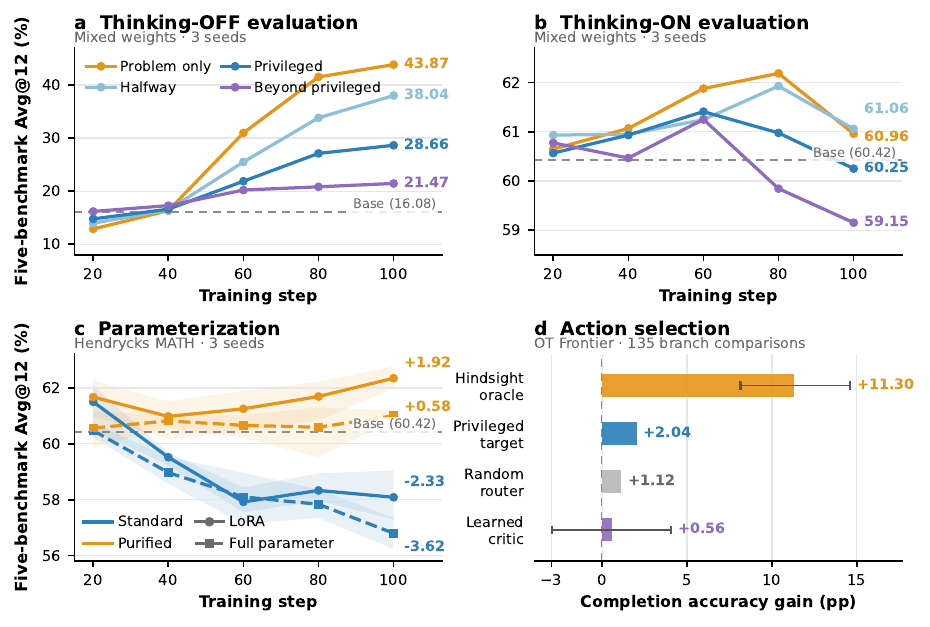}
  \vspace{-2mm}
  \caption{\textbf{Weight mixing, parameterization, and token-selection diagnostics.} (a--b) Weights mixed between problem-only and privileged-solution runs, including a 50\% extrapolation past the privileged endpoint. (c) LoRA versus full-parameter training. (d) Token selectors over 135 branch events, with 95\% intervals. (a--c) show three-seed means. Dashed lines mark the frozen Base checkpoint from which each run begins.}
  \label{fig:discussion-diagnostics}
\end{figure}

\noindent\textbf{Teacher success does not guarantee student gains.} To examine what the solution adds, we mix weights from matched problem-only and privileged-solution Standard OPSD runs at Qwen3-4B. At Step 100, the privileged endpoint and extrapolation beyond it score below problem-only weights under Thinking OFF, and slightly below under Thinking ON (Figure~\ref{fig:discussion-diagnostics}a,b). The privileged endpoint therefore offers no accuracy advantage in this comparison. The early gains and Purified improvements in RQ3 show that this late-training result does not extend to every use of solution-conditioned targets.

\vspace{1mm}
\noindent\textbf{The largest gains are consistent with reasoning arbitrage.} Problem-only teachers often match or exceed privileged-solution teachers (RQ3), showing that gains need not depend on privileged solutions. On verified Unsolved questions, the final checkpoints after approximately 16 epochs reach only 3.2\% to 7.2\% Thinking-OFF Avg@16, compared with 61.7\% to 78.2\% under Thinking ON (Appendix Table~\hyperref[app:tab-f1g]{F1g}). These questions remain difficult under the Thinking-OFF interface despite substantial success under Thinking ON. We hypothesize that some gains arise from making Thinking-ON behavior easier to elicit under Thinking OFF, resembling the compression role of \citet{kim2026opsd}. The mechanism underlying this pattern remains unresolved.

\vspace{1mm}
\noindent\textbf{The late decline persists beyond the default training settings.} Standard OPSD ends below Base with full-parameter training (Figure~\ref{fig:discussion-diagnostics}c), unclipped FKL, and longer rollouts (Appendix Tables~\hyperref[app:tab-h1b]{H1b} and~\hyperref[app:tab-h1d]{H1d}). These controls show that the decline persists across the tested settings, although the clipping and rollout controls use a single training seed. The matched exact-FKL OLMo comparison (RQ3) further shows that changing target construction can improve retention while keeping the loss fixed.

\noindent\textbf{Choosing useful targets and applying corrections require separate tests.} At Qwen3-4B, a hindsight oracle choosing between the student's and the target's preferred next token yields an estimated 11.30pp gain over 135 audited branch events, yet a held-out critic ranks near chance with pooled AUROC 0.452 (Figure~\ref{fig:discussion-diagnostics}d). FKL-selected sparse replay also shows no clear advantage over matched random replay in the single-seed pilot (Appendix~\ref{app:diagnostic-selection}). Useful local alternatives therefore do not by themselves establish reliable gains from selection or subsequent training. In RQ4, lower correction NLL likewise does not consistently improve recovery during generation. Target availability, selection, retention, and generation should be evaluated separately. Evaluations should also compare runs differing only in whether the teacher receives the solution, and evaluate Base and trained students under both thinking modes.

\vspace{-2mm}
\section{Conclusion}
\label{sec:conclusion}
\vspace{-2mm}
\enlargethispage{\baselineskip}

We audit how worked solutions shape supervision and student behavior in OPSD across three Qwen3 scales. Privileged solutions improve teacher solving and recovery, but do not reliably improve error recognition or correctness-sensitive token supervision. Learned corrections likewise do not consistently improve student recovery during generation. The largest gains at 4B and 8B recur with problem-only teachers and track the teacher's thinking mode. Purified OPSD better preserves accuracy, demonstrating that solution-conditioned training can still yield benefits. Together, these findings show that teacher competence and student accuracy gains alone do not establish transfer from privileged solutions. Evaluations should isolate what the solution adds through matched problem-only controls, assess both thinking modes, and test whether improvements persist through training and appear in unaided generation.


\bibliographystyle{assets/plainnat}
\bibliography{custom}

\clearpage
\beginappendix
\normalsize
\setcounter{dbltopnumber}{2}
\renewcommand{\dbltopfraction}{0.95}
\renewcommand{\dblfloatpagefraction}{0.8}
\setlength{\abovecaptionskip}{0.5pt}
\setlength{\belowcaptionskip}{0.5pt}
\makeatletter
\setlength{\@dblfptop}{0pt}
\setlength{\@dblfpsep}{8pt}
\setlength{\@dblfpbot}{0pt plus 1fil}
\makeatother

\noindent\textbf{Appendix organization.} Appendix~\ref{app:related} first situates the audit relative to concurrent work. Appendix~\ref{app:data} documents data preparation and the common training, prompt, and evaluation setup, and Appendix~\ref{app:figure1} reports the complete Figure~\ref{fig:interface-factorial} results. Appendices~\ref{app:source}-\ref{app:expression} provide the full Source, Encoding, Transmission, and Expression analyses, respectively. Appendix~\ref{app:discussion} reports additional diagnostics.

\section{Extended related work}
\label{app:related}

\noindent\textbf{Privileged information and on-policy self-distillation.} Learning using privileged information studies supervision that is available during training but absent at deployment \citep{vapnik2009new}, and generalized distillation unifies it with knowledge distillation \citep{lopezpaz2016unifying,hinton2015distilling}. On-policy distillation evaluates teacher targets on states sampled by the student rather than on fixed demonstrations \citep{agarwal2024onpolicy}. OPSD applies this idea with a frozen copy of the student conditioned on a privileged solution \citep{zhao2026selfdistilled}. Closely related formulations use a demonstration-conditioned EMA teacher for continual learning, an evolving feedback-conditioned teacher for reinforcement learning, a context-conditioned teacher for on-policy context distillation, or jointly trained privileged and unprivileged views for agent training \citep{shenfeld2026selfdistillation,hubotter2026reinforcement,ye2026onpolicy,penaloza2026privileged}. Recent variants alter the target or its credit assignment: Purified OPSD subtracts a reference-only component \citep{shen2026purified}, $\beta$-OPSD interpolates between a reference policy and the privileged teacher and adds return-to-go credit \citep{xu2026bopsd}, and AD-OPSD softly anchors high-risk tokens to a frozen base-model prior \citep{peng2026diagnosing}. Concurrent work also changes the teacher's privileged view or the supervision channel. PS-OPSD replaces a full solution with problem-space guidance \citep{zhao2026problemspace}. U-OPSD constructs pseudo-solutions from self-consistency \citep{li2026unsupervised}. Privileged Hidden Flow adds hidden-state transition matching \citep{li2026hiddenflow}. Context-design experiments compare full solutions with answer-only and multiple abstraction levels \citep{tian2026selfteacher}. Broader extensions select verified teachers across domains or apply privileged self-distillation to multi-turn GUI interaction \citep{he2026multiteacher,zhang2026gui}. We use Purified and $\beta$-OPSD as matched diagnostic controls rather than as a method leaderboard (\S\ref{sec:transmission-retention} and Appendix~\ref{app:discussion}).

\vspace{1mm}
\noindent\textbf{Source.} \citet{kaur2026rethinking} directly evaluate the gold-demonstration-conditioned model, whose pass@k remains high even as its rollouts shorten. RQ1 separately tests whether the frozen privileged teacher can solve problems that the Base model fails, repair audited erroneous student prefixes, and recognize where errors occur, using matched problem-only, shuffled-reference, and privileged-solution inputs.

\vspace{1mm}
\noindent\textbf{Encoding.} \citet{kim2026selfdistillation} trace self-distillation degradation to suppressed epistemic verbalization when the teacher conditions on rich information. \citet{kaur2026rethinking} find that privileged context degrades thinking models more at long budgets and with a full solution than with the final answer alone, and that privileged teachers penalize sampled reconsideration tokens. \citet{harne2026privileged} show that loss falls while accuracy stays flat or declines, tracing this to teacher targets pulled toward the supplied reference and to per-token losses that do not separate correct from incorrect rollouts. \citet{nguyen2026privileged} distinguish privileged-likelihood change from action value: on AIME 2025 with rollout-derived feedback, token scores discriminate correct from incorrect rollouts at near-chance level and training trails outcome-only reinforcement learning, whereas on SciKnowEval Biology score discrimination and learning agree. Concurrent context studies find that structured or abstracted guidance can outperform full solutions and that initial teacher-to-student KL does not order downstream utility \citep{zhao2026problemspace,tian2026selfteacher}. RQ2 complements these analyses by measuring target shift on identical tokens, correctness ranking across checkpoints, and sibling-branch selection at audited error states.

\vspace{1mm}
\noindent\textbf{Transmission.} \citet{ichihara2026privileged} replace the privileged solution with a problem and solution from another example and find that gains remain competitive with OPSD. \citet{shrestha2026rethinking} find that students improve without the correct reference and that another problem's solution can outperform it on several benchmarks. U-OPSD further constructs its teacher context without ground-truth supervision \citep{li2026unsupervised}. \citet{kaur2026rethinking} observe that disabling thinking during OPSD training preserves Base accuracy. \citet{kim2026opsd} find that in thinking-enabled mathematics OPSD acts more as compression than correction: training only on correct rollouts preserves accuracy while shortening responses, whereas training only on incorrect rollouts lowers it. \citet{zhu2026manyfaces} attribute OPSD failures in mathematics to the test-time absence of instance-specific privileged information, which leads the student to learn a privileged-information-free average of teachers. Concurrently, \citet{zhang2026privileged} report that reference-free distillation from a thinking-enabled teacher accounts for much of the gain on direct-response rollouts. RQ3 adds a full teacher, rollout, and evaluation factorial, a direct matched privileged-versus-problem-only contrast (Table~\hyperref[app:tab-b3]{C3}), strict Base-zero acquisition tests under repeated exposure, and retention checks across model families and domains.

\vspace{1mm}
\noindent\textbf{Expression.} \citet{kaur2026rethinking} count deliberation markers in free generation after training, but, to our knowledge, none of the analyses above tests whether corrections the student has been trained to predict are used without privileged support. RQ4 separates teacher-forced fit from unsupported and cue-supported use under matched controls.

\section{Data and common setup}
\label{app:data}

We audit the two mathematical source pools used to construct training sets or model-relative difficulty cohorts before inspecting any experimental outcome. GPT-5.6-Sol applies a fixed rubric: we retain an item only when the problem is well posed; the canonical answer is valid, consistent with the worked solution, and compatible with the final-answer evaluator; and the worked solution is valid or has only a minor issue that does not affect correctness. We exclude any item that cannot be validated under these criteria. Thus, cleaning depends on problem, answer, and reference quality rather than model performance. Code problems are instead filtered by execution (Appendix~\ref{app:data-code}). Appendices~\ref{app:figure1-training}-\ref{app:setup-eval} then describe the common training, prompt, and evaluation setup.

\subsection{OpenThoughts}
\label{app:data-ot}

From a 5,440-row candidate pool drawn from the mathematical reasoning subset of OpenThoughts \citep{guha2025openthoughts}, auditing and deduplication leave 3,322 unique usable examples. We retain a fixed 3,200-example subset for the Figure~\ref{fig:interface-factorial} training runs and the primary OpenThoughts analyses.

\subsection{Hendrycks MATH Level 5}
\label{app:data-hm}

Of the 2,304 official Hendrycks MATH Level-5 rows \citep{hendrycks2021measuring}, 2,242 pass the audit and constitute the cleaned Level-5 panel. For the separate 3,200-example training set, we deduplicate this panel, retain 2,241 usable Level-5 items, and add 959 audited Level-4 items.

\subsection{CodeContests and LiveCodeBench}
\label{app:data-code}

For CodeContests \citep{li2022competition}, we retain Python 3 problems with non-generated tests and reference programs that pass execution, yielding 3,200 training problems and a disjoint 128-problem held-out set. Exact and near-duplicate checks find no overlap with the 131-problem LiveCodeBench v6 panel \citep{jain2024livecodebench}.\looseness=-1

\subsection{Training}
\label{app:figure1-training}

Our primary Qwen3 mathematics experiments train \href{https://huggingface.co/Qwen/Qwen3-1.7B}{Qwen3-1.7B}, \href{https://huggingface.co/Qwen/Qwen3-4B}{Qwen3-4B}, and \href{https://huggingface.co/Qwen/Qwen3-8B}{Qwen3-8B} students \citep{qwen2025qwen3}. Here and throughout, Base denotes the unchanged checkpoint before additional training. The teacher is a frozen copy of that checkpoint, and only student LoRA parameters are updated \citep{hu2022lora}. We train with veRL \citep{sheng2025hybridflow} on the fixed 3,200-example OpenThoughts subset described in Appendix~\ref{app:data-ot}. Each run takes 100 optimizer steps with a global batch size of 32, processing every example once. Compared conditions within a scale share three seeds coupling data order, LoRA initialization, and rollout randomness. All local training and generation runs use NVIDIA H100 GPUs.
\begin{table*}[!t]
  \centering
  \small
  \setlength{\tabcolsep}{2pt}
  \renewcommand{\arraystretch}{1.0}
  \begin{tabularx}{\textwidth}{@{}>{\raggedright\arraybackslash}p{0.25\textwidth}>{\raggedright\arraybackslash}X@{}}
  \toprule
  \textbf{Training component} & \textbf{Setting} \\
  \midrule
  Update unit & One on-policy response per problem; global batch 32; 100 updates (one pass over 3,200 examples). \\
  Student rollout & Maximum 1,024 response tokens; temperature 1.1; top-p 0.95; top-k 20; presence penalty 0. \\
  Scoring & The frozen teacher and current student score the same sampled response tokens at every prefix under their respective inputs; loss temperature 1.1. \\
  Objective & Standard teacher-to-student forward KL; each vocabulary summand is upper-clipped at 0.05 before summation, then averaged over response positions; no post-vocabulary soft cap. \\
  Trainable parameters & LoRA rank 64, alpha 128; target modules: q\_proj, k\_proj, v\_proj, o\_proj, gate\_proj, up\_proj, and down\_proj. \\
  Optimizer & AdamW; constant learning rate $5\times10^{-6}$; no warmup; weight decay 0; gradient-norm clip 0.1. \\
  Precision and context & Student/FSDP bfloat16; frozen-teacher scoring float32; maximum sequence length 20,000 tokens (up to 18,976 prompt tokens plus 1,024 response tokens). \\
  \bottomrule
  \end{tabularx}
  \vspace{1mm}
  \caption*{\phantomsection\label{app:tab-b1}\textbf{Table B1:} Default training setup.}
\end{table*}

\begin{table*}[!t]
  \centering
  \small
  \setlength{\tabcolsep}{2pt}
  \renewcommand{\arraystretch}{1.0}
  \begin{tabularx}{\textwidth}{@{}>{\raggedright\arraybackslash}p{0.25\textwidth}>{\raggedright\arraybackslash}X@{}}
  \toprule
  \textbf{Evaluation parameter} & \textbf{Value} \\
  \midrule
  Samples per problem & 12 \\
  Decoding & Temperature 1.0; top-p 1.0; top-k disabled (-1); min-p 0; presence penalty 0. \\
  Maximum response length & 38,912 tokens \\
  Maximum context length & 40,960 tokens \\
  Inference precision & bfloat16 \\
  Scoring & Last well-formed \texttt{\textbackslash boxed\{...\}} expression after any closed thinking block; mathematical-equivalence grading. \\
  \bottomrule
  \end{tabularx}
  \vspace{1mm}
\caption*{\phantomsection\label{app:tab-b2}\textbf{Table B2:} Five-benchmark reference-free evaluation setup.}
\end{table*}

Table~\hyperref[app:tab-b1]{B1} summarizes the default training configuration. We reimplement the OPSD method of \citet{zhao2026selfdistilled} in veRL using the published configuration as a starting point. We use the reported learning rate of $5\times10^{-6}$ and the released default clipping threshold of 0.05 at every scale. The released 8B Thinking-ON script uses 0.06. The released 4B and 8B Thinking-OFF scripts instead use $10^{-6}$ and $10^{-7}$, respectively. We hold the threshold fixed across our conditions. The released loss function averages over valid tokens in each call, whereas our implementation normalizes by the valid-token count across the global minibatch.

\subsection{Thinking modes and prompts}
\label{app:figure1-interfaces}

Thinking ON and OFF are implemented with the Qwen3 chat template's \textbf{enable\_thinking} flag. The mode is set independently for teacher scoring, student rollouts, and evaluation, and each experiment states the modes it uses. All other prompt text and sampling parameters are held fixed within a comparison. The student and teacher use the following inputs.

\begin{samepage}
\noindent\textbf{Student and problem-only-teacher input:}

\nopagebreak
\begin{promptbox}
Problem: \{problem\}\\
Please reason step by step, and put your final answer within \textbackslash boxed\{\}.
\end{promptbox}
\end{samepage}

\begin{samepage}
\noindent\textbf{Privileged-solution-teacher input:}

\nopagebreak
\begin{promptbox}
Problem: \{problem\}\\
Here is a reference solution to this problem:\\
=== Reference Solution Begin ===\\
\{solution\}\\
=== Reference Solution End ===\\
After reading the reference solution above, make sure you truly understand the reasoning behind each step \textrm{\textemdash} do not copy or paraphrase it. Now, using your own words and independent reasoning, derive the same final answer to the problem above. Think step by step, explore different approaches, and don't be afraid to backtrack or reconsider if something doesn't work out:\\
Please reason step by step, and put your final answer within \textbackslash boxed\{\}.
\end{promptbox}
\end{samepage}

The shuffled-reference control uses the privileged-solution template with another problem's worked solution.

\subsection{Evaluation}
\label{app:setup-eval}

The public mathematics suite contains all problems from AIME 2024, AIME 2025, AIME 2026, HMMT November 2025, and HMMT February 2026 \citep{mathai2024aime,matharena2025aime,matharena2026aime,matharena2025hmmt,matharena2026hmmt}: 30, 30, 30, 30, and 33 items, and the macro average gives each benchmark equal weight. This suite is evaluated without the privileged solution under the stated Qwen3 thinking mode. RQ-specific audits give their own input and scoring protocols. Table~\hyperref[app:tab-b2]{B2} summarizes the suite's configuration. Avg@$k$ is mean accuracy over $k$ sampled responses per problem, and pass@$k$ is the fraction of problems with at least one correct response among $k$. Intervals over three training seeds are descriptive paired seed-bootstrap intervals rather than inferential confidence intervals.

\FloatBarrier

\section{\texorpdfstring{Figure~\ref{fig:interface-factorial}}{Figure 1}: complete results}
\label{app:figure1}

Figure~\ref{fig:interface-factorial} uses the common setup in Appendix~\ref{app:data}. Figure~\ref{fig:interface-factorial}a crosses teacher mode \{ON, OFF\}, student rollout mode \{ON, OFF\}, and evaluation mode \{ON, OFF\} at each model scale. Figure~\ref{fig:interface-factorial}b fixes the teacher to Thinking ON and the student rollout to Thinking OFF, and changes only whether the frozen teacher sees the privileged worked solution or the problem alone. 

Checkpoints at updates 20, 40, 60, 80, and 100 are evaluated on the five-benchmark suite with decoding seed 0. Figure~\ref{fig:interface-factorial} plots means over three training seeds. The frozen Base is evaluated independently with decoding seeds 0, 1, and 2 for every model and evaluation mode. Dashed lines show its three-seed mean and gray bands its 95\% seed-bootstrap interval. Colored intervals for checkpoint-minus-Base differences use 50,000 independent two-sample bootstrap resamples of the three training seeds and three Base decoding seeds. They are descriptive at n = 3 per side. In Tables~\hyperref[app:tab-b1a]{C1a}--\hyperref[app:tab-b2c]{C2c}, trained-cell accuracies and Base-relative deltas use these corresponding means, and the final column reports the same two-sample interval for the macro delta.

\subsection{\texorpdfstring{Figure~\ref{fig:interface-factorial}a}{Figure 1a}: thinking-mode factorial}

These tables expand Figure~\ref{fig:interface-factorial}a at Steps 20, 40, 60, 80, and 100. Every teacher in this factorial receives the privileged worked solution. Teacher and Student train denote the generation modes used to construct the teacher target and the on-policy student rollout, respectively. Eval denotes the reference-free generation mode used at evaluation.

\begin{table*}[!t]
  \centering
  \begingroup
  \fontsize{6.5pt}{7.5pt}\selectfont
  \setlength{\tabcolsep}{3pt}
  \renewcommand{\arraystretch}{0.95}
  \begin{tabularx}{\textwidth}{@{}cccc*{7}{>{\centering\arraybackslash}X}@{}}
  \toprule
  \textbf{Teacher} & \textbf{Student train} & \textbf{Eval} & \textbf{Step} & \textbf{AIME 2024} & \textbf{AIME 2025} & \textbf{AIME 2026} & \textbf{HMMT Nov. 2025} & \textbf{HMMT Feb. 2026} & \textbf{Macro average} & \textbf{Macro $\Delta$ 95\% CI} \\
  \midrule
  \rowcolor{baserow}Base & --- & OFF & 0 & 13.89 & 8.98 & 9.81 & 3.61 & 7.07 & 8.67 & --- \\
  \rowcolor{baserow}Base & --- & ON & 0 & 47.78 & 34.44 & 40.28 & 29.35 & 29.46 & 36.26 & --- \\
  \midrule[0.3pt]
  OFF & OFF & OFF & 20 & 12.59 (-1.30) & 10.83 (+1.85) & 9.44 (-0.37) & 3.70 (+0.09) & 7.49 (+0.42) & 8.81 (+0.14) & {[}-0.08, +0.33{]} \\
  \rowcolor{bestrow}OFF & OFF & OFF & 40 & 13.70 (-0.19) & 11.67 (+2.69) & 9.44 (-0.37) & 3.70 (+0.09) & 6.06 (-1.01) & 8.92 (+0.24) & {[}-0.02, +0.63{]} \\
  OFF & OFF & OFF & 60 & 11.85 (-2.04) & 9.07 (+0.09) & 7.41 (-2.41) & 2.78 (-0.83) & 5.05 (-2.02) & 7.23 (-1.44) & {[}-2.37, -0.92{]} \\
  OFF & OFF & OFF & 80 & 9.35 (-4.54) & 7.41 (-1.57) & 7.41 (-2.41) & 3.06 (-0.56) & 4.04 (-3.03) & 6.25 (-2.42) & {[}-3.18, -1.88{]} \\
  OFF & OFF & OFF & 100 & 9.81 (-4.07) & 6.85 (-2.13) & 6.11 (-3.70) & 2.78 (-0.83) & 3.45 (-3.62) & 5.80 (-2.87) & {[}-3.72, -2.21{]} \\
  \midrule[0.3pt]
  OFF & OFF & ON & 20 & 54.72 (+6.94) & 38.33 (+3.89) & 42.13 (+1.85) & 29.35 (+0.00) & 29.80 (+0.34) & 38.87 (+2.60) & {[}+2.03, +3.18{]} \\
  \rowcolor{bestrow}OFF & OFF & ON & 40 & 53.70 (+5.93) & 40.56 (+6.11) & 45.37 (+5.09) & 31.20 (+1.85) & 31.06 (+1.60) & 40.38 (+4.12) & {[}+3.34, +4.99{]} \\
  OFF & OFF & ON & 60 & 54.63 (+6.85) & 38.43 (+3.98) & 44.91 (+4.63) & 28.70 (-0.65) & 30.89 (+1.43) & 39.51 (+3.25) & {[}+2.58, +3.88{]} \\
  OFF & OFF & ON & 80 & 52.59 (+4.81) & 36.20 (+1.76) & 44.54 (+4.26) & 28.06 (-1.30) & 28.70 (-0.76) & 38.02 (+1.76) & {[}+0.74, +2.71{]} \\
  OFF & OFF & ON & 100 & 50.28 (+2.50) & 37.22 (+2.78) & 42.13 (+1.85) & 27.13 (-2.22) & 28.79 (-0.67) & 37.11 (+0.85) & {[}+0.10, +1.53{]} \\
  \midrule[0.3pt]
  \rowcolor{bestrow}OFF & ON & OFF & 20 & 9.17 (-4.72) & 10.09 (+1.11) & 6.76 (-3.06) & 3.98 (+0.37) & 6.23 (-0.84) & 7.25 (-1.43) & {[}-1.72, -1.21{]} \\
  OFF & ON & OFF & 40 & 9.81 (-4.07) & 8.06 (-0.93) & 7.50 (-2.31) & 2.78 (-0.83) & 4.04 (-3.03) & 6.44 (-2.24) & {[}-2.80, -1.38{]} \\
  OFF & ON & OFF & 60 & 8.89 (-5.00) & 7.59 (-1.39) & 7.22 (-2.59) & 2.96 (-0.65) & 2.61 (-4.46) & 5.86 (-2.82) & {[}-2.91, -2.73{]} \\
  OFF & ON & OFF & 80 & 8.33 (-5.56) & 7.13 (-1.85) & 6.76 (-3.06) & 2.59 (-1.02) & 2.69 (-4.38) & 5.50 (-3.17) & {[}-3.49, -2.85{]} \\
  OFF & ON & OFF & 100 & 8.15 (-5.74) & 6.48 (-2.50) & 6.94 (-2.87) & 2.22 (-1.39) & 2.53 (-4.55) & 5.26 (-3.41) & {[}-3.56, -3.27{]} \\
  \midrule[0.3pt]
  \rowcolor{bestrow}OFF & ON & ON & 20 & 48.24 (+0.46) & 32.87 (-1.57) & 37.22 (-3.06) & 24.91 (-4.44) & 26.94 (-2.53) & 34.04 (-2.23) & {[}-3.01, -1.43{]} \\
  OFF & ON & ON & 40 & 39.91 (-7.87) & 27.96 (-6.48) & 30.93 (-9.35) & 18.06 (-11.30) & 22.22 (-7.24) & 27.81 (-8.45) & {[}-9.54, -6.93{]} \\
  OFF & ON & ON & 60 & 22.41 (-25.37) & 16.57 (-17.87) & 15.00 (-25.28) & 9.17 (-20.19) & 13.47 (-15.99) & 15.32 (-20.94) & {[}-24.72, -18.66{]} \\
  OFF & ON & ON & 80 & 6.67 (-41.11) & 4.44 (-30.00) & 3.24 (-37.04) & 0.93 (-28.43) & 2.95 (-26.52) & 3.64 (-32.62) & {[}-34.45, -29.61{]} \\
  OFF & ON & ON & 100 & 2.69 (-45.09) & 1.11 (-33.33) & 0.93 (-39.35) & 0.37 (-28.98) & 1.26 (-28.20) & 1.27 (-34.99) & {[}-35.77, -34.14{]} \\
  \midrule[0.3pt]
  ON & OFF & OFF & 20 & 10.65 (-3.24) & 9.72 (+0.74) & 8.33 (-1.48) & 3.06 (-0.56) & 5.47 (-1.60) & 7.45 (-1.23) & {[}-2.41, -0.31{]} \\
  ON & OFF & OFF & 40 & 10.09 (-3.80) & 8.61 (-0.37) & 7.87 (-1.94) & 3.15 (-0.46) & 5.64 (-1.43) & 7.07 (-1.60) & {[}-1.96, -1.33{]} \\
  \rowcolor{bestrow}ON & OFF & OFF & 60 & 13.24 (-0.65) & 12.13 (+3.15) & 10.00 (+0.19) & 3.06 (-0.56) & 7.24 (+0.17) & 9.13 (+0.46) & {[}-0.10, +1.17{]} \\
  ON & OFF & OFF & 80 & 13.15 (-0.74) & 11.48 (+2.50) & 8.43 (-1.39) & 2.78 (-0.83) & 6.40 (-0.67) & 8.45 (-0.23) & {[}-0.64, +0.04{]} \\
  ON & OFF & OFF & 100 & 13.24 (-0.65) & 11.30 (+2.31) & 8.61 (-1.20) & 3.33 (-0.28) & 6.06 (-1.01) & 8.51 (-0.16) & {[}-1.28, +0.54{]} \\
  \midrule[0.3pt]
  ON & OFF & ON & 20 & 51.30 (+3.52) & 38.61 (+4.17) & 42.41 (+2.13) & 30.28 (+0.93) & 30.30 (+0.84) & 38.58 (+2.32) & {[}+1.66, +2.91{]} \\
  ON & OFF & ON & 40 & 55.00 (+7.22) & 38.89 (+4.44) & 44.72 (+4.44) & 28.15 (-1.20) & 31.57 (+2.10) & 39.66 (+3.40) & {[}+2.60, +4.14{]} \\
  \rowcolor{bestrow}ON & OFF & ON & 60 & 54.81 (+7.04) & 41.39 (+6.94) & 46.48 (+6.20) & 33.80 (+4.44) & 32.15 (+2.69) & 41.73 (+5.46) & {[}+4.57, +6.56{]} \\
  ON & OFF & ON & 80 & 54.81 (+7.04) & 41.67 (+7.22) & 45.00 (+4.72) & 30.74 (+1.39) & 31.90 (+2.44) & 40.82 (+4.56) & {[}+4.12, +5.02{]} \\
  ON & OFF & ON & 100 & 56.39 (+8.61) & 40.37 (+5.93) & 45.93 (+5.65) & 32.41 (+3.06) & 32.91 (+3.45) & 41.60 (+5.34) & {[}+4.93, +5.80{]} \\
  \midrule[0.3pt]
  ON & ON & OFF & 20 & 14.17 (+0.28) & 9.91 (+0.93) & 9.26 (-0.56) & 4.72 (+1.11) & 7.24 (+0.17) & 9.06 (+0.39) & {[}-0.03, +0.67{]} \\
  \rowcolor{bestrow}ON & ON & OFF & 40 & 14.26 (+0.37) & 12.41 (+3.43) & 9.54 (-0.28) & 3.98 (+0.37) & 7.91 (+0.84) & 9.62 (+0.95) & {[}+0.35, +1.36{]} \\
  ON & ON & OFF & 60 & 11.85 (-2.04) & 10.46 (+1.48) & 9.72 (-0.09) & 3.80 (+0.19) & 7.66 (+0.59) & 8.70 (+0.03) & {[}-0.05, +0.10{]} \\
  ON & ON & OFF & 80 & 11.67 (-2.22) & 10.37 (+1.39) & 8.52 (-1.30) & 4.07 (+0.46) & 6.48 (-0.59) & 8.22 (-0.45) & {[}-0.55, -0.35{]} \\
  ON & ON & OFF & 100 & 11.57 (-2.31) & 9.72 (+0.74) & 7.22 (-2.59) & 3.61 (+0.00) & 6.57 (-0.51) & 7.74 (-0.93) & {[}-1.24, -0.72{]} \\
  \midrule[0.3pt]
  \rowcolor{bestrow}ON & ON & ON & 20 & 51.67 (+3.89) & 35.65 (+1.20) & 40.83 (+0.56) & 29.44 (+0.09) & 31.14 (+1.68) & 37.75 (+1.48) & {[}+0.87, +2.10{]} \\
  ON & ON & ON & 40 & 49.63 (+1.85) & 33.52 (-0.93) & 39.35 (-0.93) & 27.78 (-1.57) & 27.53 (-1.94) & 35.56 (-0.70) & {[}-1.27, -0.10{]} \\
  ON & ON & ON & 60 & 47.78 (+0.00) & 31.85 (-2.59) & 37.31 (-2.96) & 25.09 (-4.26) & 27.69 (-1.77) & 33.95 (-2.32) & {[}-3.12, -1.50{]} \\
  ON & ON & ON & 80 & 49.35 (+1.57) & 32.69 (-1.76) & 36.57 (-3.70) & 25.09 (-4.26) & 28.20 (-1.26) & 34.38 (-1.88) & {[}-2.69, -1.14{]} \\
  ON & ON & ON & 100 & 47.41 (-0.37) & 32.31 (-2.13) & 37.78 (-2.50) & 23.80 (-5.56) & 27.69 (-1.77) & 33.80 (-2.46) & {[}-3.22, -1.69{]} \\
  \bottomrule
  \end{tabularx}
  \endgroup
  \vspace{1mm}
  \caption*{\phantomsection\label{app:tab-b1a}\textbf{Table C1a:} \textbf{Qwen3-1.7B} per-benchmark Avg@12 accuracy at all Figure~\ref{fig:interface-factorial}a checkpoints. Accuracies and parenthetical Base deltas use three-seed means. The final column gives the independent two-sample 95\% seed-bootstrap interval for the macro delta. Shaded rows mark Base (gray) and the step with the highest macro accuracy within each setup (yellow).}
\end{table*}
\begin{table*}[!t]
  \centering
  \begingroup
  \fontsize{6.5pt}{7.5pt}\selectfont
  \setlength{\tabcolsep}{3pt}
  \renewcommand{\arraystretch}{0.95}
  \begin{tabularx}{\textwidth}{@{}cccc*{7}{>{\centering\arraybackslash}X}@{}}
  \toprule
  \textbf{Teacher} & \textbf{Student train} & \textbf{Eval} & \textbf{Step} & \textbf{AIME 2024} & \textbf{AIME 2025} & \textbf{AIME 2026} & \textbf{HMMT Nov. 2025} & \textbf{HMMT Feb. 2026} & \textbf{Macro average} & \textbf{Macro $\Delta$ 95\% CI} \\
  \midrule
  \rowcolor{baserow}Base & --- & OFF & 0 & 22.78 & 18.43 & 15.09 & 7.87 & 16.25 & 16.08 & --- \\
  \rowcolor{baserow}Base & --- & ON & 0 & 73.33 & 65.74 & 66.48 & 55.37 & 41.16 & 60.42 & --- \\
  \midrule[0.3pt]
  \rowcolor{bestrow}OFF & OFF & OFF & 20 & 21.76 (-1.02) & 18.24 (-0.19) & 15.37 (+0.28) & 7.41 (-0.46) & 14.31 (-1.94) & 15.42 (-0.66) & {[}-1.07, -0.23{]} \\
  OFF & OFF & OFF & 40 & 22.31 (-0.46) & 16.85 (-1.57) & 14.63 (-0.46) & 7.04 (-0.83) & 12.29 (-3.96) & 14.62 (-1.46) & {[}-1.76, -1.02{]} \\
  OFF & OFF & OFF & 60 & 21.02 (-1.76) & 15.74 (-2.69) & 15.46 (+0.37) & 5.19 (-2.69) & 11.95 (-4.29) & 13.87 (-2.21) & {[}-2.66, -1.76{]} \\
  OFF & OFF & OFF & 80 & 22.22 (-0.56) & 18.98 (+0.56) & 16.57 (+1.48) & 6.02 (-1.85) & 12.63 (-3.62) & 15.28 (-0.80) & {[}-1.36, -0.24{]} \\
  OFF & OFF & OFF & 100 & 22.04 (-0.74) & 20.37 (+1.94) & 15.74 (+0.65) & 5.83 (-2.04) & 12.21 (-4.04) & 15.24 (-0.85) & {[}-1.43, -0.19{]} \\
  \midrule[0.3pt]
  OFF & OFF & ON & 20 & 75.19 (+1.85) & 66.57 (+0.83) & 66.67 (+0.19) & 52.87 (-2.50) & 40.99 (-0.17) & 60.46 (+0.04) & {[}-0.92, +0.61{]} \\
  \rowcolor{bestrow}OFF & OFF & ON & 40 & 75.93 (+2.59) & 65.09 (-0.65) & 67.96 (+1.48) & 53.61 (-1.76) & 41.08 (-0.08) & 60.73 (+0.32) & {[}+0.11, +0.49{]} \\
  OFF & OFF & ON & 60 & 75.83 (+2.50) & 65.28 (-0.46) & 68.80 (+2.31) & 50.83 (-4.54) & 42.17 (+1.01) & 60.58 (+0.16) & {[}-0.26, +0.76{]} \\
  OFF & OFF & ON & 80 & 74.07 (+0.74) & 64.81 (-0.93) & 66.48 (+0.00) & 50.19 (-5.19) & 41.84 (+0.67) & 59.48 (-0.94) & {[}-1.06, -0.81{]} \\
  OFF & OFF & ON & 100 & 73.70 (+0.37) & 65.93 (+0.19) & 67.87 (+1.39) & 52.50 (-2.87) & 42.09 (+0.93) & 60.42 (+0.00) & {[}-0.26, +0.24{]} \\
  \midrule[0.3pt]
  \rowcolor{bestrow}OFF & ON & OFF & 20 & 21.20 (-1.57) & 18.80 (+0.37) & 16.39 (+1.30) & 7.59 (-0.28) & 15.82 (-0.42) & 15.96 (-0.12) & {[}-0.79, +0.49{]} \\
  OFF & ON & OFF & 40 & 20.28 (-2.50) & 16.94 (-1.48) & 12.96 (-2.13) & 5.93 (-1.94) & 11.95 (-4.29) & 13.61 (-2.47) & {[}-3.12, -1.85{]} \\
  OFF & ON & OFF & 60 & 17.59 (-5.19) & 14.54 (-3.89) & 11.76 (-3.33) & 5.19 (-2.69) & 10.35 (-5.89) & 11.89 (-4.20) & {[}-4.83, -3.54{]} \\
  OFF & ON & OFF & 80 & 16.11 (-6.67) & 12.69 (-5.74) & 12.04 (-3.06) & 3.98 (-3.89) & 8.59 (-7.66) & 10.68 (-5.40) & {[}-6.13, -4.71{]} \\
  OFF & ON & OFF & 100 & 14.72 (-8.06) & 11.20 (-7.22) & 11.48 (-3.61) & 4.54 (-3.33) & 8.00 (-8.25) & 9.99 (-6.09) & {[}-6.61, -5.58{]} \\
  \midrule[0.3pt]
  \rowcolor{bestrow}OFF & ON & ON & 20 & 71.76 (-1.57) & 65.56 (-0.19) & 66.30 (-0.19) & 51.85 (-3.52) & 40.91 (-0.25) & 59.27 (-1.14) & {[}-1.48, -0.65{]} \\
  OFF & ON & ON & 40 & 69.63 (-3.70) & 60.00 (-5.74) & 64.81 (-1.67) & 49.35 (-6.02) & 39.98 (-1.18) & 56.76 (-3.66) & {[}-3.77, -3.55{]} \\
  OFF & ON & ON & 60 & 67.22 (-6.11) & 55.74 (-10.00) & 61.39 (-5.09) & 46.76 (-8.61) & 36.36 (-4.80) & 53.49 (-6.92) & {[}-8.55, -6.02{]} \\
  OFF & ON & ON & 80 & 57.13 (-16.20) & 44.07 (-21.67) & 50.46 (-16.02) & 37.59 (-17.78) & 31.23 (-9.93) & 44.10 (-16.32) & {[}-19.79, -11.84{]} \\
  OFF & ON & ON & 100 & 35.00 (-38.33) & 24.72 (-41.02) & 29.44 (-37.04) & 24.54 (-30.83) & 24.16 (-17.00) & 27.57 (-32.85) & {[}-36.03, -26.87{]} \\
  \midrule[0.3pt]
  ON & OFF & OFF & 20 & 22.04 (-0.74) & 17.78 (-0.65) & 13.89 (-1.20) & 7.31 (-0.56) & 12.96 (-3.28) & 14.80 (-1.29) & {[}-1.73, -0.84{]} \\
  ON & OFF & OFF & 40 & 24.81 (+2.04) & 19.35 (+0.93) & 16.94 (+1.85) & 6.85 (-1.02) & 15.15 (-1.09) & 16.62 (+0.54) & {[}-0.07, +1.15{]} \\
  ON & OFF & OFF & 60 & 29.91 (+7.13) & 24.26 (+5.83) & 23.61 (+8.52) & 12.41 (+4.54) & 19.28 (+3.03) & 21.89 (+5.81) & {[}+5.32, +6.30{]} \\
  ON & OFF & OFF & 80 & 34.26 (+11.48) & 28.98 (+10.56) & 32.69 (+17.59) & 17.78 (+9.91) & 21.97 (+5.72) & 27.13 (+11.05) & {[}+8.16, +12.81{]} \\
  \rowcolor{bestrow}ON & OFF & OFF & 100 & 37.31 (+14.54) & 31.67 (+13.24) & 32.41 (+17.31) & 19.54 (+11.67) & 22.39 (+6.14) & 28.66 (+12.58) & {[}+11.28, +13.75{]} \\
  \midrule[0.3pt]
  ON & OFF & ON & 20 & 74.26 (+0.93) & 68.80 (+3.06) & 66.39 (-0.09) & 52.31 (-3.06) & 41.08 (-0.08) & 60.57 (+0.15) & {[}-0.33, +0.59{]} \\
  ON & OFF & ON & 40 & 74.54 (+1.20) & 68.43 (+2.69) & 66.57 (+0.09) & 52.87 (-2.50) & 42.26 (+1.09) & 60.93 (+0.52) & {[}+0.09, +1.20{]} \\
  \rowcolor{bestrow}ON & OFF & ON & 60 & 75.46 (+2.13) & 68.43 (+2.69) & 68.06 (+1.57) & 52.59 (-2.78) & 42.51 (+1.35) & 61.41 (+0.99) & {[}-0.43, +1.95{]} \\
  ON & OFF & ON & 80 & 75.65 (+2.31) & 67.22 (+1.48) & 68.06 (+1.57) & 52.04 (-3.33) & 41.92 (+0.76) & 60.98 (+0.56) & {[}+0.07, +0.90{]} \\
  ON & OFF & ON & 100 & 73.24 (-0.09) & 67.50 (+1.76) & 65.37 (-1.11) & 53.15 (-2.22) & 42.00 (+0.84) & 60.25 (-0.16) & {[}-1.15, +0.75{]} \\
  \midrule[0.3pt]
  \rowcolor{bestrow}ON & ON & OFF & 20 & 22.41 (-0.37) & 19.07 (+0.65) & 15.83 (+0.74) & 8.43 (+0.56) & 15.66 (-0.59) & 16.28 (+0.20) & {[}-0.57, +0.87{]} \\
  ON & ON & OFF & 40 & 23.61 (+0.83) & 19.26 (+0.83) & 17.22 (+2.13) & 6.02 (-1.85) & 15.24 (-1.01) & 16.27 (+0.19) & {[}-0.98, +1.17{]} \\
  ON & ON & OFF & 60 & 21.48 (-1.30) & 19.17 (+0.74) & 15.37 (+0.28) & 7.31 (-0.56) & 14.39 (-1.85) & 15.55 (-0.54) & {[}-0.89, -0.09{]} \\
  ON & ON & OFF & 80 & 23.06 (+0.28) & 19.54 (+1.11) & 15.00 (-0.09) & 6.48 (-1.39) & 14.39 (-1.85) & 15.69 (-0.39) & {[}-0.98, +0.23{]} \\
  ON & ON & OFF & 100 & 22.04 (-0.74) & 18.80 (+0.37) & 15.83 (+0.74) & 6.85 (-1.02) & 13.97 (-2.27) & 15.50 (-0.58) & {[}-0.95, -0.17{]} \\
  \midrule[0.3pt]
  \rowcolor{bestrow}ON & ON & ON & 20 & 73.70 (+0.37) & 67.31 (+1.57) & 67.50 (+1.02) & 51.57 (-3.80) & 41.67 (+0.51) & 60.35 (-0.07) & {[}-0.25, +0.09{]} \\
  ON & ON & ON & 40 & 72.04 (-1.30) & 63.61 (-2.13) & 67.13 (+0.65) & 53.24 (-2.13) & 42.09 (+0.93) & 59.62 (-0.80) & {[}-1.69, -0.27{]} \\
  ON & ON & ON & 60 & 72.41 (-0.93) & 62.69 (-3.06) & 66.20 (-0.28) & 52.41 (-2.96) & 40.91 (-0.25) & 58.92 (-1.49) & {[}-2.52, -0.69{]} \\
  ON & ON & ON & 80 & 71.48 (-1.85) & 63.15 (-2.59) & 64.54 (-1.94) & 48.52 (-6.85) & 41.16 (+0.00) & 57.77 (-2.65) & {[}-3.07, -2.06{]} \\
  ON & ON & ON & 100 & 70.28 (-3.06) & 61.67 (-4.07) & 62.87 (-3.61) & 53.15 (-2.22) & 42.09 (+0.93) & 58.01 (-2.41) & {[}-2.99, -1.88{]} \\
  \bottomrule
  \end{tabularx}
  \endgroup
  \vspace{1mm}
  \caption*{\phantomsection\label{app:tab-b1b}\textbf{Table C1b:} \textbf{Qwen3-4B} per-benchmark Avg@12 accuracy at all Figure~\ref{fig:interface-factorial}a checkpoints. Accuracies and parenthetical Base deltas use three-seed means. The final column gives the independent two-sample 95\% seed-bootstrap interval for the macro delta. Shaded rows mark Base (gray) and the step with the highest macro accuracy within each setup (yellow).}
\end{table*}
\begin{table*}[!t]
  \centering
  \begingroup
  \fontsize{6.5pt}{7.5pt}\selectfont
  \setlength{\tabcolsep}{3pt}
  \renewcommand{\arraystretch}{0.95}
  \begin{tabularx}{\textwidth}{@{}cccc*{7}{>{\centering\arraybackslash}X}@{}}
  \toprule
  \textbf{Teacher} & \textbf{Student train} & \textbf{Eval} & \textbf{Step} & \textbf{AIME 2024} & \textbf{AIME 2025} & \textbf{AIME 2026} & \textbf{HMMT Nov. 2025} & \textbf{HMMT Feb. 2026} & \textbf{Macro average} & \textbf{Macro $\Delta$ 95\% CI} \\
  \midrule
  \rowcolor{baserow}Base & --- & OFF & 0 & 27.22 & 20.65 & 17.31 & 9.54 & 17.00 & 18.35 & --- \\
  \rowcolor{baserow}Base & --- & ON & 0 & 75.00 & 68.98 & 69.26 & 54.26 & 45.12 & 62.52 & --- \\
  \midrule[0.3pt]
  OFF & OFF & OFF & 20 & 21.20 (-6.02) & 17.87 (-2.78) & 14.26 (-3.06) & 6.67 (-2.87) & 13.47 (-3.54) & 14.69 (-3.65) & {[}-4.29, -3.02{]} \\
  OFF & OFF & OFF & 40 & 23.89 (-3.33) & 18.24 (-2.41) & 16.11 (-1.20) & 6.48 (-3.06) & 14.90 (-2.10) & 15.92 (-2.42) & {[}-3.08, -1.77{]} \\
  \rowcolor{bestrow}OFF & OFF & OFF & 60 & 25.46 (-1.76) & 21.11 (+0.46) & 17.31 (+0.00) & 8.15 (-1.39) & 15.40 (-1.60) & 17.49 (-0.86) & {[}-1.56, -0.11{]} \\
  OFF & OFF & OFF & 80 & 24.81 (-2.41) & 20.09 (-0.56) & 16.76 (-0.56) & 9.17 (-0.37) & 14.90 (-2.10) & 17.15 (-1.20) & {[}-1.71, -0.63{]} \\
  OFF & OFF & OFF & 100 & 26.02 (-1.20) & 21.48 (+0.83) & 15.74 (-1.57) & 8.70 (-0.83) & 14.39 (-2.61) & 17.27 (-1.08) & {[}-1.71, -0.45{]} \\
  \midrule[0.3pt]
  OFF & OFF & ON & 20 & 75.19 (+0.19) & 69.54 (+0.56) & 69.91 (+0.65) & 56.94 (+2.69) & 46.04 (+0.93) & 63.52 (+1.00) & {[}+0.25, +1.75{]} \\
  OFF & OFF & ON & 40 & 76.94 (+1.94) & 69.35 (+0.37) & 70.65 (+1.39) & 58.70 (+4.44) & 45.88 (+0.76) & 64.30 (+1.78) & {[}+0.83, +2.73{]} \\
  \rowcolor{bestrow}OFF & OFF & ON & 60 & 77.87 (+2.87) & 71.76 (+2.78) & 69.07 (-0.19) & 58.15 (+3.89) & 47.05 (+1.94) & 64.78 (+2.26) & {[}+1.57, +2.95{]} \\
  OFF & OFF & ON & 80 & 76.76 (+1.76) & 70.83 (+1.85) & 69.72 (+0.46) & 56.57 (+2.31) & 46.89 (+1.77) & 64.15 (+1.63) & {[}+1.00, +2.14{]} \\
  OFF & OFF & ON & 100 & 77.13 (+2.13) & 69.35 (+0.37) & 68.70 (-0.56) & 57.31 (+3.06) & 45.79 (+0.67) & 63.66 (+1.13) & {[}+0.46, +1.64{]} \\
  \midrule[0.3pt]
  \rowcolor{bestrow}OFF & ON & OFF & 20 & 26.02 (-1.20) & 19.35 (-1.30) & 15.09 (-2.22) & 9.44 (-0.09) & 15.57 (-1.43) & 17.10 (-1.25) & {[}-2.07, -0.34{]} \\
  OFF & ON & OFF & 40 & 24.44 (-2.78) & 19.35 (-1.30) & 13.52 (-3.80) & 8.52 (-1.02) & 14.73 (-2.27) & 16.11 (-2.23) & {[}-2.81, -1.66{]} \\
  OFF & ON & OFF & 60 & 23.70 (-3.52) & 19.07 (-1.57) & 13.98 (-3.33) & 7.13 (-2.41) & 14.23 (-2.78) & 15.62 (-2.72) & {[}-3.55, -1.78{]} \\
  OFF & ON & OFF & 80 & 22.50 (-4.72) & 18.15 (-2.50) & 14.35 (-2.96) & 6.94 (-2.59) & 12.37 (-4.63) & 14.86 (-3.48) & {[}-4.23, -2.73{]} \\
  OFF & ON & OFF & 100 & 20.93 (-6.30) & 14.26 (-6.39) & 12.04 (-5.28) & 6.48 (-3.06) & 11.20 (-5.81) & 12.98 (-5.37) & {[}-5.93, -4.80{]} \\
  \midrule[0.3pt]
  \rowcolor{bestrow}OFF & ON & ON & 20 & 75.37 (+0.37) & 69.26 (+0.28) & 68.24 (-1.02) & 58.80 (+4.54) & 45.96 (+0.84) & 63.53 (+1.00) & {[}+0.32, +1.60{]} \\
  OFF & ON & ON & 40 & 73.98 (-1.02) & 66.57 (-2.41) & 67.78 (-1.48) & 57.22 (+2.96) & 45.45 (+0.34) & 62.20 (-0.32) & {[}-1.20, +0.57{]} \\
  OFF & ON & ON & 60 & 69.91 (-5.09) & 64.17 (-4.81) & 66.85 (-2.41) & 54.81 (+0.56) & 43.10 (-2.02) & 59.77 (-2.76) & {[}-4.09, -1.54{]} \\
  OFF & ON & ON & 80 & 72.22 (-2.78) & 65.09 (-3.89) & 66.85 (-2.41) & 53.98 (-0.28) & 43.18 (-1.94) & 60.27 (-2.26) & {[}-3.16, -1.35{]} \\
  OFF & ON & ON & 100 & 64.91 (-10.09) & 55.74 (-13.24) & 60.83 (-8.43) & 46.94 (-7.31) & 39.14 (-5.98) & 53.51 (-9.01) & {[}-10.49, -7.84{]} \\
  \midrule[0.3pt]
  ON & OFF & OFF & 20 & 22.31 (-4.91) & 17.78 (-2.87) & 14.44 (-2.87) & 6.39 (-3.15) & 16.16 (-0.84) & 15.42 (-2.93) & {[}-3.87, -1.87{]} \\
  ON & OFF & OFF & 40 & 23.98 (-3.24) & 20.93 (+0.28) & 18.43 (+1.11) & 8.98 (-0.56) & 16.50 (-0.51) & 17.76 (-0.58) & {[}-1.12, -0.02{]} \\
  ON & OFF & OFF & 60 & 30.37 (+3.15) & 23.06 (+2.41) & 22.78 (+5.46) & 13.43 (+3.89) & 19.44 (+2.44) & 21.81 (+3.47) & {[}+2.35, +4.93{]} \\
  ON & OFF & OFF & 80 & 37.50 (+10.28) & 26.76 (+6.11) & 28.15 (+10.83) & 17.22 (+7.69) & 22.39 (+5.39) & 26.40 (+8.06) & {[}+5.14, +12.48{]} \\
  \rowcolor{bestrow}ON & OFF & OFF & 100 & 48.06 (+20.83) & 34.44 (+13.80) & 39.81 (+22.50) & 30.74 (+21.20) & 27.27 (+10.27) & 36.07 (+17.72) & {[}+14.13, +24.26{]} \\
  \midrule[0.3pt]
  ON & OFF & ON & 20 & 75.74 (+0.74) & 70.37 (+1.39) & 70.46 (+1.20) & 56.94 (+2.69) & 46.72 (+1.60) & 64.05 (+1.52) & {[}+0.89, +2.11{]} \\
  ON & OFF & ON & 40 & 76.11 (+1.11) & 70.93 (+1.94) & 70.19 (+0.93) & 58.33 (+4.07) & 45.79 (+0.67) & 64.27 (+1.75) & {[}+1.11, +2.31{]} \\
  ON & OFF & ON & 60 & 77.22 (+2.22) & 70.19 (+1.20) & 72.50 (+3.24) & 58.06 (+3.80) & 44.61 (-0.51) & 64.52 (+1.99) & {[}+1.36, +2.49{]} \\
  \rowcolor{bestrow}ON & OFF & ON & 80 & 78.15 (+3.15) & 70.93 (+1.94) & 70.83 (+1.57) & 57.22 (+2.96) & 46.13 (+1.01) & 64.65 (+2.13) & {[}+1.47, +2.61{]} \\
  ON & OFF & ON & 100 & 76.11 (+1.11) & 70.56 (+1.57) & 70.46 (+1.20) & 59.91 (+5.65) & 45.88 (+0.76) & 64.58 (+2.06) & {[}+1.14, +2.98{]} \\
  \midrule[0.3pt]
  ON & ON & OFF & 20 & 26.76 (-0.46) & 18.33 (-2.31) & 15.65 (-1.67) & 9.17 (-0.37) & 15.82 (-1.18) & 17.15 (-1.20) & {[}-1.70, -0.64{]} \\
  ON & ON & OFF & 40 & 24.63 (-2.59) & 20.28 (-0.37) & 16.30 (-1.02) & 9.91 (+0.37) & 15.82 (-1.18) & 17.39 (-0.96) & {[}-1.37, -0.44{]} \\
  ON & ON & OFF & 60 & 26.30 (-0.93) & 20.00 (-0.65) & 15.00 (-2.31) & 9.17 (-0.37) & 15.99 (-1.01) & 17.29 (-1.05) & {[}-1.71, -0.39{]} \\
  ON & ON & OFF & 80 & 25.83 (-1.39) & 20.19 (-0.46) & 15.74 (-1.57) & 7.96 (-1.57) & 16.41 (-0.59) & 17.23 (-1.12) & {[}-2.20, -0.16{]} \\
  \rowcolor{bestrow}ON & ON & OFF & 100 & 26.85 (-0.37) & 19.91 (-0.74) & 16.57 (-0.74) & 8.89 (-0.65) & 15.49 (-1.52) & 17.54 (-0.80) & {[}-1.39, -0.21{]} \\
  \midrule[0.3pt]
  \rowcolor{bestrow}ON & ON & ON & 20 & 76.48 (+1.48) & 68.98 (+0.00) & 69.91 (+0.65) & 56.94 (+2.69) & 45.20 (+0.08) & 63.50 (+0.98) & {[}-0.06, +1.90{]} \\
  ON & ON & ON & 40 & 74.63 (-0.37) & 67.59 (-1.39) & 68.43 (-0.83) & 57.87 (+3.61) & 46.21 (+1.09) & 62.95 (+0.42) & {[}-0.32, +1.17{]} \\
  ON & ON & ON & 60 & 72.13 (-2.87) & 65.46 (-3.52) & 65.00 (-4.26) & 54.63 (+0.37) & 42.17 (-2.95) & 59.88 (-2.64) & {[}-3.57, -1.79{]} \\
  ON & ON & ON & 80 & 70.93 (-4.07) & 63.15 (-5.83) & 64.35 (-4.91) & 54.44 (+0.19) & 41.84 (-3.28) & 58.94 (-3.58) & {[}-4.26, -2.99{]} \\
  ON & ON & ON & 100 & 71.30 (-3.70) & 63.52 (-5.46) & 63.24 (-6.02) & 54.35 (+0.09) & 43.69 (-1.43) & 59.22 (-3.30) & {[}-5.07, -1.84{]} \\
  \bottomrule
  \end{tabularx}
  \endgroup
  \vspace{1mm}
  \caption*{\phantomsection\label{app:tab-b1c}\textbf{Table C1c:} \textbf{Qwen3-8B} per-benchmark Avg@12 accuracy at all Figure~\ref{fig:interface-factorial}a checkpoints. Accuracies and parenthetical Base deltas use three-seed means. The final column gives the independent two-sample 95\% seed-bootstrap interval for the macro delta. Shaded rows mark Base (gray) and the step with the highest macro accuracy within each setup (yellow).}
\end{table*}

\subsection{\texorpdfstring{Figure~\ref{fig:interface-factorial}b}{Figure 1b}: privileged-solution versus problem-only teacher}

These tables expand Figure~\ref{fig:interface-factorial}b at Steps 20, 40, 60, 80, and 100 for the problem-only teacher. The privileged-solution arm is the Thinking-ON teacher, Thinking-OFF student rows of Tables~\hyperref[app:tab-b1a]{C1a}--\hyperref[app:tab-b1c]{C1c}. Both teachers are frozen copies of the student. Because the two arms share the same three seed tuples, we also compare their seed-level macro averages directly (Table~\hyperref[app:tab-b3]{C3}).

\begin{table*}[!t]
  \centering
  \begingroup
  \fontsize{6.5pt}{7.5pt}\selectfont
  \setlength{\tabcolsep}{3pt}
  \renewcommand{\arraystretch}{0.95}
  \begin{tabularx}{\textwidth}{@{}ccc*{7}{>{\centering\arraybackslash}X}@{}}
  \toprule
  \textbf{Teacher input} & \textbf{Eval} & \textbf{Step} & \textbf{AIME 2024} & \textbf{AIME 2025} & \textbf{AIME 2026} & \textbf{HMMT Nov. 2025} & \textbf{HMMT Feb. 2026} & \textbf{Macro average} & \textbf{Macro $\Delta$ 95\% CI} \\
  \midrule
  \rowcolor{baserow}Base & OFF & 0 & 13.89 & 8.98 & 9.81 & 3.61 & 7.07 & 8.67 & --- \\
  \rowcolor{baserow}Base & ON & 0 & 47.78 & 34.44 & 40.28 & 29.35 & 29.46 & 36.26 & --- \\
  \midrule[0.3pt]
  Problem only & OFF & 20 & 8.43 (-5.46) & 6.67 (-2.31) & 7.87 (-1.94) & 3.52 (-0.09) & 4.55 (-2.53) & 6.21 (-2.47) & {[}-2.59, -2.34{]} \\
  Problem only & OFF & 40 & 8.98 (-4.91) & 6.20 (-2.78) & 6.30 (-3.52) & 3.24 (-0.37) & 4.97 (-2.10) & 5.94 (-2.74) & {[}-3.12, -2.16{]} \\
  Problem only & OFF & 60 & 12.78 (-1.11) & 9.81 (+0.83) & 8.80 (-1.02) & 4.17 (+0.56) & 8.84 (+1.77) & 8.88 (+0.21) & {[}-0.44, +0.66{]} \\
  Problem only & OFF & 80 & 14.17 (+0.28) & 10.74 (+1.76) & 8.52 (-1.30) & 4.35 (+0.74) & 9.01 (+1.94) & 9.36 (+0.68) & {[}+0.08, +1.38{]} \\
  \rowcolor{bestrow}Problem only & OFF & 100 & 14.07 (+0.19) & 10.56 (+1.57) & 9.91 (+0.09) & 4.63 (+1.02) & 10.19 (+3.11) & 9.87 (+1.20) & {[}+0.21, +1.75{]} \\
  \midrule[0.3pt]
  Problem only & ON & 20 & 51.20 (+3.43) & 37.50 (+3.06) & 39.54 (-0.74) & 28.98 (-0.37) & 30.22 (+0.76) & 37.49 (+1.23) & {[}+0.67, +1.78{]} \\
  Problem only & ON & 40 & 51.02 (+3.24) & 40.46 (+6.02) & 42.13 (+1.85) & 29.63 (+0.28) & 29.38 (-0.08) & 38.52 (+2.26) & {[}+1.71, +2.81{]} \\
  \rowcolor{bestrow}Problem only & ON & 60 & 52.59 (+4.81) & 39.44 (+5.00) & 45.00 (+4.72) & 31.02 (+1.67) & 30.30 (+0.84) & 39.67 (+3.41) & {[}+2.61, +4.12{]} \\
  Problem only & ON & 80 & 51.57 (+3.80) & 39.63 (+5.19) & 43.98 (+3.70) & 27.96 (-1.39) & 30.89 (+1.43) & 38.81 (+2.55) & {[}+1.21, +4.14{]} \\
  Problem only & ON & 100 & 53.24 (+5.46) & 39.91 (+5.46) & 42.59 (+2.31) & 29.07 (-0.28) & 30.22 (+0.76) & 39.01 (+2.74) & {[}+1.75, +3.94{]} \\
  \bottomrule
  \end{tabularx}
  \endgroup
  \vspace{1mm}
  \caption*{\phantomsection\label{app:tab-b2a}\textbf{Table C2a:} \textbf{Qwen3-1.7B} problem-only-teacher per-benchmark Avg@12 accuracy at all Figure~\ref{fig:interface-factorial}b checkpoints. Accuracies and parenthetical Base deltas use three-seed means. The final column gives the independent two-sample 95\% seed-bootstrap interval for the macro delta. Shaded rows mark Base (gray) and the step with the highest macro accuracy within each setup (yellow).}
\end{table*}
\begin{table*}[!t]
  \centering
  \begingroup
  \fontsize{6.5pt}{7.5pt}\selectfont
  \setlength{\tabcolsep}{3pt}
  \renewcommand{\arraystretch}{0.95}
  \begin{tabularx}{\textwidth}{@{}ccc*{7}{>{\centering\arraybackslash}X}@{}}
  \toprule
  \textbf{Teacher input} & \textbf{Eval} & \textbf{Step} & \textbf{AIME 2024} & \textbf{AIME 2025} & \textbf{AIME 2026} & \textbf{HMMT Nov. 2025} & \textbf{HMMT Feb. 2026} & \textbf{Macro average} & \textbf{Macro $\Delta$ 95\% CI} \\
  \midrule
  \rowcolor{baserow}Base & OFF & 0 & 22.78 & 18.43 & 15.09 & 7.87 & 16.25 & 16.08 & --- \\
  \rowcolor{baserow}Base & ON & 0 & 73.33 & 65.74 & 66.48 & 55.37 & 41.16 & 60.42 & --- \\
  \midrule[0.3pt]
  Problem only & OFF & 20 & 19.44 (-3.33) & 13.52 (-4.91) & 13.33 (-1.76) & 5.28 (-2.59) & 12.88 (-3.37) & 12.89 (-3.19) & {[}-3.65, -2.73{]} \\
  Problem only & OFF & 40 & 22.96 (+0.19) & 18.24 (-0.19) & 16.57 (+1.48) & 8.80 (+0.93) & 15.15 (-1.09) & 16.35 (+0.26) & {[}-0.48, +1.02{]} \\
  Problem only & OFF & 60 & 40.93 (+18.15) & 32.59 (+14.17) & 31.67 (+16.57) & 25.83 (+17.96) & 23.99 (+7.74) & 31.00 (+14.92) & {[}+12.88, +16.62{]} \\
  Problem only & OFF & 80 & 53.52 (+30.74) & 43.06 (+24.63) & 47.78 (+32.69) & 33.24 (+25.37) & 30.13 (+13.89) & 41.55 (+25.46) & {[}+23.93, +26.97{]} \\
  \rowcolor{bestrow}Problem only & OFF & 100 & 56.39 (+33.61) & 48.06 (+29.63) & 48.24 (+33.15) & 35.28 (+27.41) & 31.40 (+15.15) & 43.87 (+27.79) & {[}+22.17, +36.53{]} \\
  \midrule[0.3pt]
  Problem only & ON & 20 & 74.91 (+1.57) & 66.30 (+0.56) & 68.15 (+1.67) & 52.22 (-3.15) & 41.67 (+0.51) & 60.65 (+0.23) & {[}-0.24, +0.68{]} \\
  Problem only & ON & 40 & 75.19 (+1.85) & 67.41 (+1.67) & 68.70 (+2.22) & 52.96 (-2.41) & 41.08 (-0.08) & 61.07 (+0.65) & {[}+0.50, +0.79{]} \\
  Problem only & ON & 60 & 74.07 (+0.74) & 69.91 (+4.17) & 67.50 (+1.02) & 54.72 (-0.65) & 43.18 (+2.02) & 61.88 (+1.46) & {[}+0.98, +1.82{]} \\
  \rowcolor{bestrow}Problem only & ON & 80 & 75.37 (+2.04) & 69.91 (+4.17) & 67.13 (+0.65) & 55.00 (-0.37) & 43.52 (+2.36) & 62.19 (+1.77) & {[}+1.41, +2.29{]} \\
  Problem only & ON & 100 & 73.33 (+0.00) & 68.61 (+2.87) & 66.76 (+0.28) & 53.33 (-2.04) & 42.76 (+1.60) & 60.96 (+0.54) & {[}+0.22, +0.87{]} \\
  \bottomrule
  \end{tabularx}
  \endgroup
  \vspace{1mm}
  \caption*{\phantomsection\label{app:tab-b2b}\textbf{Table C2b:} \textbf{Qwen3-4B} problem-only-teacher per-benchmark Avg@12 accuracy at all Figure~\ref{fig:interface-factorial}b checkpoints. Accuracies and parenthetical Base deltas use three-seed means. The final column gives the independent two-sample 95\% seed-bootstrap interval for the macro delta. Shaded rows mark Base (gray) and the step with the highest macro accuracy within each setup (yellow).}
\end{table*}
\begin{table*}[!t]
  \centering
  \begingroup
  \fontsize{6.5pt}{7.5pt}\selectfont
  \setlength{\tabcolsep}{3pt}
  \renewcommand{\arraystretch}{0.95}
  \begin{tabularx}{\textwidth}{@{}ccc*{7}{>{\centering\arraybackslash}X}@{}}
  \toprule
  \textbf{Teacher input} & \textbf{Eval} & \textbf{Step} & \textbf{AIME 2024} & \textbf{AIME 2025} & \textbf{AIME 2026} & \textbf{HMMT Nov. 2025} & \textbf{HMMT Feb. 2026} & \textbf{Macro average} & \textbf{Macro $\Delta$ 95\% CI} \\
  \midrule
  \rowcolor{baserow}Base & OFF & 0 & 27.22 & 20.65 & 17.31 & 9.54 & 17.00 & 18.35 & --- \\
  \rowcolor{baserow}Base & ON & 0 & 75.00 & 68.98 & 69.26 & 54.26 & 45.12 & 62.52 & --- \\
  \midrule[0.3pt]
  Problem only & OFF & 20 & 22.41 (-4.81) & 15.65 (-5.00) & 13.89 (-3.43) & 5.93 (-3.61) & 14.14 (-2.86) & 14.40 (-3.94) & {[}-4.49, -3.38{]} \\
  Problem only & OFF & 40 & 26.02 (-1.20) & 22.59 (+1.94) & 20.28 (+2.96) & 9.44 (-0.09) & 17.85 (+0.84) & 19.24 (+0.89) & {[}+0.18, +1.66{]} \\
  Problem only & OFF & 60 & 41.94 (+14.72) & 32.50 (+11.85) & 30.83 (+13.52) & 23.70 (+14.17) & 24.83 (+7.83) & 30.76 (+12.42) & {[}+10.01, +13.95{]} \\
  Problem only & OFF & 80 & 57.96 (+30.74) & 43.24 (+22.59) & 48.70 (+31.39) & 38.43 (+28.89) & 32.07 (+15.07) & 44.08 (+25.74) & {[}+23.30, +27.84{]} \\
  \rowcolor{bestrow}Problem only & OFF & 100 & 58.24 (+31.02) & 43.89 (+23.24) & 49.54 (+32.22) & 39.26 (+29.72) & 32.41 (+15.40) & 44.67 (+26.32) & {[}+25.17, +27.74{]} \\
  \midrule[0.3pt]
  Problem only & ON & 20 & 76.76 (+1.76) & 69.07 (+0.09) & 69.44 (+0.19) & 59.26 (+5.00) & 45.96 (+0.84) & 64.10 (+1.58) & {[}+0.87, +2.28{]} \\
  Problem only & ON & 40 & 77.31 (+2.31) & 68.98 (+0.00) & 69.44 (+0.19) & 58.43 (+4.17) & 45.88 (+0.76) & 64.01 (+1.48) & {[}+0.81, +2.11{]} \\
  Problem only & ON & 60 & 77.41 (+2.41) & 71.02 (+2.04) & 70.56 (+1.30) & 58.80 (+4.54) & 46.46 (+1.35) & 64.85 (+2.32) & {[}+1.59, +3.06{]} \\
  \rowcolor{bestrow}Problem only & ON & 80 & 76.48 (+1.48) & 71.76 (+2.78) & 70.56 (+1.30) & 59.17 (+4.91) & 46.55 (+1.43) & 64.90 (+2.38) & {[}+1.68, +3.08{]} \\
  Problem only & ON & 100 & 76.02 (+1.02) & 72.31 (+3.33) & 68.98 (-0.28) & 57.69 (+3.43) & 47.47 (+2.36) & 64.49 (+1.97) & {[}+1.29, +2.59{]} \\
  \bottomrule
  \end{tabularx}
  \endgroup
  \vspace{1mm}
  \caption*{\phantomsection\label{app:tab-b2c}\textbf{Table C2c:} \textbf{Qwen3-8B} problem-only-teacher per-benchmark Avg@12 accuracy at all Figure~\ref{fig:interface-factorial}b checkpoints. Accuracies and parenthetical Base deltas use three-seed means. The final column gives the independent two-sample 95\% seed-bootstrap interval for the macro delta. Shaded rows mark Base (gray) and the step with the highest macro accuracy within each setup (yellow).}
\end{table*}

\begin{table*}[!t]
  \centering
  \begingroup
  \scriptsize
  \setlength{\tabcolsep}{6pt}
  \renewcommand{\arraystretch}{1.0}
  \begin{tabular}{@{}ccccc@{}}
  \toprule
  \textbf{Model} & \textbf{Eval} & \textbf{Step} & \textbf{Problem-only $-$ privileged (pp)} & \textbf{95\% paired seed-bootstrap CI} \\
  \midrule
  \rowcolor{pairedrow}Qwen3-1.7B & OFF & 20 & -1.24 & {[}-2.28, -0.06{]} \\
  \rowcolor{pairedrow}Qwen3-1.7B & OFF & 40 & -1.13 & {[}-1.41, -0.84{]} \\
  Qwen3-1.7B & OFF & 60 & -0.25 & {[}-0.51, +0.10{]} \\
  \rowcolor{bestrow}Qwen3-1.7B & OFF & 80 & +0.91 & {[}+0.04, +2.02{]} \\
  Qwen3-1.7B & OFF & 100 & +1.36 & {[}-0.33, +2.93{]} \\
  \midrule[0.3pt]
  \rowcolor{pairedrow}Qwen3-1.7B & ON & 20 & -1.09 & {[}-1.78, -0.58{]} \\
  \rowcolor{pairedrow}Qwen3-1.7B & ON & 40 & -1.14 & {[}-1.84, -0.65{]} \\
  \rowcolor{pairedrow}Qwen3-1.7B & ON & 60 & -2.06 & {[}-4.03, -0.87{]} \\
  \rowcolor{pairedrow}Qwen3-1.7B & ON & 80 & -2.02 & {[}-3.36, -0.10{]} \\
  \rowcolor{pairedrow}Qwen3-1.7B & ON & 100 & -2.59 & {[}-3.54, -1.19{]} \\
  \midrule[0.3pt]
  \rowcolor{pairedrow}Qwen3-4B & OFF & 20 & -1.91 & {[}-2.22, -1.74{]} \\
  Qwen3-4B & OFF & 40 & -0.28 & {[}-1.09, +0.27{]} \\
  \rowcolor{bestrow}Qwen3-4B & OFF & 60 & +9.11 & {[}+6.77, +11.29{]} \\
  \rowcolor{bestrow}Qwen3-4B & OFF & 80 & +14.41 & {[}+12.85, +15.77{]} \\
  \rowcolor{bestrow}Qwen3-4B & OFF & 100 & +15.21 & {[}+10.88, +22.82{]} \\
  \midrule[0.3pt]
  Qwen3-4B & ON & 20 & +0.08 & {[}-0.36, +0.52{]} \\
  Qwen3-4B & ON & 40 & +0.13 & {[}-0.73, +0.61{]} \\
  Qwen3-4B & ON & 60 & +0.47 & {[}-1.03, +2.06{]} \\
  \rowcolor{bestrow}Qwen3-4B & ON & 80 & +1.21 & {[}+0.53, +1.58{]} \\
  Qwen3-4B & ON & 100 & +0.71 & {[}-0.26, +1.38{]} \\
  \midrule[0.3pt]
  \rowcolor{pairedrow}Qwen3-8B & OFF & 20 & -1.02 & {[}-1.98, -0.51{]} \\
  \rowcolor{bestrow}Qwen3-8B & OFF & 40 & +1.47 & {[}+0.91, +2.02{]} \\
  \rowcolor{bestrow}Qwen3-8B & OFF & 60 & +8.95 & {[}+7.49, +10.75{]} \\
  \rowcolor{bestrow}Qwen3-8B & OFF & 80 & +17.68 & {[}+10.56, +21.53{]} \\
  \rowcolor{bestrow}Qwen3-8B & OFF & 100 & +8.60 & {[}+0.90, +13.30{]} \\
  \midrule[0.3pt]
  Qwen3-8B & ON & 20 & +0.05 & {[}-0.74, +0.70{]} \\
  \rowcolor{pairedrow}Qwen3-8B & ON & 40 & -0.26 & {[}-0.32, -0.16{]} \\
  \rowcolor{bestrow}Qwen3-8B & ON & 60 & +0.33 & {[}+0.02, +0.94{]} \\
  Qwen3-8B & ON & 80 & +0.25 & {[}-0.03, +0.62{]} \\
  Qwen3-8B & ON & 100 & -0.09 & {[}-0.62, +0.72{]} \\
  \bottomrule
  \end{tabular}
  \endgroup
  \vspace{1mm}
  \caption*{\phantomsection\label{app:tab-b3}\textbf{Table C3:} Direct matched macro contrast between the Figure~\ref{fig:interface-factorial}b teacher inputs at every checkpoint. Positive values favor the problem-only teacher. Intervals use 50,000 paired bootstrap resamples of the three training-seed differences. Shaded rows mark intervals that exclude zero, favoring the problem-only teacher (yellow) or the privileged-solution teacher (blue). With three training seeds per arm, these intervals are descriptive.}
\end{table*}

\section{RQ1: Source}
\label{app:source}

Appendix~\ref{app:source-ot} reports the OpenThoughts recovery, recognition, and response audits behind \S\ref{sec:source-recovery}-\S\ref{sec:source-recognition}. Appendix~\ref{app:source-hm} reports the Hendrycks MATH Level 5 hard-problem solving analysis behind \S\ref{sec:source-solving}, together with cross-dataset recovery and recognition replications. Dataset construction appears in Appendix~\ref{app:data}.

\begin{table*}[!t]
  \centering
  \scriptsize
  \setlength{\tabcolsep}{2pt}
  \renewcommand{\arraystretch}{1.0}
  \begin{tabular*}{\textwidth}{@{\extracolsep{\fill}}ccccccc@{}}
  \toprule
  \textbf{Corpus} & \textbf{Model} & \textbf{Screened} & \textbf{Audited source} & \textbf{Mixed-outcome / candidate} & \textbf{Consensus eligible} & \textbf{Selected} \\
  \midrule
  OpenThoughts & Qwen3-1.7B & 3,200 & --- & 646 & 296 & 50 \\
  OpenThoughts & Qwen3-4B & 3,200 & --- & 183 & 113 & 50 \\
  OpenThoughts & Qwen3-8B & 3,200 & --- & 164 & 78 & 50 \\
  Hendrycks MATH Level 5 & Qwen3-1.7B & 2,242 & 111 & 180 & 149 & 50 \\
  Hendrycks MATH Level 5 & Qwen3-4B & 2,242 & 44 & 80 & 69 & 50 \\
  Hendrycks MATH Level 5 & Qwen3-8B & 2,242 & 35 & 79 & 53 & 50 \\
  \bottomrule
  \end{tabular*}
  \vspace{1mm}
  \caption*{\phantomsection\label{app:tab-d1b}\textbf{Table D1b:} Cohort construction and error-audit attrition. Audited-source counts apply only to the Hendrycks MATH Level 5 original-question analysis. The selected recovery panels contain 50 pairs per scale. OpenThoughts recognition uses 200 pair states per scale (Appendix~\ref{app:source-ot-recognition}).}
\end{table*}

\subsection{OpenThoughts: recovery and recognition}
\label{app:source-ot}

This section gives the protocols and complete results for recovery after errors (\S\ref{sec:source-recovery}) and error recognition (\S\ref{sec:source-recognition}).

\subsubsection{Cohort construction}

For each model scale, we draw 12 reference-free Thinking-ON samples per OpenThoughts problem before any source-conditioned generation. The RQ1 mixed-outcome pool contains problems with 1--11/12 correct samples, yielding 646, 183, and 164 problems at Qwen3-1.7B, 4B, and 8B. This broader pool differs from the 1--5/12 Frontier bucket used in RQ3. It is frozen before error-boundary selection, and screening answers use the canonical evaluator.

\subsubsection{Recovery after errors}
\label{app:source-ot-recovery}

For each mixed-outcome problem, we select one correct and one incorrect non-truncated trajectory while minimizing their token-length difference and randomize display order. GPT-5.6-Sol and Claude Opus 4.8 independently mark the earliest 128-token chunk containing a specific recoverable mathematical error without seeing terminal outcomes. Incompleteness, exploration, and a different valid strategy do not count as errors. Consensus produces 296, 113, and 78 eligible pairs; 50 per scale are selected across four bins for the number of tokens before the error (0--1,023, 1,024--2,047, 2,048--4,095, and at least 4,096) before any Qwen condition is evaluated.

Recovery begins from a fixed prefix ending after both judges' marked error chunks and draws four matched continuations under problem-only, shuffled-reference, and privileged-solution conditions using four prespecified decoding streams, with the frozen teacher in Thinking ON at temperature 1.0. The shuffled-reference condition supplies a worked solution to a different problem in place of the privileged solution. GPT-5.6-Sol and Claude Opus 4.8 independently assign task-level valid-solution labels to all 1,800 continuations, with a blinded pass adjudicating disagreements. A continuation must repair or abandon the inherited error and reach the requested result. Ambiguous responses count as incorrect. Confidence intervals resample source problems while preserving the matched continuations.

\begin{table*}[!t]
  \centering
  \scriptsize
  \setlength{\tabcolsep}{2pt}
  \renewcommand{\arraystretch}{1.0}
  \begin{tabular*}{\textwidth}{@{\extracolsep{\fill}}cccccccc@{}}
  \toprule
  \textbf{Model} & \textbf{Prefixes / clusters} & \textbf{Problem only} & \textbf{Shuffled} & \textbf{Privileged} & \textbf{Privileged $-$ problem} & \textbf{Privileged $-$ shuffled} & \textbf{Trunc. P/S/V} \\
  \midrule
  Qwen3-1.7B & 50 / 46 & 48.50\% & 41.50\% & 74.50\% & +26.00 {[}+16.35, +36.06{]} & +33.00 {[}+21.50, +44.61{]} & 0.0/0.5/0.0\% \\
  Qwen3-4B & 50 / 46 & 66.00\% & 51.00\% & 88.50\% & +22.50 {[}+11.27, +34.31{]} & +37.50 {[}+24.53, +50.00{]} & 0.0/0.0/0.0\% \\
  Qwen3-8B & 50 / 46 & 63.00\% & 53.50\% & 88.50\% & +25.50 {[}+15.87, +36.00{]} & +35.00 {[}+23.21, +47.34{]} & 0.0/0.0/0.0\% \\
  \bottomrule
  \end{tabular*}
  \vspace{1mm}
  \caption*{\phantomsection\label{app:tab-d2}\textbf{Table D2:} OpenThoughts task-level recovery from 50 prefixes containing verified errors per scale. Each arm has four continuations per prefix. Brackets are paired 95\% source-problem-cluster bootstrap intervals. P/S/V denotes problem-only/shuffled/privileged.}
\end{table*}

\vspace{1mm}
\noindent\textbf{Privileged solutions support recovery after verified errors.} Privileged-solution recovery reaches 74.50--88.50\% and exceeds shuffled references by 33.00--37.50pp. Every interval excludes zero. Task-level grading rejects invalid proofs and continuations that do not repair the prefix, yet the privileged-context advantage remains large.

\subsubsection{Error recognition}
\label{app:source-ot-recognition}

We construct 200 audited prefix pairs per scale. Each contains two prefixes from the same trajectory: a valid prefix ending before the earlier judge mark and an erroneous prefix ending after the later mark. These panels span 200, 172, and 156 unique source questions at 1.7B, 4B, and 8B, respectively. Repeated states are kept together in uncertainty estimates. The frozen teacher judges whether each prefix is correct so far, with and without the privileged solution, using deterministic Thinking-ON decoding at temperature 0. It is instructed not to solve the full problem and to end with exactly one line, either ``\texttt{VERDICT: ERROR}'' or ``\texttt{VERDICT: NO\_ERROR\_VISIBLE}''. \texttt{ERROR} requires a substantive mathematical error, invalid inference, or false claim already visible in the prefix. A response parses only when a closed thinking block is followed by exactly one prescribed verdict. A missing close, missing verdict, or multiple verdicts counts as incorrect, and parse rate is reported separately. Accuracy on valid prefixes (specificity) measures correct acceptance. Accuracy on erroneous prefixes (recall) measures correct flagging. Overall prefix accuracy (balanced accuracy) averages the two, and pair accuracy (joint accuracy) requires both judgments in a pair to be correct. All metrics use the same 200 pairs. An always-accept baseline has 100\% accuracy on valid prefixes, 0\% on erroneous prefixes, 50\% overall prefix accuracy, and 0\% pair accuracy. The response cap is min(38,912, 40,960 minus rendered input tokens).

\begin{table*}[!t]
  \centering
  \scriptsize
  \setlength{\tabcolsep}{2pt}
  \renewcommand{\arraystretch}{1.0}
  \begin{tabular*}{\textwidth}{@{\extracolsep{\fill}}ccccccccc@{}}
  \toprule
  \textbf{Model} & \textbf{Pairs} & \textbf{Valid N/R} & \textbf{Error N/R} & \textbf{Overall N/R} & \textbf{Pair N/R} & \textbf{Pair R $-$ N} & \textbf{Parse N/R} & \textbf{Trunc. N/R} \\
  \midrule
  Qwen3-1.7B & 200 & 56.0/65.0\% & 23.0/25.5\% & 39.5/45.2\% & 9.5/13.0\% & +3.50 {[}-2.00, +9.50{]} & 78.8/78.8\% & 3.0/0.8\% \\
  Qwen3-4B & 200 & 78.5/82.0\% & 14.5/13.0\% & 46.5/47.5\% & 9.0/6.5\% & -2.50 {[}-7.50, +2.50{]} & 89.0/87.0\% & 2.5/5.5\% \\
  Qwen3-8B & 200 & 86.0/89.0\% & 18.5/16.5\% & 52.2/52.8\% & 14.0/12.5\% & -1.50 {[}-6.50, +3.50{]} & 94.0/97.0\% & 0.5/0.2\% \\
  \bottomrule
  \end{tabular*}
  \vspace{1mm}
  \caption*{\phantomsection\label{app:tab-d3}\textbf{Table D3:} OpenThoughts explicit error recognition. Valid and Error report accuracy on valid and erroneous prefixes. N/R denotes problem only/privileged solution. Pair R $-$ N is in pp. Intervals use 50,000 paired source-question-cluster bootstrap resamples, preserving repeated pair states from the same source question.}
\end{table*}

\vspace{1mm}
\noindent\textbf{Privileged solutions do not reliably improve recognition.} Accuracy on erroneous prefixes remains only 13.0--25.5\%. Privileged context raises overall prefix accuracy by +5.75pp {[}+0.50, +11.00{]} at 1.7B, +1.00pp {[}-3.25, +5.25{]} at 4B, and +0.50pp {[}-3.00, +4.25{]} at 8B. Intervals for pair accuracy include zero at every scale. Parse rates of 78.8--97.0\% and truncation rates of 0.2--5.5\% rule out the response cap as the main explanation. Unparsed responses count as incorrect, and parse rates differ by at most 3pp between conditions within each scale.

\subsubsection{Blind response audit}

We apply an outcome-blind response audit to OpenThoughts. Appendix~\ref{app:source-hm-audit} applies the same protocol to Hendrycks MATH Level 5. Each teacher mode contains 225 response pairs: 150 automatically identified problem-only-wrong to privileged-correct pairs (50 per scale) and 75 wrong-to-wrong controls (25 per scale). Candidate order is randomized. GPT-5.6-Sol sees the problem and worked solution as mathematical evidence but not the generation condition or automatic-verifier outcome. We report mechanism labels only when the judge independently confirms that the problem-only response is incorrect and the privileged response is correct.

The mechanism label identifies the dominant change in the successful response: a local repair preserves the original route, global strategy selection replaces it, verification or final-answer correction changes only the terminal check or answer, close reproduction follows distinctive reference steps, and search pruning or supplied fact captures other direct reductions in search. Deployment is labeled self-contained when the successful response justifies its reasoning on its own and privilege-dependent/direct when distinctive imported steps are not independently supported.

\begin{table*}[!t]
  \centering
  \scriptsize
  \setlength{\tabcolsep}{6pt}
  \renewcommand{\arraystretch}{1.0}

  \begin{tabular}{@{}cccc@{}}
  \toprule
  \textbf{Audit dimension} & \textbf{Label} & \textbf{Thinking ON (n = 40)} & \textbf{Thinking OFF (n = 66)} \\
  \midrule
  Mechanism & Local repair & 14 (35.0\%) & 22 (33.3\%) \\
  Mechanism & Global strategy selection & 9 (22.5\%) & 19 (28.8\%) \\
  Mechanism & Verification or final-answer correction & 13 (32.5\%) & 8 (12.1\%) \\
  Mechanism & Close reference reproduction & 4 (10.0\%) & 8 (12.1\%) \\
  Mechanism & Search pruning or supplied fact & 0 (0.0\%) & 9 (13.6\%) \\
  Deployment & Self-contained derivation & 35 (87.5\%) & 58 (87.9\%) \\
  Deployment & Privilege-dependent or direct & 5 (12.5\%) & 8 (12.1\%) \\
  \bottomrule
  \end{tabular}
  \vspace{1mm}
  \caption*{\phantomsection\label{app:tab-d5}\textbf{Table D5:} Mechanism and deployment labels among blind-confirmed OpenThoughts improvements. Categories within each audit dimension are mutually exclusive.}
\end{table*}

\vspace{1mm}
\noindent\textbf{OpenThoughts improvements are usually not copies of the privileged solution.} Local repair and strategy selection together account for 23/40 (57.5\%) confirmed improvements under Thinking ON and 41/66 (62.1\%) under Thinking OFF. Close reproduction accounts for only 10.0\% and 12.1\%. The successful response is judged self-contained in 87.5\% and 87.9\% of cases. The Hendrycks MATH Level 5 audit uses a slightly different label set, so we compare only the shared close-reproduction and self-contained labels: on that corpus, close reproduction is the largest category and roughly half of improvements remain privilege-dependent or direct (Appendix~\ref{app:source-hm-audit}). Because the blind judge confirms only 40/150 and 66/150 automatically selected wrong-to-right pairs, these proportions are descriptive sensitivity evidence. The matched outcome contrasts in main-text Tables~\ref{tab:source-solving}-\ref{tab:source-recovery} remain the primary evidence.

\begin{figure*}[!t]
  \centering
  \includegraphics[width=0.9\textwidth]{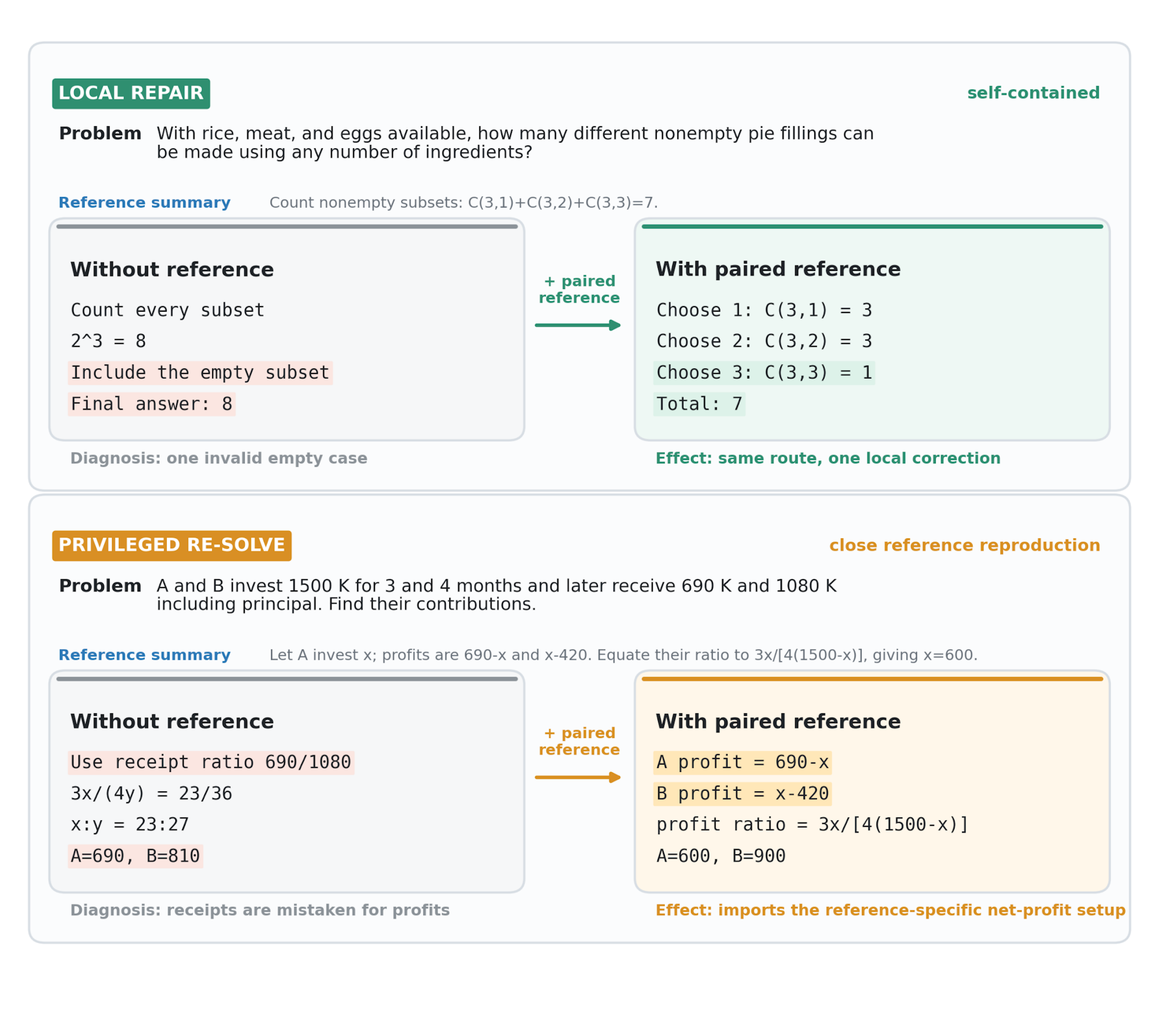}
  \vspace{-10mm}
  \caption*{\phantomsection\label{app:fig-d2}\textbf{Figure D2}: Two OpenThoughts examples from the blind response audit under Teacher Thinking OFF. The top shows a self-contained local repair that removes one invalid case. The bottom shows a privilege-dependent re-solve that imports the reference-specific net-profit setup. Examples are condensed for legibility.}
\end{figure*}

\subsection{Hendrycks MATH Level 5: solving and replications}
\label{app:source-hm}

We use Hendrycks MATH Level 5 for the hard-problem solving analysis in \S\ref{sec:source-solving} and for cross-dataset replications of recovery and recognition. The section reports high-precision cohort construction, hard-problem solving, recovery after errors, explicit error recognition, and a blind response-transition audit.

\subsubsection{Cohort construction}
\label{app:source-hm-cohort}

We start from the cleaned 2,242-problem Hendrycks MATH Level 5 corpus described in Appendix~\ref{app:data-hm}. For each frozen Qwen3 checkpoint, 12 reference-free Thinking-ON screening samples are generated per problem at temperature 1.0, top-p 1.0, top-k disabled, and seed 0. An answer-equivalence screen identifies 125, 62, and 53 candidate 0/12 questions at Qwen3-1.7B, 4B, and 8B. To obtain a high-precision task-level cohort, GPT-5.6-Sol and Claude Opus 4.8 independently judge all 2,880 candidate responses against the problem and verified solution. They agree on 98.65\% of responses. Every disagreement is resolved by a separate blinded adjudication pass. We retain a question only when all 12 responses are judged incorrect, yielding 111, 44, and 35 questions. The cohorts are locked before any fresh source-condition outcome is inspected.

For the error-boundary analyses, we begin from model-relative mixed-outcome problems with both correct and incorrect screening trajectories. A deterministic priority rule selects one correct and one incorrect candidate per problem while favoring non-truncated, closed, shorter responses. This yields 180, 80, and 79 candidate pairs. GPT-5.6-Sol and Claude Opus 4.8 independently audit randomized candidates in 128-token chunks; 149, 69, and 53 pairs satisfy the consensus criteria. Before any Qwen recovery or recognition output is examined, 50 pairs per scale are selected deterministically while preserving subject and error-position coverage. Table~\hyperref[app:tab-d1b]{D1b} summarizes this attrition.

\subsubsection{Hard-problem solving}

On each audited 0/12 cohort, the frozen teacher produces four fresh Thinking-ON samples under problem-only, subject- and length-matched shuffled-reference, and privileged-solution input at temperature 1.0, top-p 1.0, and top-k disabled. Each shuffled donor has the same subject, comes from a different question with a different gold answer, and minimizes reference-token-length difference without donor reuse within scale. Remaining ties use the lower source index. GPT-5.6-Sol and Claude Opus 4.8 independently assign task-level valid-solution labels to all 2,280 fresh responses against the problem and verified solution. The rubric accepts equivalent answer forms but, for proof or construction tasks, requires a valid derivation rather than a terminal-answer match alone. The judges agree on 88.55\% of responses, and a separate blinded pass resolves every disagreement. All responses are present and remain within the native context window.

\begin{table*}[!t]
  \centering
  \scriptsize
  \setlength{\tabcolsep}{2pt}
  \renewcommand{\arraystretch}{1.0}
  \begin{tabular*}{\textwidth}{@{\extracolsep{\fill}}ccccccc@{}}
  \toprule
  \textbf{Model} & \textbf{n} & \textbf{Problem only} & \textbf{Shuffled} & \textbf{Privileged} & \textbf{Privileged $-$ problem} & \textbf{Privileged $-$ shuffled} \\
  \midrule
  Qwen3-1.7B & 111 & 2.93\% & 1.13\% & 40.54\% & +37.61 {[}+31.31, +44.14{]} & +39.41 {[}+33.11, +45.95{]} \\
  Qwen3-4B & 44 & 1.14\% & 1.70\% & 64.20\% & +63.07 {[}+51.14, +74.43{]} & +62.50 {[}+50.57, +73.86{]} \\
  Qwen3-8B & 35 & 0.00\% & 1.43\% & 58.57\% & +58.57 {[}+46.43, +70.71{]} & +57.14 {[}+45.71, +68.57{]} \\
  \bottomrule
  \end{tabular*}
  \vspace{1mm}
  \caption*{\phantomsection\label{app:tab-d6}\textbf{Table D6:} Hendrycks MATH Level 5 task-level valid-solution accuracy on independently audited 0/12 cohorts. Each condition uses four fresh samples per problem. Brackets are paired 95\% question-bootstrap confidence intervals.}
\end{table*}

\vspace{1mm}
\noindent\textbf{Privileged solutions provide strong, problem-specific source capability.} Privileged-solution input sharply raises valid-solution accuracy (40.54--64.20\%), while problem-only and shuffled-reference accuracy remain at or below 2.93\%. Every privileged-solution contrast excludes zero, showing that the gain depends on information matched to the current problem rather than solution-like context alone.

\subsubsection{Recovery after errors}

Using the 50 consensus pairs per scale selected in Appendix~\ref{app:source-hm-cohort}, recovery begins from a fixed prefix ending after both judges' marked error chunks. The frozen teacher generates four Thinking-ON continuations, one per prespecified decoding stream, under problem-only, subject- and length-matched shuffled-reference, and privileged-solution conditions, with temperature 1.0, top-p 1.0, and top-k disabled. The primary privileged-minus-shuffled comparison holds the prefix, prompt wrapper, and decoding seed fixed while changing whether the supplied solution matches the problem. Unlike the OpenThoughts recovery audit, which uses task-level judges, final answers here are scored with the canonical exact-answer evaluator, so Hendrycks MATH Level 5 recovery rates are not directly comparable to Table~\hyperref[app:tab-d2]{D2}. Intervals resample source-problem clusters, and every selected sequence respects the native context window.

\begin{table*}[!t]
  \centering
  \scriptsize
  \setlength{\tabcolsep}{2pt}
  \renewcommand{\arraystretch}{1.0}
  \begin{tabular*}{\textwidth}{@{\extracolsep{\fill}}cccccccc@{}}
  \toprule
  \textbf{Model} & \textbf{Prefixes / clusters} & \textbf{Problem only} & \textbf{Shuffled} & \textbf{Privileged} & \textbf{Privileged $-$ problem} & \textbf{Privileged $-$ shuffled} & \textbf{Trunc. P/S/V} \\
  \midrule
  Qwen3-1.7B & 50 / 46 & 26.00\% & 22.00\% & 92.50\% & +66.50 {[}+55.29, +76.96{]} & +70.50 {[}+60.11, +80.10{]} & 0.5/0.0/0.0\% \\
  Qwen3-4B & 50 / 41 & 39.00\% & 27.50\% & 97.50\% & +58.50 {[}+46.76, +69.68{]} & +70.00 {[}+57.87, +80.80{]} & 1.0/0.0/0.0\% \\
  Qwen3-8B & 50 / 41 & 37.00\% & 36.50\% & 96.00\% & +59.00 {[}+47.92, +69.39{]} & +59.50 {[}+47.73, +70.28{]} & 1.0/0.5/0.0\% \\
  \bottomrule
  \end{tabular*}
  \vspace{1mm}
  \caption*{\phantomsection\label{app:tab-d7}\textbf{Table D7:} Hendrycks MATH Level 5 recovery after errors. Each of 50 prefixes containing an audited error has four matched continuations per arm. Contrasts in pp show 95\% source-problem-cluster bootstrap intervals from 50,000 resamples. P/S/V denotes problem-only/shuffled/privileged.}
\end{table*}

\vspace{1mm}
\noindent\textbf{Task-matched context remains useful after a verified error.} The privileged-minus-shuffled contrasts in Table~\hyperref[app:tab-d7]{D7} reach 70.50, 70.00, and 59.50pp across the three scales. Shuffled references do not improve over problem-only continuations, ruling out extra solution-like text and the teacher wrapper as sufficient explanations.

\subsubsection{Error recognition}

Error boundaries are fixed before the Qwen3 judgments. GPT-5.6-Sol and Claude Opus 4.8 independently inspect randomized candidate trajectories with the problem and worked solution as mathematical evidence but without terminal outcomes. Each labels the reasoning as valid so far, containing a recoverable mathematical error, or ambiguous. A recoverable error must be a visible false claim or invalid inference that could still be retracted. Incompleteness, exploration, and a different valid strategy do not count. Each judge identifies the earliest 128-token error chunk and provides a quotation and explanation. We retain only consensus errors.

We select 50 trajectories per scale, stratified by subject and error position. From each trajectory, we form a valid prefix ending before the earlier annotated error chunk and an erroneous prefix ending after the later chunk. The frozen teacher judges each prefix with and without the privileged solution using deterministic Thinking-ON decoding at temperature 0. It must end with exactly one line, either ``\texttt{VERDICT: ERROR}'' or ``\texttt{VERDICT: NO\_ERROR\_VISIBLE}''. Accuracy measures correct acceptance of valid prefixes and correct flagging of erroneous prefixes. Pair accuracy requires both judgments to be correct. The response cap is min(38,912, 40,960 minus rendered input tokens), and confidence intervals resample matched problem pairs.

\begin{table*}[!t]
  \centering
  \scriptsize
  \setlength{\tabcolsep}{2pt}
  \renewcommand{\arraystretch}{1.0}
  \begin{tabular*}{\textwidth}{@{\extracolsep{\fill}}ccccccccc@{}}
  \toprule
  \textbf{Model} & \textbf{Pairs} & \textbf{Valid N/R} & \textbf{Error N/R} & \textbf{Overall N/R} & \textbf{Pair N/R} & \textbf{Pair R $-$ N} & \textbf{Parse N/R} & \textbf{Trunc. N/R} \\
  \midrule
  Qwen3-1.7B & 50 & 42/52\% & 40/42\% & 41/47\% & 12/22\% & +10.00 {[}-4.00, +24.00{]} & 83/84\% & 2/2\% \\
  Qwen3-4B & 50 & 72/66\% & 30/38\% & 51/52\% & 18/16\% & -2.00 {[}-14.00, +10.00{]} & 92/97\% & 2/1\% \\
  Qwen3-8B & 50 & 72/76\% & 40/20\% & 56/48\% & 24/12\% & -12.00 {[}-26.00, +2.00{]} & 99/98\% & 0/1\% \\
  \bottomrule
  \end{tabular*}
  \vspace{1mm}
  \caption*{\phantomsection\label{app:tab-d8}\textbf{Table D8:} Hendrycks MATH Level 5 explicit error recognition. Valid and Error report accuracy on valid and erroneous prefixes. N/R denotes problem only/privileged solution. Pair R $-$ N is in pp. Intervals for pair accuracy use 20,000 paired problem-bootstrap resamples.}
\end{table*}

\vspace{1mm}
\noindent\textbf{We find no evidence that privileged context improves recognition on Hendrycks MATH Level 5.} In Table~\hyperref[app:tab-d8]{D8}, pair accuracy changes by +10, -2, and -12pp across scales, and every confidence interval includes zero. Accuracy on valid and erroneous prefixes also changes direction across scales. Parse rates range from 83\% to 99\%, and truncation rates from 0\% to 2\%, so missing or truncated judgments do not explain the result. Together, Tables~\hyperref[app:tab-d6]{D6}--\hyperref[app:tab-d8]{D8} reproduce the main asymmetry: the privileged solution greatly improves solving and recovery, but we find no evidence that it identifies the error boundary.

\subsubsection{Blind response audit}
\label{app:source-hm-audit}

Final-answer accuracy does not reveal how privileged-solution responses differ from problem-only responses. We therefore conduct an outcome-blind audit of independently generated complete answers under Teacher Thinking ON and OFF. The 225 prespecified response-pair slots contain 75 slots per scale and were selected using Teacher-ON recorded outcomes. All 225 judgments completed under Teacher Thinking OFF and 219 under Teacher Thinking ON. The latter 219 are the slots available in both modes. Response order is randomized, and GPT-5.6-Sol sees the problem and worked solution as mathematical evidence but not the generation condition or automatic-verifier outcome. Mechanism proportions condition on pairs for which the blind judge independently confirms that the problem-only response is incorrect and the privileged response is correct. Because the responses do not share a prefix, the labels describe differences between complete answers rather than a causal repair of one trajectory.

The mechanism label identifies the dominant response change: local repair preserves the original route, new strategy replaces it, supplied fact imports a decisive intermediate result, close reproduction follows distinctive reference steps, and other collects search-pruning, final-answer, or mixed changes. Deployment is labeled self-contained when the successful response supplies sufficient independent support and privilege-dependent/direct when it does not.

\begin{table*}[!t]
  \centering
  \scriptsize
  \setlength{\tabcolsep}{2pt}
  \renewcommand{\arraystretch}{1.0}

  \resizebox{\textwidth}{!}{%
  \begin{tabular}{@{}cccccccccc@{}}
  \toprule
  \textbf{Teacher} & \textbf{Matched slots} & \textbf{Confirmed} & \textbf{Local} & \textbf{New strategy} & \textbf{Supplied fact} & \textbf{Close reproduction} & \textbf{Other} & \textbf{Self-contained} & \textbf{Privilege/direct} \\
  \midrule
  Thinking ON & 219 & 105 & 12 (11.4\%) & 23 (21.9\%) & 24 (22.9\%) & 35 (33.3\%) & 11 (10.5\%) & 48 (45.7\%) & 57 (54.3\%) \\
  Thinking OFF & 219 & 80 & 11 (13.8\%) & 21 (26.3\%) & 7 (8.8\%) & 34 (42.5\%) & 7 (8.8\%) & 39 (48.8\%) & 41 (51.3\%) \\
  \bottomrule
  \end{tabular}%
  }
  \vspace{1mm}
  \caption*{\phantomsection\label{app:tab-d9a}\textbf{Table D9a:} Hendrycks MATH Level 5 blind-audit labels among judge-confirmed wrong-to-right pairs. Percentages condition on the Confirmed column. Categories are descriptive because slots were selected using Teacher-ON recorded outcomes.}
\end{table*}

\begin{table*}[!t]
  \centering
  \scriptsize
  \setlength{\tabcolsep}{6pt}
  \renewcommand{\arraystretch}{1.0}
  \begin{tabular}{@{}cccccc@{}}
  \toprule
  \textbf{Teacher} & \textbf{Model} & \textbf{Problem-only tokens} & \textbf{Privileged tokens} & \textbf{Reduction} & \textbf{Confirmed} \\
  \midrule
  Thinking ON & Qwen3-1.7B & 17,949 & 10,927 & 39.1\% & 33 \\
  Thinking ON & Qwen3-4B & 18,693 & 10,817 & 42.1\% & 36 \\
  Thinking ON & Qwen3-8B & 17,900 & 11,033 & 38.4\% & 36 \\
  Thinking OFF & Qwen3-1.7B & 1,980 & 1,289 & 34.9\% & 24 \\
  Thinking OFF & Qwen3-4B & 2,558 & 1,537 & 39.9\% & 30 \\
  Thinking OFF & Qwen3-8B & 2,414 & 1,461 & 39.5\% & 26 \\
  \bottomrule
  \end{tabular}
  \vspace{1mm}
  \caption*{\phantomsection\label{app:tab-d9b}\textbf{Table D9b:} Hendrycks MATH Level 5 response length and blind-confirmed improvement counts for the 219 slots completed in both teacher modes. Token columns are means over matched slots.}
\end{table*}

\vspace{1mm}
\noindent\textbf{Close reproduction is the largest single category.} It accounts for 35/105 confirmed improvements (33.3\%) under Teacher Thinking ON and 34/80 (42.5\%) under Teacher Thinking OFF. Same-strategy local repair accounts for only 12/105 (11.4\%) and 11/80 (13.8\%). Overall, 54.3\% and 51.3\% of improvements import or reproduce reference-specific steps without enough independent support to stand alone. Privileged responses are also 35--42\% shorter than problem-only responses at every scale in both modes. These descriptive results show that high source accuracy on Hendrycks MATH Level 5 often reflects direct use of hidden information rather than independent reasoning.

\begin{figure*}[!t]
  \centering
  \includegraphics[width=0.9\textwidth]{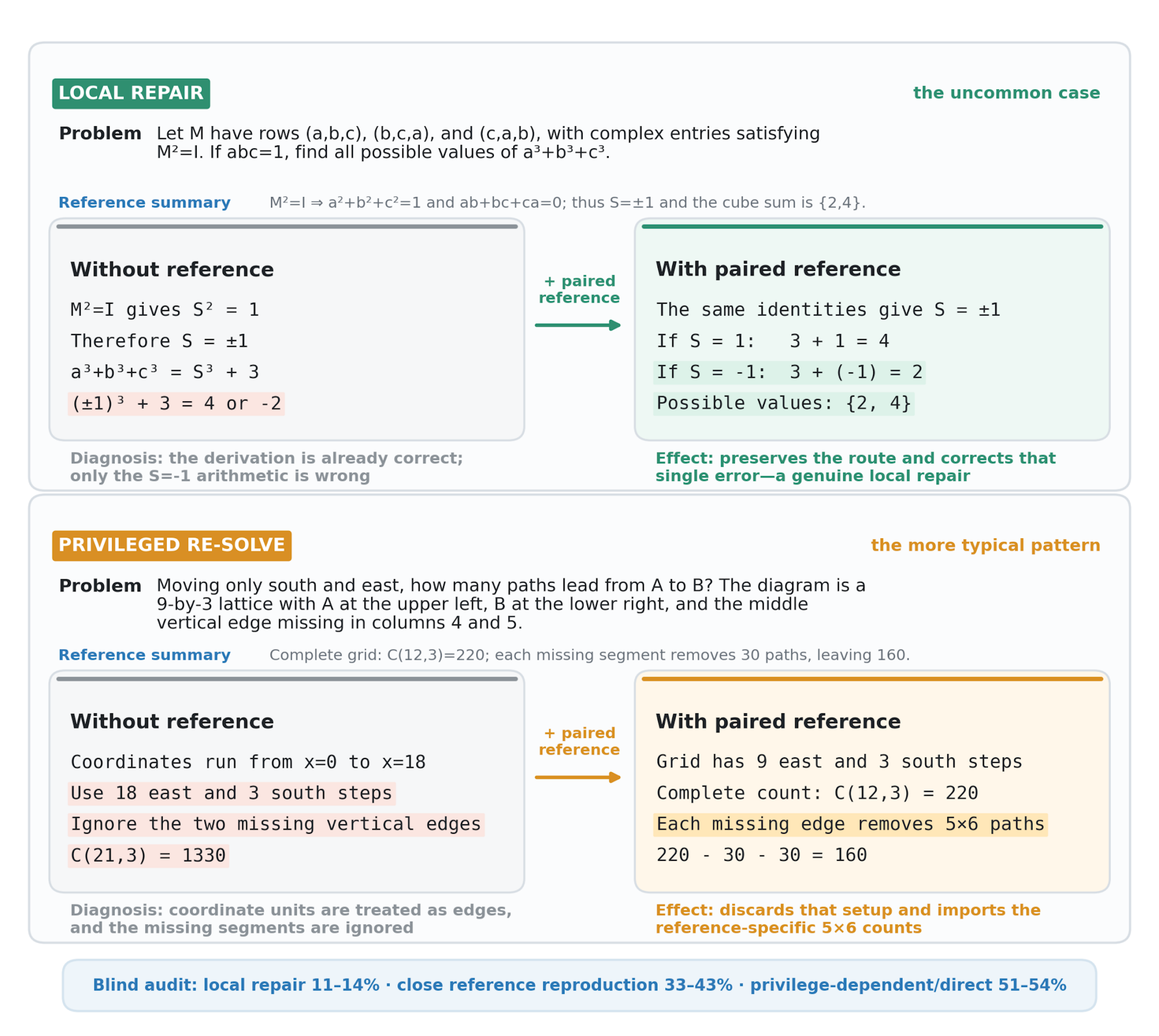}
  \caption*{\phantomsection\label{app:fig-d4}\textbf{Figure D4}: Two Hendrycks MATH Level 5 examples from the blind response audit. The top shows a same-route local repair. The bottom shows the more common pattern of replacing the original setup with reference-specific steps. Examples are condensed for legibility.}
\end{figure*}

\vspace{1mm}
\noindent\textbf{Hendrycks MATH Level 5 replicates the solving-recognition gap.} The frozen teacher can exploit a privileged solution both from scratch and after an error, but privileged access does not reliably help it identify the error boundary.

\subsection{Execution notes}
\label{app:source-exec}

\noindent\textbf{OpenThoughts.} All 1,800 planned OpenThoughts recovery generations are present and receive two independent task-level judgments, with disagreements resolved by blinded adjudication. Ambiguous labels count as incorrect in the reported estimates. The recognition panel contains four judgments per pair state (valid/erroneous prefix × problem only/privileged solution), for 2,400 judgments in total. All retained sequences remain within the native context window, and failed attempts are excluded by scheduler status alone.

\noindent\textbf{Hendrycks MATH Level 5.} All 2,880 screening responses, 2,280 fresh original-question responses, 1,800 recovery continuations, and 600 recognition judgments are present in the final Hendrycks MATH Level 5 analysis. Every screening and fresh original-question response receives two independent task-level judgments, followed by blinded adjudication of disagreements. The response-transition audit contains 225 completed Thinking-OFF judgments and 219 completed Thinking-ON judgments. The 219 matched slots are used in Tables~\hyperref[app:tab-d9a]{D9a}--\hyperref[app:tab-d9b]{D9b}. To enforce the Qwen3 native 40,960-token context window, the recovery audit uses 103 protocol-matched replacements covering two previously missing 50-row arm/seed cells and three previously over-length rows. The recognition audit uses five replacements covering four missing rows and one over-length row. The largest retained total sequence length is exactly 40,960 tokens. Replacement sources are chosen from scheduler-success states before response text is inspected, and all recovery outcomes use the paper\textquotesingle s canonical exact-answer evaluator.

\section{RQ2: Encoding}
\label{app:encoding}

This appendix follows the three RQ2 tests in the main text. Section~\ref{app:encoding-shift} documents target-shift scoring and its Hendrycks MATH Level 5 replication. Section~\ref{app:encoding-ranking} reports verified correctness-ranking dynamics, using OpenThoughts for Thinking-OFF and Hendrycks MATH Level 5 for Thinking-ON. Section~\ref{app:encoding-branch} gives the branch-selection protocol and results on both corpora.

\subsection{Target shift}
\label{app:encoding-shift}

\subsubsection{Construction and scoring}

For each model and student mode, we retain questions with at least one automatically scored correct and incorrect response among 12 reference-free frozen-Base samples generated at temperature 1.0, top-p 1.0, and top-k disabled. Thinking-OFF trajectories come from OpenThoughts and Thinking-ON trajectories from Hendrycks MATH Level 5. One trajectory of each type is chosen within question to minimize response-length mismatch, and both must contain at least 1,024 clean response tokens. This yields 200, 183, and 168 OpenThoughts pairs for Thinking-OFF trajectories at Qwen3-1.7B, 4B, and 8B, and 200 Hendrycks MATH Level 5 pairs per scale for Thinking-ON trajectories. The first 512 response tokens form the fixed context and the following 512-token suffix is scored. At every suffix position, we compute the exact full-vocabulary teacher-to-student FKL using the problem-only and privileged-solution teacher templates in Appendix~\ref{app:figure1-interfaces}. Overall FKL is averaged within trajectory and then across questions. $\Delta\text{FKL}_{\text{ref}}$ subtracts the problem-only value from the privileged-solution value on identical tokens. The paired problem bootstrap uses 20,000 replicates. Table~\hyperref[app:tab-e1a]{E1a} adds confidence intervals to the overall estimates in Table~\ref{tab:reference-fkl}.

\vspace{1mm}
\noindent\textbf{Token-family taxonomy.} We decode each scored token, strip surrounding whitespace, and lowercase it before assigning one mutually exclusive family. Epistemic markers are the exact normalized tokens \emph{wait, actually, check, hmm, perhaps, maybe, alternatively, seems, might,} and \emph{likely}. Tokens containing a digit, a TeX delimiter, or a Unicode mathematical symbol are math/numeric. Alphabetic ASCII and other non-ASCII text are lexical. Tokens composed only of punctuation or symbols form the punctuation/symbol family. Whitespace, special tokens, and control tokens form the final family. Table~\hyperref[app:tab-e1b]{E1b} reports conditional per-token means for both cohorts. These values do not sum to Overall because each is normalized by the number of tokens in that family.

\begin{table*}[!t]
  \centering
  \scriptsize
  \setlength{\tabcolsep}{2pt}
  \renewcommand{\arraystretch}{1.0}
  \begin{tabular*}{\textwidth}{@{\extracolsep{\fill}}ccccc@{}}
  \toprule
  \textbf{Student trajectories} & \textbf{Teacher mode} & \textbf{Qwen3-1.7B} & \textbf{Qwen3-4B} & \textbf{Qwen3-8B} \\
  \midrule
  Thinking OFF & Thinking OFF & 0.0508 {[}0.0463, 0.0558{]} & 0.0369 {[}0.0339, 0.0400{]} & 0.0325 {[}0.0296, 0.0356{]} \\
  Thinking OFF & Thinking ON & 0.0423 {[}0.0381, 0.0471{]} & 0.0350 {[}0.0320, 0.0381{]} & 0.0315 {[}0.0285, 0.0345{]} \\
  Thinking ON & Thinking OFF & 0.1382 {[}0.1302, 0.1463{]} & 0.1042 {[}0.0975, 0.1109{]} & 0.0993 {[}0.0928, 0.1062{]} \\
  Thinking ON & Thinking ON & 0.0761 {[}0.0678, 0.0850{]} & 0.0689 {[}0.0621, 0.0761{]} & 0.0583 {[}0.0519, 0.0650{]} \\
  \bottomrule
  \end{tabular*}
  \vspace{1mm}
  \caption*{\phantomsection\label{app:tab-e1a}\textbf{Table E1a:} $\Delta\text{FKL}_{\text{ref}}$ on fixed trajectories (OpenThoughts for Thinking OFF, Hendrycks MATH Level 5 for Thinking ON). Brackets are paired 95\% bootstrap intervals.}
\end{table*}

\begin{table*}[!t]
  \centering
  \scriptsize
  \setlength{\tabcolsep}{2pt}
  \renewcommand{\arraystretch}{1.0}
  \begin{tabular*}{\textwidth}{@{\extracolsep{\fill}}cccccccc@{}}
  \toprule
  \textbf{Model} & \textbf{Student} & \textbf{Teacher} & \textbf{Math / num.} & \textbf{Lexical} & \textbf{Epistemic} & \textbf{Punct. / sym.} & \textbf{Control / space} \\
  \midrule
  Qwen3-1.7B & OFF & OFF & 0.031 & 0.090 & \textbf{0.215} & 0.031 & 0.024 \\
  & OFF & ON & 0.025 & 0.075 & \textbf{0.144} & 0.027 & 0.022 \\
  & ON & OFF & 0.025 & 0.188 & \textbf{0.811} & 0.101 & 0.059 \\
  & ON & ON & 0.016 & 0.109 & \textbf{0.518} & 0.036 & 0.024 \\
  \midrule
  Qwen3-4B & OFF & OFF & 0.020 & 0.065 & \textbf{0.205} & 0.028 & 0.011 \\
  & OFF & ON & 0.019 & 0.060 & \textbf{0.187} & 0.030 & 0.012 \\
  & ON & OFF & 0.017 & 0.135 & \textbf{0.587} & 0.067 & 0.039 \\
  & ON & ON & 0.014 & 0.091 & \textbf{0.436} & 0.032 & 0.024 \\
  \midrule
  Qwen3-8B & OFF & OFF & 0.018 & 0.059 & \textbf{0.163} & 0.026 & 0.009 \\
  & OFF & ON & 0.017 & 0.056 & \textbf{0.150} & 0.027 & 0.009 \\
  & ON & OFF & 0.026 & 0.131 & \textbf{0.442} & 0.063 & 0.039 \\
  & ON & ON & 0.015 & 0.081 & \textbf{0.270} & 0.027 & 0.019 \\
  \bottomrule
  \end{tabular*}
  \vspace{1mm}
  \caption*{\phantomsection\label{app:tab-e1b}\textbf{Table E1b:} Token-family $\Delta\text{FKL}_{\text{ref}}$ (OpenThoughts for Thinking OFF, Hendrycks MATH Level 5 for Thinking ON). Values are conditional per-token means. Student and Teacher denote thinking modes.}
\end{table*}

\subsubsection{Hendrycks MATH Level 5 replication}

We begin from 200 frozen Hendrycks MATH Level 5 questions per scale that have mixed outcomes under Thinking ON. Each receives 12 reference-free Thinking-OFF samples under the same temperature 1.0, top-p 1.0, top-k-disabled, 38,912-response-token, and 40,960-context-token protocol. Requiring mixed outcomes in both modes leaves 91, 101, and 108 matched questions at Qwen3-1.7B, 4B, and 8B. Both trajectory modes then use the same fixed questions, minimum-length-mismatch pairing rule, and scored positions. Tables~\hyperref[app:tab-e2]{E2} and \hyperref[app:tab-e3]{E3} repeat the signal-magnitude and token-family analyses on these cleaned Hendrycks MATH Level 5 cohorts, using the same 20,000-draw problem bootstrap.

\begin{table*}[!t]
  \centering
  \scriptsize
  \setlength{\tabcolsep}{2pt}
  \renewcommand{\arraystretch}{1.0}
  \begin{tabular*}{\textwidth}{@{\extracolsep{\fill}}ccccc@{}}
  \toprule
  \textbf{Student trajectories} & \textbf{Teacher mode} & \textbf{Qwen3-1.7B} & \textbf{Qwen3-4B} & \textbf{Qwen3-8B} \\
  \midrule
  Thinking OFF & Thinking OFF & 0.0375 {[}0.0331, 0.0422{]} & 0.0370 {[}0.0324, 0.0421{]} & 0.0301 {[}0.0267, 0.0338{]} \\
  Thinking OFF & Thinking ON & 0.0324 {[}0.0281, 0.0368{]} & 0.0360 {[}0.0311, 0.0414{]} & 0.0306 {[}0.0269, 0.0345{]} \\
  Thinking ON & Thinking OFF & 0.1437 {[}0.1296, 0.1581{]} & 0.0971 {[}0.0883, 0.1061{]} & 0.0954 {[}0.0867, 0.1045{]} \\
  Thinking ON & Thinking ON & 0.0863 {[}0.0713, 0.1026{]} & 0.0624 {[}0.0538, 0.0718{]} & 0.0579 {[}0.0486, 0.0680{]} \\
  \bottomrule
  \end{tabular*}
  \vspace{1mm}
  \caption*{\phantomsection\label{app:tab-e2}\textbf{Table E2:} Hendrycks MATH Level 5 $\Delta\text{FKL}_{\text{ref}}$ replication on matched fixed trajectories. Brackets are 95\% confidence intervals.}
\end{table*}

\begin{table*}[!t]
  \centering
  \scriptsize
  \setlength{\tabcolsep}{2pt}
  \renewcommand{\arraystretch}{1.0}
  \begin{tabular*}{\textwidth}{@{\extracolsep{\fill}}cccccccc@{}}
  \toprule
  \textbf{Model} & \textbf{Student} & \textbf{Teacher} & \textbf{Math / num.} & \textbf{Lexical} & \textbf{Epistemic} & \textbf{Punct. / sym.} & \textbf{Control / space} \\
  \midrule
  Qwen3-1.7B & OFF & OFF & 0.023 & 0.069 & \textbf{0.132} & 0.023 & 0.012 \\
  & OFF & ON & 0.021 & 0.058 & \textbf{0.085} & 0.021 & 0.013 \\
  & ON & OFF & 0.025 & 0.202 & \textbf{0.827} & 0.104 & 0.057 \\
  & ON & ON & 0.017 & 0.128 & \textbf{0.586} & 0.041 & 0.022 \\
  \midrule
  Qwen3-4B & OFF & OFF & 0.019 & 0.061 & \textbf{0.105} & 0.026 & 0.013 \\
  & OFF & ON & 0.019 & 0.057 & \textbf{0.097} & 0.028 & 0.013 \\
  & ON & OFF & 0.016 & 0.131 & \textbf{0.555} & 0.063 & 0.033 \\
  & ON & ON & 0.014 & 0.085 & \textbf{0.414} & 0.030 & 0.019 \\
  \midrule
  Qwen3-8B & OFF & OFF & 0.015 & 0.053 & \textbf{0.138} & 0.021 & 0.011 \\
  & OFF & ON & 0.015 & 0.052 & \textbf{0.133} & 0.024 & 0.012 \\
  & ON & OFF & 0.029 & 0.128 & \textbf{0.414} & 0.063 & 0.040 \\
  & ON & ON & 0.017 & 0.082 & \textbf{0.254} & 0.029 & 0.018 \\
  \bottomrule
  \end{tabular*}
  \vspace{1mm}
  \caption*{\phantomsection\label{app:tab-e3}\textbf{Table E3:} Hendrycks MATH Level 5 token-family $\Delta\text{FKL}_{\text{ref}}$ replication on fixed suffixes. Values are conditional per-token means.}
\end{table*}

\vspace{1mm}
\noindent\textbf{The magnitude result replicates across corpora.} Every model×student-mode×teacher-mode cell in Tables~\hyperref[app:tab-e1a]{E1a} and \hyperref[app:tab-e2]{E2} has a positive $\Delta\text{FKL}_{\text{ref}}$ interval that excludes zero. Epistemic markers have the largest conditional shift in every corresponding cell in Tables~\hyperref[app:tab-e1b]{E1b} and \hyperref[app:tab-e3]{E3}. These results establish that the reference changes the target, but they do not establish that the change points toward correct reasoning.

\subsection{Correctness ranking}
\label{app:encoding-ranking}

\subsubsection{Cohorts and metric}

GPT-5.6-Sol and Claude Opus 4.8 independently evaluate both trajectories in all 1,151 automatically selected correct-incorrect pairs from Appendix~\ref{app:encoding-shift}, with a separate blinded pass resolving disagreements. We retain a pair only when the final labels preserve its correct-incorrect ordering; 894 pairs remain. The Thinking-OFF cohorts come from OpenThoughts and contain 174, 150, and 139 pairs at Qwen3-1.7B, 4B, and 8B. The Thinking-ON cohorts come from Hendrycks MATH Level 5 and contain 156, 145, and 130 pairs. Excluded pairs are not relabeled into a new contrast.

Each trajectory score averages the per-token readout over a fixed 512-token suffix. Exact FKL measures the full teacher-to-student divergence. The executed loss clips each vocabulary summand at 0.05 before aggregation, matching the training objective. AUROC treats the incorrect trajectory as the positive class, gives ties half credit, and equals 0.5 at chance. This trajectory-level ranking is a coarse diagnostic rather than a necessary condition for useful token-local learning. Section~\ref{app:encoding-branch} provides the more local test at matched error states. Step 0 uses one frozen-Base estimate. Later checkpoints average three prespecified training seeds. Confidence intervals use 20,000 paired problem resamples at Step 0 and hierarchical seed-and-problem resamples thereafter.

\subsubsection{Training dynamics}

\begin{figure*}[!t]
  \centering
  \includegraphics[width=0.55\textwidth]{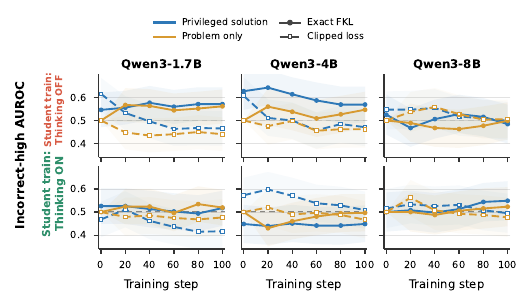}
  \caption*{\phantomsection\label{app:fig-e1}\textbf{Figure E1}: Incorrect-high AUROC across checkpoints with 95\% intervals (shaded). Blue/orange denotes privileged-solution/problem-only context. Circles/squares denote exact FKL/clipped loss. The gray line marks chance. Step 0 is the frozen Base with problem resampling. Later checkpoints average three training seeds with hierarchical seed-and-problem resampling.}
\end{figure*}

We rescore the verified pairs at Steps 0, 20, 40, 60, 80, and 100. The student distribution comes from the corresponding checkpoint, while the teacher remains the frozen Thinking-ON Base model. At Step 0, the problem-only readout is constant, yielding AUROC 0.5 under the tie convention above. Tables~\hyperref[app:tab-e4a]{E4a}--\hyperref[app:tab-e4b]{E4b} report Step-100 estimates with confidence intervals, and Figure~\hyperref[app:fig-e1]{E1} plots every checkpoint with its interval.

\begin{table*}[!t]
  \centering
  \scriptsize
  \setlength{\tabcolsep}{2pt}
  \renewcommand{\arraystretch}{1.0}
  \begin{tabular*}{\textwidth}{@{\extracolsep{\fill}}ccccc@{}}
  \toprule
  \textbf{Student training} & \textbf{Model (pairs)} & \textbf{Privileged-solution AUROC} & \textbf{Problem-only AUROC} & \textbf{Privileged minus problem} \\
  \midrule
  Thinking OFF & Qwen3-1.7B (174) & 0.571 {[}0.496, 0.644{]} & 0.561 {[}0.487, 0.634{]} & +0.010 {[}-0.033, +0.054{]} \\
  & Qwen3-4B (150) & 0.569 {[}0.489, 0.647{]} & 0.547 {[}0.469, 0.624{]} & +0.022 {[}-0.033, +0.076{]} \\
  & Qwen3-8B (139) & 0.484 {[}0.400, 0.568{]} & 0.499 {[}0.417, 0.578{]} & -0.014 {[}-0.055, +0.024{]} \\
  \midrule
  Thinking ON & Qwen3-1.7B (156) & 0.517 {[}0.440, 0.592{]} & 0.519 {[}0.442, 0.596{]} & -0.002 {[}-0.075, +0.071{]} \\
  & Qwen3-4B (145) & 0.448 {[}0.370, 0.529{]} & 0.497 {[}0.416, 0.577{]} & -0.048 {[}-0.124, +0.028{]} \\
  & Qwen3-8B (130) & 0.549 {[}0.464, 0.633{]} & 0.523 {[}0.438, 0.608{]} & +0.026 {[}-0.056, +0.108{]} \\
  \bottomrule
  \end{tabular*}
  \vspace{1mm}
  \caption*{\phantomsection\label{app:tab-e4a}\textbf{Table E4a:} Step-100 \textbf{exact-FKL} incorrect-high AUROC on verified fixed pairs. Intervals use a hierarchical seed-and-problem bootstrap.}
\end{table*}

\begin{table*}[!t]
  \centering
  \scriptsize
  \setlength{\tabcolsep}{2pt}
  \renewcommand{\arraystretch}{1.0}
  \begin{tabular*}{\textwidth}{@{\extracolsep{\fill}}ccccc@{}}
  \toprule
  \textbf{Student training} & \textbf{Model (pairs)} & \textbf{Privileged-solution AUROC} & \textbf{Problem-only AUROC} & \textbf{Privileged minus problem} \\
  \midrule
  Thinking OFF & Qwen3-1.7B (174) & 0.466 {[}0.391, 0.540{]} & 0.441 {[}0.368, 0.513{]} & +0.025 {[}-0.036, +0.086{]} \\
  & Qwen3-4B (150) & 0.471 {[}0.389, 0.553{]} & 0.462 {[}0.382, 0.540{]} & +0.009 {[}-0.051, +0.067{]} \\
  & Qwen3-8B (139) & 0.499 {[}0.417, 0.580{]} & 0.506 {[}0.424, 0.590{]} & -0.007 {[}-0.053, +0.036{]} \\
  \midrule
  Thinking ON & Qwen3-1.7B (156) & 0.417 {[}0.342, 0.494{]} & 0.476 {[}0.400, 0.556{]} & -0.060 {[}-0.128, +0.004{]} \\
  & Qwen3-4B (145) & 0.508 {[}0.428, 0.589{]} & 0.467 {[}0.386, 0.547{]} & +0.041 {[}-0.021, +0.103{]} \\
  & Qwen3-8B (130) & 0.495 {[}0.410, 0.579{]} & 0.477 {[}0.392, 0.562{]} & +0.018 {[}-0.054, +0.090{]} \\
  \bottomrule
  \end{tabular*}
  \vspace{1mm}
  \caption*{\phantomsection\label{app:tab-e4b}\textbf{Table E4b:} Step-100 \textbf{clipped-loss} incorrect-high AUROC on verified fixed pairs. Intervals use a hierarchical seed-and-problem bootstrap.}
\end{table*}

\vspace{1mm}
\noindent\textbf{Correctness ranking remains unstable.} Exact FKL shows a temporary privileged-context advantage for Qwen3-4B Thinking-OFF at Steps 20--60 (+0.082 [0.007, 0.158], +0.076 [0.007, 0.144], and +0.078 [0.016, 0.142]), but no pattern persists across scale or mode. Only two other cells exclude zero, each at a single checkpoint with a lower bound of 0.002: Qwen3-4B Thinking-ON clipped loss at Step 40 (+0.080) and Qwen3-8B Thinking-OFF exact FKL at Step 60 (+0.065). At Step 100, every privileged-minus-problem interval in Tables~\hyperref[app:tab-e4a]{E4a}--\hyperref[app:tab-e4b]{E4b} includes zero. The benchmark gains in Figure~\ref{fig:interface-factorial} therefore do not coincide with a reliable privilege-specific ranking of incorrect trajectories.

\subsection{Branch choice}
\label{app:encoding-branch}

\subsubsection{Protocol}

From each audited prefix ending before an error, the frozen Thinking-ON student sees only the problem and prefix. Candidate proposals use seeds 100--115, temperature 1.0, top-p 0.95, and top-k 20. The first four unique, full-length 128-token branches in seed order are retained. The same frozen checkpoint then scores each branch at temperature 1.1 as a problem-only student and as a privileged-solution Thinking-ON teacher. The student selector chooses the branch with the highest student mean log likelihood. The privileged selector chooses the branch with the highest teacher-minus-student mean log likelihood. Each fixed branch then receives four matched reference-free Thinking-ON continuations using outcome seeds 0, 1, 2, and 3 under the same sampling settings. The 38,912-token response budget includes the prefix and branch, and all sequences remain within the native 40,960-token context window. Comparing sibling branches from the same state removes between-problem difficulty, and every selected outcome receives blinded task-level adjudication. Final answers are first scored by the paper\textquotesingle s canonical exact-answer evaluator. The main estimates use the blinded full-population adjudication in Section~\ref{app:encoding-adjudication}. All branches, scores, and continuations are fixed before outcome adjudication or selector comparison. State-bootstrap intervals use 50,000 replicates and preserve the four candidates and their matched continuations as a unit.

\subsubsection{Cohort construction and adjudication}
\label{app:encoding-adjudication}

State selection is frozen before branch generation. GPT-5.6-Sol and Claude Opus 4.8 independently inspect one non-truncated Base trajectory per problem, selected by a prespecified hash ordering, in 128-token chunks and retain a state only when both identify a recoverable mathematical error with at least 256 tokens of valid reasoning before that error. No state is filtered using selector scores or downstream outcomes.

Before completing the 4B and 8B outcome evaluations, we preregistered the following interpretation rule: describe weak branch-level information only if a positive OpenThoughts effect appears at two or more Qwen3 scales. Otherwise describe any 1.7B result as scale-specific. We report all six corpus × scale cells regardless of outcome and treat cross-corpus agreement as stronger evidence than one interval excluding zero.

\noindent\textbf{Outcome adjudication.} Because the automatic verifier misses many correct answers on these continuations (Table~\hyperref[app:tab-e7]{E7}), final branch-success estimates replace automatic labels with a task-level protocol. It evaluates all 4,928 unique rollouts selected by either selector. GPT-5.6-Sol judges every response while blind to corpus, model scale, selector identity, branch score, and automatic label. The 1,374 responses for which this judgment disagreed with the automatic label or was ambiguous were independently reviewed by Claude Opus 4.8. A third blinded GPT-5.6-Sol pass resolved the 280 disagreements between the two model judges. The primary analysis conservatively counts the remaining 163 ambiguous responses as incorrect (Table~\hyperref[app:tab-e5]{E5}). Table~\hyperref[app:tab-e6]{E6} reports the two alternative treatments.

\noindent\textbf{Expanded-panel construction.} The frozen expansion targets 150 source states per corpus × scale cell, with at most one state per source question. Every retained state satisfies the same dual-judge error-consensus rule and supplies four unique, full-length candidate branches before selector scoring. Candidate coverage removes one OpenThoughts Qwen3-1.7B state, leaving 149. All 150 OpenThoughts Qwen3-4B and 8B states remain. The available Hendrycks MATH Level 5 consensus pools retain 150, 68, and 59 states at 1.7B, 4B, and 8B, respectively. No state is filtered using selector scores or downstream outcomes.

\begin{table*}[!t]
  \centering
  \scriptsize
  \setlength{\tabcolsep}{2pt}
  \renewcommand{\arraystretch}{1.0}
  \begin{tabular*}{\textwidth}{@{\extracolsep{\fill}}ccccccc@{}}
  \toprule
  \textbf{Corpus} & \textbf{Model} & \textbf{States} & \textbf{Student} & \textbf{Privileged} & \textbf{Privileged $-$ student} & \textbf{Improve / tie / harm} \\
  \midrule
  OpenThoughts & Qwen3-1.7B & 149 & 57.38\% & 63.09\% & +5.70 {[}+1.51, +10.07{]} & 36 / 98 / 15 \\
  OpenThoughts & Qwen3-4B & 150 & 73.67\% & 74.33\% & +0.67 {[}-1.67, +3.00{]} & 14 / 126 / 10 \\
  OpenThoughts & Qwen3-8B & 150 & 80.50\% & 79.33\% & -1.17 {[}-3.67, +1.17{]} & 9 / 129 / 12 \\
  \midrule
  Hendrycks MATH Level 5 & Qwen3-1.7B & 150 & 30.50\% & 29.50\% & -1.00 {[}-4.50, +2.50{]} & 26 / 97 / 27 \\
  Hendrycks MATH Level 5 & Qwen3-4B & 68 & 41.18\% & 45.96\% & +4.78 {[}+0.00, +9.93{]} & 15 / 47 / 6 \\
  Hendrycks MATH Level 5 & Qwen3-8B & 59 & 50.00\% & 49.58\% & -0.42 {[}-5.08, +4.66{]} & 7 / 42 / 10 \\
  \bottomrule
  \end{tabular*}
  \vspace{1mm}
  \caption*{\phantomsection\label{app:tab-e5}\textbf{Table E5:} Fully adjudicated branch-selection results. Accuracy averages four matched continuations per selected branch. Intervals use 50,000 state-bootstrap resamples.}
  \vspace{3mm}
  \begin{tabular*}{\textwidth}{@{\extracolsep{\fill}}ccccc@{}}
  \toprule
  \textbf{Corpus} & \textbf{Model} & \textbf{Ambiguous = incorrect} & \textbf{Ambiguous = 0.5} & \textbf{Ambiguous excluded} \\
  \midrule
  OpenThoughts & Qwen3-1.7B & +5.70 {[}+1.51, +10.07{]} (n=149) & +5.70 {[}+1.68, +9.82{]} (n=149) & +5.29 {[}+1.24, +9.52{]} (n=148) \\
  OpenThoughts & Qwen3-4B & +0.67 {[}-1.67, +3.00{]} (n=150) & +0.58 {[}-1.83, +3.08{]} (n=150) & +0.74 {[}-1.87, +3.40{]} (n=147) \\
  OpenThoughts & Qwen3-8B & -1.17 {[}-3.67, +1.17{]} (n=150) & -1.50 {[}-3.83, +0.58{]} (n=150) & -1.40 {[}-3.80, +0.67{]} (n=149) \\
  \midrule
  Hendrycks MATH Level 5 & Qwen3-1.7B & -1.00 {[}-4.50, +2.50{]} (n=150) & -0.67 {[}-4.33, +2.92{]} (n=150) & -0.56 {[}-4.28, +3.11{]} (n=150) \\
  Hendrycks MATH Level 5 & Qwen3-4B & +4.78 {[}+0.00, +9.93{]} (n=68) & +4.60 {[}-0.37, +9.74{]} (n=68) & +4.78 {[}-0.25, +9.80{]} (n=68) \\
  Hendrycks MATH Level 5 & Qwen3-8B & -0.42 {[}-5.08, +4.66{]} (n=59) & -1.69 {[}-6.14, +2.97{]} (n=59) & -1.01 {[}-5.75, +3.74{]} (n=58) \\
  \bottomrule
  \end{tabular*}
  \vspace{1mm}
  \caption*{\phantomsection\label{app:tab-e6}\textbf{Table E6:} Ambiguity sensitivity of the privileged-minus-student effect under three scoring rules.}
\end{table*}

\begin{table*}[!t]
  \centering
  \scriptsize
  \setlength{\tabcolsep}{6pt}
  \renewcommand{\arraystretch}{1.0}
  \begin{tabular}{@{}lccc@{}}
  \toprule
  \textbf{Stratum} & \textbf{Audited / non-ambiguous} & \textbf{False negatives} & \textbf{Judged correct share} \\
  \midrule
  Qwen3-1.7B & 67 / 67 & 11 & 16.4\% {[}9.4, 27.1{]} \\
  Qwen3-4B & 67 / 66 & 36 & 54.5\% {[}42.6, 66.0{]} \\
  Qwen3-8B & 66 / 65 & 42 & 64.6\% {[}52.5, 75.1{]} \\
  \midrule[0.3pt]
  Both selectors & 50 / 49 & 20 & 40.8\% {[}28.2, 54.8{]} \\
  Student only & 51 / 50 & 23 & 46.0\% {[}33.0, 59.6{]} \\
  Privileged only & 51 / 51 & 24 & 47.1\% {[}34.1, 60.5{]} \\
  Neither selector & 48 / 48 & 22 & 45.8\% {[}32.6, 59.7{]} \\
  \bottomrule
  \end{tabular}
  \vspace{1mm}
  \caption*{\phantomsection\label{app:tab-e7}\textbf{Table E7:} Dual-judge audit of 200 OpenThoughts branch continuations labeled incorrect by the automatic verifier. A false negative is a response the judges deem correct. The final column divides these counts by non-ambiguous audited responses. The upper rows split the sample by model scale and the lower rows split the same sample by selector category. Wilson intervals use non-ambiguous judgments. The balanced sample is descriptive rather than population-weighted.}
\end{table*}

\vspace{1mm}
\noindent\textbf{Branch selection is not robust.} The preregistered two-scale OpenThoughts criterion is not met after full adjudication. Only the 1.7B OpenThoughts interval strictly excludes zero. The Hendrycks MATH Level 5 4B interval reaches zero at its lower endpoint and becomes unresolved when ambiguous judgments receive half credit or are excluded. Under either treatment, the OpenThoughts 1.7B estimate remains positive, while every other interval includes zero. We therefore describe the detected advantage as local and scale-specific rather than as a general branch-ranking capability.

\subsubsection{Verifier audit}

We audit whether automatic answer-label errors could create the selector contrast. Before inspection, we sample 200 responses labeled incorrect from the expanded OpenThoughts experiment, balancing the three scales and four selector categories: chosen by both selectors, only the student selector, only the privileged selector, or neither. At most one response is retained per audited error state. GPT-5.6-Sol and Claude Opus 4.8 independently receive the problem, reference solution, extracted answer, and response, but not model scale, selector identity, or branch scores. They judge mathematical equivalence, including option-label/value equivalence. Raw agreement is 97.0\%. A separate blinded pass adjudicates the six disagreements.

\vspace{1mm}
\noindent\textbf{Final selector estimates use full adjudication.} In this balanced sample of verifier-negative responses, the share judged correct ranges from 16.4\% at 1.7B to 64.6\% at 8B. The privileged-only and student-only shares are 47.1\% and 46.0\%, a +1.06pp contrast with a bootstrap interval of {[}-18.71, +20.78{]}. This interval is too wide to validate selector differences of only a few pp, which motivated the full-population adjudication in Section~\ref{app:encoding-adjudication}. Table~\hyperref[app:tab-e7]{E7} is therefore a diagnostic of the automatic verifier, not the source of the paper's final branch-success estimates.

\section{RQ3: Transmission}
\label{app:transmission}

Section~\ref{app:transmission-main} gives the OpenThoughts analyses of pass@8, repeated training, access to the supplied solution, and thinking mode behind Figures~\ref{fig:training-pass8}-\ref{fig:privilege-interface}. Section~\ref{app:retention} reports the longitudinal retention controls behind Figure~\ref{fig:retention}, including Qwen mathematics, OLMo, and coding. Because the protocols differ, all comparisons are made only within matched designs.

\subsection{OpenThoughts acquisition and thinking-mode audit}
\label{app:transmission-main}

\subsubsection{Question selection and training epochs}

We screen all 3,200 examples separately for each Qwen3 scale and student generation mode. Each problem receives 12 reference-free frozen-Base samples under the same Thinking-ON or Thinking-OFF mode later used to define that training condition. \textbf{Unsolved} contains problems with 0/12 correct samples. \textbf{Frontier} contains problems with 1--5/12. Student Thinking-OFF runs therefore train on OFF-defined buckets, and Student Thinking-ON runs train on ON-defined buckets. Buckets are model-relative and are never pooled across scales or modes.

When a bucket contains at most 160 problems, we retain all of them. Larger Unsolved pools are capped at 160 with a fixed hash ordering. Larger Frontier pools are capped at 160 while balancing the five Base-success counts. Every retained row has a non-empty privileged solution, a valid reference answer matching the dataset answer, and a verifier-compatible final answer. The worked-solution audit labels every retained solution valid or minor-issue. None is invalid. Training otherwise follows Appendix~\ref{app:figure1-training}. We count training epochs as the number of examples processed divided by the number of selected training questions: $e = 32s/N$, where $s$ is the optimizer step, 32 is the batch size, and $N$ is the training-set size. We compare checkpoints nearest 1, 2, 4, 8, and 16 epochs within each set. Table~\hyperref[app:tab-f1a]{F1a} lists the numbers of selected questions and the exact optimizer steps. These steps differ across sets, so epoch labels are approximate.

\begin{table*}[!t]
  \centering
  \scriptsize
  \setlength{\tabcolsep}{2pt}
  \renewcommand{\arraystretch}{1.0}
  \begin{tabular*}{\textwidth}{@{\extracolsep{\fill}}cccccc@{}}
  \toprule
  \textbf{Student mode} & \textbf{Model} & \textbf{Unsolved avail. / used} & \textbf{Unsolved steps} & \textbf{Frontier avail. / used} & \textbf{Frontier steps} \\
  \midrule
  Thinking OFF & Qwen3-1.7B & 261 / 160 & 5 / 10 / 20 / 40 / 80 & 721 / 160 & 5 / 10 / 20 / 40 / 80 \\
  Thinking OFF & Qwen3-4B & 121 / 121 & 4 / 8 / 15 / 30 / 61 & 381 / 160 & 5 / 10 / 20 / 40 / 80 \\
  Thinking OFF & Qwen3-8B & 156 / 156 & 5 / 10 / 20 / 39 / 78 & 399 / 160 & 5 / 10 / 20 / 40 / 80 \\
  \midrule
  Thinking ON & Qwen3-1.7B & 106 / 106 & 3 / 7 / 13 / 27 / 53 & 111 / 111 & 3 / 7 / 14 / 28 / 56 \\
  Thinking ON & Qwen3-4B & 107 / 107 & 3 / 7 / 13 / 27 / 54 & 45 / 45 & 1 / 3 / 6 / 11 / 23 \\
  Thinking ON & Qwen3-8B & 97 / 97 & 3 / 6 / 12 / 24 / 49 & 35 / 35 & 1 / 2 / 4 / 9 / 18 \\
  \bottomrule
  \end{tabular*}
  \vspace{1mm}
  \caption*{\phantomsection\label{app:tab-f1a}\textbf{Table F1a:} OpenThoughts question-set sizes and the exact optimizer steps at checkpoints nearest 1 / 2 / 4 / 8 / 16 training epochs. Epoch count is $32s/N$, with optimizer step $s$ and training-set size $N$.}
\end{table*}

\subsubsection{Pass@8 and frequent success on Frontier questions}

Each Thinking-ON-teacher, student-mode, and teacher-input condition is trained separately on its corresponding bucket with three matched training seeds. Every checkpoint is evaluated reference-free on the same questions under both student thinking modes using eight fresh samples per problem, temperature 1.0, top-p 1.0, top-k disabled, and a 38,912-token response cap. Figure~\ref{fig:training-pass8} reports pass@8 for matched student training and evaluation modes: the number of questions with at least one correct answer among eight samples, divided by the number evaluated. The Base points use a separate eight-sample evaluation after the 12-sample screening. Checkpoint values average the three training seeds equally, and shading shows the seed range. The two thinking modes use separately selected question sets, so their curves are not a same-question comparison.

Table~\hyperref[app:tab-f1b]{F1b} retains the stricter Frontier analysis: the percentage of questions with matched-mode Base below 7/8 that reach at least 7/8 correct responses after training. Rates are computed within seed and then averaged equally. Because initially unsuccessful questions are more sensitive to false-negative grading, Section~\ref{app:verified-unsolved} uses the separate dual-judge 0/16 rerun. Automatic-verifier Unsolved threshold rates are omitted.

\begin{table*}[!t]
  \centering
  \scriptsize
  \setlength{\tabcolsep}{2pt}
  \renewcommand{\arraystretch}{1.0}
  \begin{tabular*}{\textwidth}{@{\extracolsep{\fill}}ccccccc@{}}
  \toprule
  \textbf{Model} & \textbf{Bucket} & \textbf{Train} & \textbf{n OFF} & \textbf{Eval OFF: 1 / 2 / 4 / 8 / 16} & \textbf{n ON} & \textbf{Eval ON: 1 / 2 / 4 / 8 / 16} \\
  \midrule
  Qwen3-1.7B & Frontier & OFF & 159 & 0.6 / 1.5 / 1.5 / 3.4 / 6.1 & 34 & 23.5 / 29.4 / 30.4 / 44.1 / 42.2 \\
  Qwen3-1.7B & Frontier & ON & 98 & 1.4 / 0.0 / 0.0 / 1.7 / 1.0 & 108 & 1.2 / 2.2 / 3.4 / 6.8 / 7.1 \\
  \midrule
  Qwen3-4B & Frontier & OFF & 155 & 2.2 / 1.5 / 3.7 / 6.7 / 20.0 & 16 & 12.5 / 12.5 / 8.3 / 10.4 / 10.4 \\
  Qwen3-4B & Frontier & ON & 25 & 5.3 / 4.0 / 5.3 / 6.7 / 5.3 & 44 & 2.3 / 1.5 / 3.0 / 2.3 / 5.3 \\
  \midrule
  Qwen3-8B & Frontier & OFF & 156 & 0.6 / 1.1 / 0.4 / 4.9 / 23.7 & 16 & 10.4 / 16.7 / 8.3 / 10.4 / 20.8 \\
  Qwen3-8B & Frontier & ON & 20 & 6.7 / 11.7 / 15.0 / 6.7 / 13.3 & 35 & 1.0 / 1.9 / 1.9 / 1.9 / 2.9 \\
  \bottomrule
  \end{tabular*}
  \vspace{1mm}
  \caption*{\phantomsection\label{app:tab-f1b}\textbf{Table F1b:} Frontier questions reaching at least 7/8 correct responses from a matched-mode Base below 7/8 (\%). Train and evaluation columns denote student thinking mode. The teacher is Thinking ON with the privileged solution. Cells are means across three matched training seeds. Values follow checkpoints nearest 1, 2, 4, 8, and 16 epochs.}
\end{table*}

\subsubsection{Privilege and thinking-mode controls}

\begin{figure*}[!t]
  \centering
  \includegraphics[width=0.8\textwidth]{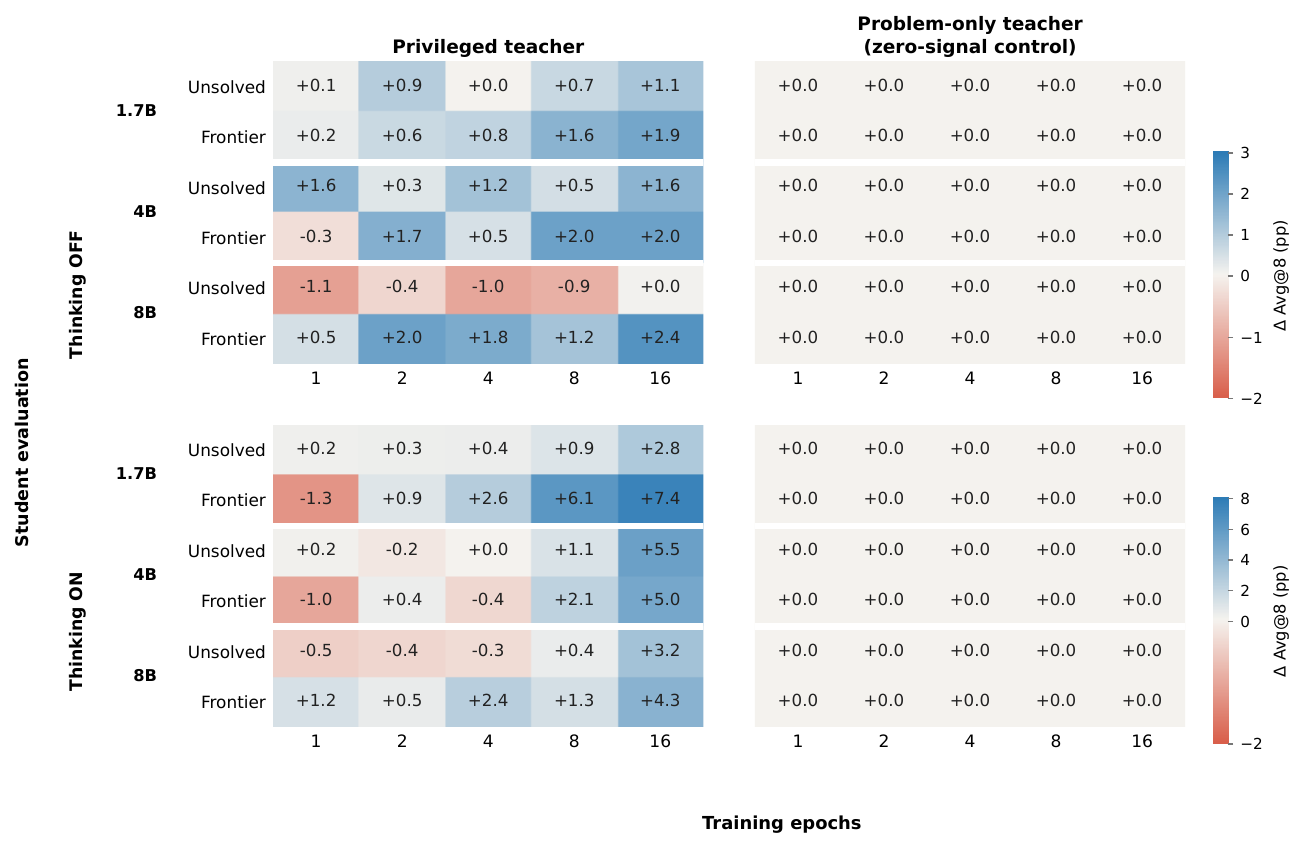}
  \phantomsection\label{app:fig-f1}
  \caption*{\textbf{Figure F1}: Student Thinking-ON training control on mode-specific OpenThoughts buckets. The left column reports privileged-solution OPSD. Cells show reference-free Avg@8 change from the corresponding mode-specific Base, averaged over three training seeds. The right column is the exact zero-signal problem-only control. Checkpoints are nearest 1, 2, 4, 8, and 16 epochs. Color scales differ across evaluation modes.}
\end{figure*}

Table~\hyperref[app:tab-f1c]{F1c} gives every cell of Figure~\ref{fig:privilege-interface}. The teacher remains Thinking ON and the student trains Thinking OFF. Only privileged-solution versus problem-only teacher input changes. Table~\hyperref[app:tab-f1d]{F1d} gives the Student Thinking-ON companion in Figure~\hyperref[app:fig-f1]{F1}. For the matched Thinking-ON problem-only arm, teacher and student receive identical inputs and begin from identical weights, so the OPSD loss and gradient are exactly zero and every cell remains at Base.

\begin{table*}[!t]
  \centering
  \scriptsize
  \setlength{\tabcolsep}{2pt}
  \renewcommand{\arraystretch}{0.9}
  \begin{tabular*}{\textwidth}{@{\extracolsep{\fill}}ccccc@{}}
  \toprule
  \textbf{Model} & \textbf{Bucket} & \textbf{Teacher input} & \textbf{Eval OFF $\Delta$: 1 / 2 / 4 / 8 / 16} & \textbf{Eval ON $\Delta$: 1 / 2 / 4 / 8 / 16} \\
  \midrule
  Qwen3-1.7B & Unsolved & Privileged & +1.8 / +2.1 / +2.8 / +4.1 / +7.7 & +0.5 / +0.7 / +2.9 / +4.0 / +5.1 \\
  Qwen3-1.7B & Frontier & Privileged & +0.1 / +2.0 / +2.8 / +2.8 / +9.7 & -0.1 / +0.5 / +1.4 / +2.2 / +2.2 \\
  Qwen3-4B & Unsolved & Privileged & +0.5 / +1.9 / +6.0 / +9.6 / +23.0 & -0.0 / -0.4 / -0.1 / +0.6 / +1.5 \\
  Qwen3-4B & Frontier & Privileged & +1.3 / +3.1 / +5.7 / +10.2 / +27.7 & -0.5 / -0.5 / -0.8 / -0.6 / -1.2 \\
  Qwen3-8B & Unsolved & Privileged & +0.7 / +1.5 / +3.7 / +15.9 / +24.3 & +0.7 / +0.0 / +0.5 / +0.6 / -0.9 \\
  Qwen3-8B & Frontier & Privileged & +2.0 / +0.8 / +0.0 / +12.3 / +32.3 & -0.3 / -0.6 / +0.1 / -0.7 / -0.1 \\
  \midrule
  Qwen3-1.7B & Unsolved & Problem only & +1.9 / +1.7 / +1.1 / +2.3 / +11.0 & +1.1 / +1.1 / +1.2 / +3.1 / +3.7 \\
  Qwen3-1.7B & Frontier & Problem only & -0.5 / +1.1 / -1.5 / -0.6 / +16.7 & -0.3 / +0.1 / +0.6 / +1.3 / +2.0 \\
  Qwen3-4B & Unsolved & Problem only & -0.1 / +1.8 / +2.6 / +3.2 / +22.1 & -1.0 / -0.2 / -0.1 / +0.6 / +1.7 \\
  Qwen3-4B & Frontier & Problem only & +0.5 / +0.1 / -2.2 / +3.1 / +34.0 & -0.7 / -0.9 / -0.9 / -0.5 / -1.2 \\
  Qwen3-8B & Unsolved & Problem only & +1.1 / +1.1 / +0.8 / +5.4 / +17.3 & +0.3 / +0.5 / -0.0 / +0.5 / -2.9 \\
  Qwen3-8B & Frontier & Problem only & -0.1 / +0.6 / -2.6 / +9.6 / +34.2 & -0.6 / +0.2 / -0.1 / -0.1 / -1.3 \\
  \bottomrule
  \end{tabular*}
  \phantomsection\label{app:tab-f1c}
  \vspace{1mm}
  \caption*{\textbf{Table F1c:} Complete Figure~\ref{fig:privilege-interface} Avg@8 changes from matched Base (pp), averaged equally over three training seeds. Student training is Thinking OFF. Values follow checkpoints nearest 1, 2, 4, 8, and 16 epochs.}
\end{table*}

\begin{table*}[!t]
  \centering
  \scriptsize
  \setlength{\tabcolsep}{2pt}
  \renewcommand{\arraystretch}{0.9}
  \begin{tabular*}{\textwidth}{@{\extracolsep{\fill}}ccccc@{}}
  \toprule
  \textbf{Model} & \textbf{Bucket} & \textbf{Teacher input} & \textbf{Eval OFF $\Delta$: 1 / 2 / 4 / 8 / 16} & \textbf{Eval ON $\Delta$: 1 / 2 / 4 / 8 / 16} \\
  \midrule
  Qwen3-1.7B & Unsolved & Privileged & +0.1 / +0.9 / +0.0 / +0.7 / +1.1 & +0.2 / +0.3 / +0.4 / +0.9 / +2.8 \\
  Qwen3-1.7B & Frontier & Privileged & +0.2 / +0.6 / +0.8 / +1.6 / +1.9 & -1.3 / +0.9 / +2.6 / +6.1 / +7.4 \\
  \midrule
  Qwen3-4B & Unsolved & Privileged & +1.6 / +0.3 / +1.2 / +0.5 / +1.6 & +0.2 / -0.2 / -0.0 / +1.1 / +5.5 \\
  Qwen3-4B & Frontier & Privileged & -0.3 / +1.7 / +0.5 / +2.0 / +2.0 & -1.0 / +0.4 / -0.4 / +2.1 / +5.0 \\
  \midrule
  Qwen3-8B & Unsolved & Privileged & -1.1 / -0.4 / -1.0 / -0.9 / +0.0 & -0.5 / -0.4 / -0.3 / +0.4 / +3.2 \\
  Qwen3-8B & Frontier & Privileged & +0.5 / +2.0 / +1.8 / +1.2 / +2.4 & +1.2 / +0.5 / +2.4 / +1.3 / +4.3 \\
  \bottomrule
  \end{tabular*}
  \phantomsection\label{app:tab-f1d}
  \vspace{1mm}
  \caption*{\textbf{Table F1d:} Complete Figure~\hyperref[app:fig-f1]{F1} Avg@8 changes from matched Base (pp), averaged equally over three training seeds. Student training is Thinking ON. The problem-only zero-signal control is 0.0 in every cell. Values follow checkpoints nearest 1, 2, 4, 8, and 16 epochs.}
\end{table*}

\subsubsection{Retention between checkpoints}

Table~\hyperref[app:tab-f1e]{F1e} preserves question and training-seed identity to test whether a Frontier conversion at one checkpoint remains stable at the next. Each entry pools the three seeds: P is the number of question-seed units stable at the earlier checkpoint and R is the subset for which the same question in the same seed remains stable at the next checkpoint. These are descriptive pooled counts, not counts of unique questions.

\begin{table*}[!t]
  \centering
  \scriptsize
  \setlength{\tabcolsep}{2pt}
  \renewcommand{\arraystretch}{0.9}
  \begin{tabular*}{\textwidth}{@{\extracolsep{\fill}}ccccccccccc@{}}
  \toprule
  & & & \multicolumn{4}{c}{\textbf{Eval OFF (R/P)}} & \multicolumn{4}{c}{\textbf{Eval ON (R/P)}} \\
  \cmidrule(lr){4-7}\cmidrule(l){8-11}
  \textbf{Model} & \textbf{Bucket} & \textbf{Train} & \textbf{1$\rightarrow$2} & \textbf{2$\rightarrow$4} & \textbf{4$\rightarrow$8} & \textbf{8$\rightarrow$16} & \textbf{1$\rightarrow$2} & \textbf{2$\rightarrow$4} & \textbf{4$\rightarrow$8} & \textbf{8$\rightarrow$16} \\
  \midrule
  Qwen3-1.7B & Frontier & OFF & 0/3 & 1/7 & 1/7 & 3/16 & 17/24 & 19/30 & 27/31 & 35/45 \\
  Qwen3-1.7B & Frontier & ON & 0/4 & --- & --- & 1/5 & 0/4 & 1/7 & 4/11 & 8/22 \\
  \midrule
  Qwen3-4B & Frontier & OFF & 1/10 & 2/7 & 5/17 & 21/31 & 3/6 & 2/6 & 2/4 & 4/5 \\
  Qwen3-4B & Frontier & ON & 1/4 & 2/3 & 2/4 & 3/5 & 1/3 & 0/2 & 0/4 & 0/3 \\
  \midrule
  Qwen3-8B & Frontier & OFF & 0/3 & 0/5 & 1/2 & 14/23 & 4/5 & 4/8 & 4/4 & 5/5 \\
  Qwen3-8B & Frontier & ON & 2/4 & 5/7 & 3/9 & 3/4 & 0/1 & 0/2 & 1/2 & 1/2 \\
  \bottomrule
  \end{tabular*}
  \vspace{1mm}
  \caption*{\phantomsection\label{app:tab-f1e}\textbf{Table F1e:} Frontier retention between checkpoints nearest adjacent epochs (1, 2, 4, 8, and 16) with a Thinking-ON privileged-solution teacher. Entries report retained/prior-stable question-seed units (R/P), pooled over three seeds. A dash marks P = 0.}
\end{table*}

\subsubsection{Held-out transfer}

To separate question-level fitting from broader transfer, we evaluate each Student Thinking-OFF checkpoint nearest 16 epochs on AIME 2024, AIME 2025, AIME 2026, HMMT November 2025, and HMMT February 2026. None of these benchmark questions appears in the OpenThoughts training buckets. Table~\hyperref[app:tab-f1f]{F1f} reports checkpoint-minus-matched-Base macro Avg@12. Intervals independently resample three training seeds and three Base decoding seeds. The listed n is the number of OpenThoughts bucket questions used for training in each seed.

\begin{table*}[!t]
  \centering
  \scriptsize
  \setlength{\tabcolsep}{2pt}
  \renewcommand{\arraystretch}{0.9}
  \begin{tabular*}{\textwidth}{@{\extracolsep{\fill}}cccccc@{}}
  \toprule
  \textbf{Model} & \textbf{Bucket} & \textbf{Teacher input} & \textbf{Train n} & \textbf{Eval OFF $\Delta$ {[}95\% CI{]}} & \textbf{Eval ON $\Delta$ {[}95\% CI{]}} \\
  \midrule
  Qwen3-1.7B & Unsolved & Privileged & 160 & -0.05 {[}-0.23, +0.19{]} & +4.74 {[}+4.21, +5.27{]} \\
  Qwen3-1.7B & Unsolved & Problem only & 160 & +3.78 {[}+2.88, +4.76{]} & +5.21 {[}+4.41, +5.94{]} \\
  Qwen3-1.7B & Frontier & Privileged & 160 & +0.43 {[}-0.05, +0.72{]} & +5.14 {[}+4.42, +5.78{]} \\
  Qwen3-1.7B & Frontier & Problem only & 160 & +4.38 {[}+2.34, +6.20{]} & +5.63 {[}+3.95, +7.03{]} \\
  \midrule
  Qwen3-4B & Unsolved & Privileged & 121 & +9.12 {[}+7.91, +10.15{]} & +1.31 {[}+1.11, +1.54{]} \\
  Qwen3-4B & Unsolved & Problem only & 121 & +15.98 {[}+14.97, +16.80{]} & +1.74 {[}+1.34, +2.03{]} \\
  Qwen3-4B & Frontier & Privileged & 160 & +10.66 {[}+9.77, +11.57{]} & +1.24 {[}+0.99, +1.57{]} \\
  Qwen3-4B & Frontier & Problem only & 160 & +22.70 {[}+22.03, +23.31{]} & +1.40 {[}+0.58, +1.93{]} \\
  \midrule
  Qwen3-8B & Unsolved & Privileged & 156 & +5.68 {[}+4.23, +7.83{]} & +1.98 {[}+1.31, +2.44{]} \\
  Qwen3-8B & Unsolved & Problem only & 156 & +20.39 {[}+18.90, +21.57{]} & +2.53 {[}+1.79, +3.26{]} \\
  Qwen3-8B & Frontier & Privileged & 160 & +9.50 {[}+6.79, +11.58{]} & +2.04 {[}+1.35, +2.74{]} \\
  Qwen3-8B & Frontier & Problem only & 160 & +22.53 {[}+20.63, +25.70{]} & +2.16 {[}+1.29, +3.09{]} \\
  \bottomrule
  \end{tabular*}
  \phantomsection\label{app:tab-f1f}
  \vspace{1mm}
  \caption*{\textbf{Table F1f:} Held-out five-benchmark Avg@12 changes from the matched-mode Base after approximately 16 epochs. All checkpoints use a Thinking-ON teacher and Student Thinking-OFF training. Brackets give descriptive 95\% bootstrap intervals.}
\end{table*}

\subsubsection{Questions with no correct Base responses}
\label{app:verified-unsolved}

To check initially unsuccessful questions more carefully, GPT-5.6-Sol and Claude Opus 4.8 independently grade 16 frozen-Base samples per Student Thinking-OFF question, with blinded adjudication of disagreements. Eligibility requires all 16 samples to be incorrect and none ambiguous, leaving 60, 24, and 11 questions at 1.7B, 4B, and 8B. We rerun privileged-solution and problem-only training on these verified question sets. At each checkpoint, the canonical answer-equivalence evaluator scores 16 fresh samples per question. The dual-judge audit determines which questions are included and does not relabel checkpoint generations. Stable acquisition requires at least 14/16 correct samples under matched Thinking-OFF evaluation. Thinking-ON results use the same questions selected under Thinking OFF and therefore measure cross-mode reachability rather than zero-start acquisition. The same evaluator credits 55.6--79.2\% of Thinking-ON samples on these questions (Table~\hyperref[app:tab-f1g]{F1g}), so its false negatives cannot by themselves explain the near-zero Thinking-OFF rates.

\begin{table*}[!t]
  \centering
  \scriptsize
  \setlength{\tabcolsep}{2pt}
  \renewcommand{\arraystretch}{1.0}
  \resizebox{\textwidth}{!}{%
  \begin{tabular}{@{}ccccccc@{}}
  \toprule
  \textbf{Model} & \textbf{Teacher input} & \textbf{n} & \textbf{Eval OFF Avg@16} & \textbf{Eval OFF stable} & \textbf{Eval ON Avg@16} & \textbf{Eval ON stable} \\
  \midrule
  Qwen3-1.7B & Privileged & 60 & 3.5 / 3.4 / 3.8 / 4.8 / 6.4 & 0 / 0 / 0 / 0 / 0 & 55.6 / 56.3 / 56.5 / 60.7 / 63.9 & 31.7 / 33.3 / 36.1 / 43.3 / 45.6 \\
  Qwen3-1.7B & Problem only & 60 & 3.4 / 3.8 / 3.3 / 3.1 / 7.2 & 0 / 0 / 0 / 0 / 0 & 55.9 / 56.5 / 55.7 / 57.8 / 61.7 & 33.3 / 31.7 / 31.1 / 36.1 / 45.6 \\
  \midrule
  Qwen3-4B & Privileged & 24 & 3.0 / 3.6 / 2.7 / 4.6 / 5.1 & 0 / 0 / 0 / 0 / 0 & 64.0 / 64.0 / 65.3 / 64.1 / 65.6 & 47.2 / 44.4 / 51.4 / 48.6 / 48.6 \\
  Qwen3-4B & Problem only & 24 & 2.5 / 3.4 / 3.9 / 3.8 / 5.0 & 0 / 0 / 0 / 0 / 0 & 64.4 / 64.1 / 63.9 / 64.2 / 65.3 & 47.2 / 50.0 / 47.2 / 47.2 / 47.2 \\
  \midrule
  Qwen3-8B & Privileged & 11 & 4.0 / 6.6 / 4.4 / 4.2 / 3.2 & 0 / 0 / 0 / 0 / 0 & 79.2 / 77.1 / 77.7 / 78.0 / 78.2 & 72.7 / 63.6 / 72.7 / 69.7 / 72.7 \\
  Qwen3-8B & Problem only & 11 & 5.9 / 5.9 / 5.5 / 4.7 / 5.9 & 0 / 0 / 0 / 0 / 0 & 77.3 / 76.7 / 78.4 / 78.8 / 78.0 & 63.6 / 69.7 / 66.7 / 69.7 / 72.7 \\
  \bottomrule
  \end{tabular}%
  }
  \phantomsection\label{app:tab-f1g}
  \vspace{1mm}
  \caption*{\textbf{Table F1g:} Verified Unsolved results at checkpoints nearest 1 / 2 / 4 / 8 / 16 epochs. Avg@16 and stable rates (\%) average three training seeds. Stable means at least 14/16 correct samples. OFF stable is strict acquisition, and ON stable is reachability on questions selected using the Thinking-OFF Base.}
\end{table*}

\subsection{Purification and retention}
\label{app:retention}

\subsubsection{Purified target}
\label{app:purified-target}

We implement Purified OPSD following \citet{shen2026purified}. For every on-policy student prefix $\hat{y}_{<t}$, the same frozen Base model is evaluated under three inputs: a teacher distribution $p_T(\cdot\mid \hat{y}_{<t},q,r)$ with question $q$ and reference $r$; a reference-only probe $p_{\mathrm{ref}}(\cdot\mid \hat{y}_{<t},r)$ with the question removed; and a clean Base distribution $p_0(\cdot\mid \hat{y}_{<t},q)$ with the reference removed. The student $p_\theta(\cdot\mid \hat{y}_{<t},q)$ is the only network evaluated with gradients.

Over the full vocabulary $V$, the raw transferable signal is $\Delta_{it}(v)=\log p_T(v)-\log p_{\text{ref}}(v)$. Subtracting its vocabulary mean gives the centered signal $\bar{\Delta}_{it}(v)$, to which we apply the soft clip of \citet{shen2026purified}, $\widetilde{\Delta}_{it}(v)=c\tanh(\bar{\Delta}_{it}(v)/c)$, with $c=10$. The target is $p_{\text{target}}=\operatorname{softmax}(\log p_0+\widetilde{\Delta}_{it}/\beta_{\text{PMI}})$, with $\beta_{\text{PMI}}=1$. Canonical Purified runs minimize the response-position mean of the symmetric generalized Jensen--Shannon divergence between $p_\theta$ and $p_{\text{target}}$ (mixture coefficient 0.5), using the full vocabulary without an additional token-level loss clip after vocabulary aggregation. The PMI correction strength and JSD mixture coefficient are distinct hyperparameters.

Our Standard comparator is the paper-wide executed baseline: the privileged teacher distribution is used directly with teacher-to-student forward KL, with each vocabulary contribution upper-clipped at 0.05. Thus the canonical Standard-versus-Purified curves compare the complete executed recipes. The OLMo RawExact-versus-PurifiedExact pair separately holds exact FKL fixed and changes only target construction.

\subsubsection{Qwen mathematics and order controls}

The Qwen analyses train Qwen3-1.7B, 4B, and 8B on the same audited OpenThoughts corpus for 100 updates with three matched training seeds per arm. Teacher and student rollout modes are Thinking ON. Evaluation is reference-free Thinking ON Avg@12 on AIME 2024, AIME 2025, AIME 2026, HMMT November 2025, and HMMT February 2026, with temperature 1.0, top-p 1.0, top-k disabled, and a 38,912-token response cap. We first macro-average the five benchmarks within each seed and then average seeds equally. The frozen Base is independently repeated with decoding seeds 0, 1, and 2. Base-relative intervals use 50,000 independent two-sample bootstrap resamples of the three training seeds and three Base decoding seeds. Pure Standard and Purified runs are evaluated at Steps 20, 40, 60, 80, and 100. The switch trajectories change objective after Step 20. Matched-composition endpoints additionally reverse 20/80 and 80/20 update blocks. Table~\hyperref[app:tab-f2a]{F2a} gives every plotted point and the across-seed range used as Figure~\ref{fig:retention}a shading.

\begin{table*}[!t]
  \centering
  \scriptsize
  \setlength{\tabcolsep}{2pt}
  \renewcommand{\arraystretch}{1.0}
  \resizebox{\textwidth}{!}{%
  \begin{tabular}{@{}cccccccc@{}}
  \toprule
  \textbf{Model} & \textbf{Schedule} & \textbf{Base mean {[}95\% CI{]}} & \textbf{Step 20} & \textbf{Step 40} & \textbf{Step 60} & \textbf{Step 80} & \textbf{Step 100} \\
  \midrule
  Qwen3-1.7B & Standard & 36.26 {[}35.80, 36.60{]} & 36.88 {[}36.75, 37.04{]} & 35.93 {[}35.71, 36.10{]} & 34.18 {[}34.04, 34.38{]} & 34.87 {[}33.71, 35.97{]} & 34.35 {[}33.27, 34.91{]} \\
  Qwen3-1.7B & Purified & 36.26 {[}35.80, 36.60{]} & 36.77 {[}36.06, 37.66{]} & 37.51 {[}36.71, 37.92{]} & 36.71 {[}36.61, 36.78{]} & 37.44 {[}36.96, 37.99{]} & 38.12 {[}37.10, 38.79{]} \\
  Qwen3-1.7B & S20$\rightarrow$P80 & 36.26 {[}35.80, 36.60{]} & 36.88 {[}36.75, 37.04{]} & 36.57 {[}35.46, 37.59{]} & 37.85 {[}37.02, 38.44{]} & 38.27 {[}38.07, 38.57{]} & 37.30 {[}36.08, 38.30{]} \\
  Qwen3-1.7B & P20$\rightarrow$S80 & 36.26 {[}35.80, 36.60{]} & 36.77 {[}36.06, 37.66{]} & 38.01 {[}37.52, 38.59{]} & 34.98 {[}34.14, 36.21{]} & 34.95 {[}34.53, 35.24{]} & 35.03 {[}33.90, 35.63{]} \\
  \midrule
  Qwen3-4B & Standard & 60.42 {[}60.33, 60.51{]} & 61.17 {[}60.71, 61.80{]} & 59.77 {[}59.67, 59.89{]} & 58.78 {[}58.61, 58.88{]} & 57.72 {[}57.49, 58.06{]} & 57.79 {[}57.36, 58.53{]} \\
  Qwen3-4B & Purified & 60.42 {[}60.33, 60.51{]} & 61.30 {[}60.54, 62.21{]} & 61.31 {[}60.69, 62.21{]} & 61.04 {[}60.72, 61.32{]} & 61.10 {[}60.67, 61.44{]} & 61.16 {[}61.04, 61.30{]} \\
  Qwen3-4B & S20$\rightarrow$P80 & 60.42 {[}60.33, 60.51{]} & 61.17 {[}60.71, 61.80{]} & 60.16 {[}59.63, 60.55{]} & 60.27 {[}59.57, 61.39{]} & 60.81 {[}60.42, 61.18{]} & 61.25 {[}60.83, 61.72{]} \\
  Qwen3-4B & P20$\rightarrow$S80 & 60.42 {[}60.33, 60.51{]} & 61.30 {[}60.54, 62.21{]} & 61.88 {[}61.21, 63.03{]} & 60.29 {[}60.14, 60.49{]} & 59.62 {[}58.63, 60.62{]} & 59.07 {[}58.55, 59.40{]} \\
  \midrule
  Qwen3-8B & Standard & 62.52 {[}62.06, 63.20{]} & 63.72 {[}63.64, 63.84{]} & 63.58 {[}62.98, 64.24{]} & 60.31 {[}59.24, 60.91{]} & 59.16 {[}58.44, 59.83{]} & 58.96 {[}58.67, 59.33{]} \\
  Qwen3-8B & Purified & 62.52 {[}62.06, 63.20{]} & 64.75 {[}64.49, 64.89{]} & 63.73 {[}63.01, 64.52{]} & 63.26 {[}62.27, 64.07{]} & 64.03 {[}63.31, 64.99{]} & 64.11 {[}63.18, 65.26{]} \\
  Qwen3-8B & S20$\rightarrow$P80 & 62.52 {[}62.06, 63.20{]} & 63.72 {[}63.64, 63.84{]} & 63.79 {[}63.30, 64.31{]} & 63.40 {[}63.31, 63.54{]} & 63.75 {[}63.52, 63.91{]} & 63.74 {[}63.46, 64.06{]} \\
  Qwen3-8B & P20$\rightarrow$S80 & 62.52 {[}62.06, 63.20{]} & 64.75 {[}64.49, 64.89{]} & 63.98 {[}63.03, 64.94{]} & 63.42 {[}63.08, 63.61{]} & 60.11 {[}59.46, 60.62{]} & 59.58 {[}59.12, 59.85{]} \\
  \bottomrule
  \end{tabular}%
  }
  \vspace{1mm}
  \caption*{\phantomsection\label{app:tab-f2a}\textbf{Table F2a:} Complete Qwen mathematics trajectories in Figure~\ref{fig:retention}a. Checkpoints are three-training-seed mean Avg@12, with brackets giving the seed range. Base is the mean over three independent decoding seeds. Its brackets give the 95\% seed-bootstrap interval. S and P denote Standard and Purified OPSD.}
\end{table*}

At Step 100, independent two-sample seed-bootstrap intervals for Standard minus Base (pp) are -1.92 {[}-2.92, -1.11{]}, -2.63 {[}-3.07, -1.92{]}, and -3.57 {[}-4.23, -2.97{]} at 1.7B, 4B, and 8B. The corresponding Purified-minus-Base estimates are +1.86 {[}+0.90, +2.69{]}, +0.74 {[}+0.60, +0.89{]}, and +1.58 {[}+0.51, +2.73{]}.

\begin{table*}[!t]
  \centering
  \scriptsize
  \setlength{\tabcolsep}{5pt}
  \renewcommand{\arraystretch}{1.0}
  \begin{tabular}{@{}llrrr@{}}
  \toprule
  \textbf{Model} & \textbf{Composition} & \textbf{Standard last} & \textbf{Purified last} & \textbf{P-last $-$ S-last} \\
  \midrule
  Qwen3-1.7B & 80\% S + 20\% P & 34.02 & 36.62 & +2.60 {[}+2.16, +3.11{]} \\
  Qwen3-1.7B & 20\% S + 80\% P & 37.93 & 37.84 & -0.09 {[}-0.54, +0.39{]} \\
  \midrule
  Qwen3-4B & 80\% S + 20\% P & 59.30 & 59.90 & +0.60 {[}+0.19, +1.08{]} \\
  Qwen3-4B & 20\% S + 80\% P & 61.25 & 60.88 & -0.37 {[}-0.94, +0.36{]} \\
  \midrule
  Qwen3-8B & 80\% S + 20\% P & 59.94 & 63.22 & +3.27 {[}+3.16, +3.42{]} \\
  Qwen3-8B & 20\% S + 80\% P & 63.80 & 63.95 & +0.15 {[}-0.89, +1.19{]} \\
  \bottomrule
  \end{tabular}
  \phantomsection\label{app:tab-f2c}
  \vspace{1mm}
  \caption*{\textbf{Table F2c:} Fixed-composition OpenThoughts order audit. These are separately trained runs, so their endpoints need not match the switch trajectories in Table~\hyperref[app:tab-f2a]{F2a}. Scores are five-benchmark Avg@12 averaged over three paired training seeds. Brackets give paired seed-bootstrap 95\% confidence intervals.}
\end{table*}

\noindent\textbf{Extension to 200 updates.} We continue one prespecified Qwen3-4B Standard and Purified seed from Step 100 and evaluate at Steps 150 and 200. Standard scores 56.68 and 56.77, while Purified scores 61.58 and 61.94. A separate frozen-Base evaluation under the same protocol scores 60.56. This single-seed extension shows that the retention separation persists for another 100 updates rather than disappearing immediately after the main evaluation horizon.

\subsubsection{OLMo exact-FKL control}

We train OLMo-3-7B-Think \citep{olmo2025olmo3} on Hendrycks MATH with three prespecified seeds per arm and evaluate every checkpoint on the same five held-out mathematics benchmarks. RawExact and PurifiedExact use the same prompt, temperature, LoRA configuration, 1,024-token training-rollout cap, 100-update schedule, and unclipped exact forward-KL loss. Only target construction differs. PurifiedExact retains the canonical three-pass, centered, tanh-clipped PMI target ($\beta=1$, $c=10$) but deliberately replaces the canonical JSD with exact FKL. It is therefore a target-construction ablation, not a second claim of exact paper replication. The released clipped Standard run falls from 54.39\% at Step 20 to 14.08\% at Step 100. Comparing it directly with canonical Purified would confound target construction with loss and temperature, so those rows are supplementary diagnostics.

\begin{table*}[!t]
  \centering
  \scriptsize
  \setlength{\tabcolsep}{2pt}
  \renewcommand{\arraystretch}{1.0}
  \begin{tabular*}{\textwidth}{@{\extracolsep{\fill}}cccccc@{}}
  \toprule
  \textbf{Arm} & \textbf{Step 20} & \textbf{Step 40} & \textbf{Step 60} & \textbf{Step 80} & \textbf{Step 100} \\
  \midrule
  RawExact & 54.98 {[}54.92, 55.11{]} & 53.50 {[}52.22, 55.45{]} & 52.44 {[}51.37, 53.11{]} & 52.69 {[}51.97, 53.33{]} & 52.35 {[}52.06, 52.88{]} \\
  PurifiedExact & 57.43 {[}56.84, 57.84{]} & 57.66 {[}57.35, 58.13{]} & 57.04 {[}56.44, 57.88{]} & 58.13 {[}57.34, 58.73{]} & 58.52 {[}58.09, 59.16{]} \\
  Standard, clipped & 54.39 & 51.68 & 43.58 & 29.24 & 14.08 \\
  Purified, canonical & 57.89 & 57.52 & 59.59 & 57.08 & 57.82 \\
  \bottomrule
  \end{tabular*}
  \vspace{1mm}
  \caption*{\phantomsection\label{app:tab-f3}\textbf{Table F3:} OLMo-3-7B-Think five-benchmark Avg@12 across training on Hendrycks MATH. RawExact and PurifiedExact report three-seed means with seed ranges in brackets. The remaining supplementary rows are seed-0 diagnostics. Frozen Base is 55.76\%.}
\end{table*}

\noindent\textbf{Three-seed exact-FKL replication.} Mean RawExact Avg@12 across Steps 20/40/60/80/100 is 54.98 / 53.50 / 52.44 / 52.69 / 52.35. Mean PurifiedExact is 57.43 / 57.66 / 57.04 / 58.13 / 58.52. At Step 100, RawExact scores 52.88 / 52.12 / 52.06 across seeds, while PurifiedExact scores 58.32 / 59.16 / 58.09. The +6.17pp mean endpoint separation therefore replicates across all three seeds.

\subsubsection{Code replication}

Qwen3-4B trains for 100 updates on 3,200 audited CodeContests problems \citep{li2022competition} with batch size 32, LoRA rank 64, and Thinking ON for student rollouts and the privileged teacher. The public student prompt and teacher solution/reprompt templates are byte-identical to the released SDPO code \citep{hubotter2026reinforcement}. The frozen teacher receives the judge-verified reference program in the successful-attempt slot. Purified uses the target defined by \citet{shen2026purified} and summarized in Section~\ref{app:purified-target}. This is an SDPO-prompt-aligned domain transfer, not a replication of SDPO's data or RL objective. Evaluation is reference-free Thinking ON with four samples per problem, temperature 0.6, top-p 0.95, top-k disabled, and a 38,912-token response cap. LiveCodeBench v6 (LCBv6) \citep{jain2024livecodebench} contains 131 problems and the disjoint CodeContests held-out set contains 128. Base is one fixed decoding evaluation. The 8K and 16K checkpoint summaries average three matched training seeds. The 32K pilot uses one. Standard improves early but degrades as truncation rises, while Purified remains near Base. Tables~\hyperref[app:tab-f4a]{F4a}--\hyperref[app:tab-f4b]{F4b} give the complete coding trajectories and budget sensitivity.

\begin{table*}[!t]
  \centering
  \scriptsize
  \setlength{\tabcolsep}{2pt}
  \renewcommand{\arraystretch}{1.0}
  \begin{tabular*}{\textwidth}{@{\extracolsep{\fill}}ccccccc@{}}
  \toprule
  \textbf{Benchmark} & \textbf{Arm} & \textbf{Metric} & \textbf{Base} & \textbf{Step 20} & \textbf{Step 60} & \textbf{Step 100} \\
  \midrule
  LCBv6 & Standard & Accuracy & 44.66 & 46.88 {[}45.99, 47.90{]} & 43.26 {[}42.37, 43.70{]} & 40.27 {[}39.69, 41.03{]} \\
   &  & Truncation & 3.82 & 4.20 {[}3.44, 4.96{]} & 15.33 {[}13.74, 16.41{]} & 29.39 {[}28.82, 30.34{]} \\
   & Purified & Accuracy & 44.66 & 43.00 {[}41.22, 43.89{]} & 43.00 {[}41.98, 44.47{]} & 44.72 {[}43.70, 45.99{]} \\
   &  & Truncation & 3.82 & 5.34 {[}4.96, 5.73{]} & 4.64 {[}4.58, 4.77{]} & 3.75 {[}3.24, 4.39{]} \\
  \midrule[0.3pt]
  CodeContests & Standard & Accuracy & 42.58 & 46.35 {[}46.09, 46.68{]} & 38.48 {[}37.30, 39.45{]} & 35.29 {[}34.96, 35.55{]} \\
   &  & Truncation & 6.25 & 6.45 {[}6.25, 6.64{]} & 26.43 {[}24.61, 28.91{]} & 44.21 {[}43.36, 45.51{]} \\
   & Purified & Accuracy & 42.58 & 41.67 {[}40.82, 42.77{]} & 40.23 {[}39.45, 41.60{]} & 41.86 {[}40.23, 44.14{]} \\
   &  & Truncation & 6.25 & 5.53 {[}4.49, 6.45{]} & 7.68 {[}5.47, 8.98{]} & 6.25 {[}5.27, 7.23{]} \\
  \bottomrule
  \end{tabular*}
  \vspace{1mm}
  \caption*{\phantomsection\label{app:tab-f4a}\textbf{Table F4a:} Complete 8K coding trajectories in Figure~\ref{fig:retention}c-d. Rows give Avg@4 accuracy and truncation percentage. Trained checkpoints report three-seed mean {[}minimum, maximum{]}.}
\end{table*}

\begin{table*}[!t]
  \centering
  \scriptsize
  \setlength{\tabcolsep}{2pt}
  \renewcommand{\arraystretch}{1.0}
  \begin{tabular*}{\textwidth}{@{\extracolsep{\fill}}ccccccccc@{}}
  \toprule
  \textbf{Budget} & \textbf{Seeds} & \textbf{Benchmark} & \textbf{Arm} & \textbf{Steps} & \textbf{Metric} & \textbf{Checkpoint 1} & \textbf{Checkpoint 2} & \textbf{Checkpoint 3} \\
  \midrule
  16K & 3 & LCBv6 & Standard & 20/60/100 & Accuracy & 47.07 {[}46.18, 48.47{]} & 39.76 {[}38.55, 40.46{]} & 37.60 {[}37.02, 38.55{]} \\
   &  &  &  &  & Truncation & 2.86 {[}2.48, 3.24{]} & 25.38 {[}23.09, 28.05{]} & 45.29 {[}43.51, 46.37{]} \\
  16K & 3 & LCBv6 & Purified & 20/60/100 & Accuracy & 44.53 {[}42.75, 46.18{]} & 43.58 {[}41.98, 44.66{]} & 44.78 {[}43.32, 46.76{]} \\
   &  &  &  &  & Truncation & 4.39 {[}4.01, 4.58{]} & 4.45 {[}3.63, 5.53{]} & 4.07 {[}3.82, 4.39{]} \\
  \midrule[0.3pt]
  16K & 3 & CodeContests & Standard & 20/60/100 & Accuracy & 46.35 {[}45.70, 47.07{]} & 34.11 {[}33.01, 35.35{]} & 26.17 {[}25.20, 26.95{]} \\
   &  &  &  &  & Truncation & 4.75 {[}3.52, 7.03{]} & 43.36 {[}41.41, 46.88{]} & 62.57 {[}61.91, 63.09{]} \\
  16K & 3 & CodeContests & Purified & 20/60/100 & Accuracy & 41.21 {[}39.84, 43.55{]} & 40.62 {[}40.43, 40.82{]} & 39.71 {[}38.09, 40.62{]} \\
   &  &  &  &  & Truncation & 5.79 {[}4.69, 6.64{]} & 7.16 {[}6.45, 7.81{]} & 5.92 {[}5.86, 6.05{]} \\
  \midrule[0.3pt]
  32K & 1 & LCBv6 & Standard & 10/20/40 & Accuracy & 44.85 & 47.52 & 38.36 \\
   &  &  &  &  & Truncation & 3.05 & 4.39 & 41.41 \\
  32K & 1 & LCBv6 & Purified & 10/20/40 & Accuracy & 46.37 & 44.66 & 44.66 \\
   &  &  &  &  & Truncation & 4.58 & 5.15 & 5.92 \\
  \midrule[0.3pt]
  32K & 1 & CodeContests & Standard & 10/20/40 & Accuracy & 41.60 & 45.90 & 27.34 \\
   &  &  &  &  & Truncation & 4.88 & 5.08 & 64.65 \\
  32K & 1 & CodeContests & Purified & 10/20/40 & Accuracy & 41.41 & 40.62 & 42.97 \\
   &  &  &  &  & Truncation & 6.05 & 6.45 & 5.27 \\
  \bottomrule
  \end{tabular*}
  \vspace{1mm}
  \caption*{\phantomsection\label{app:tab-f4b}\textbf{Table F4b:} Complete coding training-budget sensitivity trajectories. Checkpoints 1--3 follow the listed steps. 16K entries are three-seed mean {[}minimum, maximum{]}, whereas the 32K pilot is one seed. All evaluations use the Figure~\ref{fig:retention}c-d long-context protocol.}
\end{table*}

\begin{table*}[!t]
  \centering
  \scriptsize
  \setlength{\tabcolsep}{2pt}
  \renewcommand{\arraystretch}{1.0}
  \begin{tabular*}{\textwidth}{@{\extracolsep{\fill}}ccccc@{}}
  \toprule
  \textbf{Benchmark} & \textbf{Schedule} & \textbf{Accuracy mean {[}seed range{]}} & \textbf{$\Delta$ from Base} & \textbf{Truncation mean {[}seed range{]}} \\
  \midrule
  LCBv6 & S20$\rightarrow$P80 & 44.59 {[}41.98, 46.18{]} & -0.06 & 4.96 {[}4.39, 5.34{]} \\
  LCBv6 & P80$\rightarrow$S20 & 41.54 {[}40.46, 42.94{]} & -3.12 & 29.58 {[}28.63, 30.92{]} \\
  LCBv6 & P20$\rightarrow$S80 & 39.12 {[}37.60, 41.60{]} & -5.53 & 25.25 {[}19.85, 30.73{]} \\
  LCBv6 & S80$\rightarrow$P20 & 44.47 {[}44.08, 44.85{]} & -0.19 & 3.82 {[}3.44, 4.39{]} \\
  \midrule
  CodeContests & S20$\rightarrow$P80 & 41.15 {[}38.87, 42.38{]} & -1.43 & 6.32 {[}5.47, 7.23{]} \\
  CodeContests & P80$\rightarrow$S20 & 30.08 {[}28.91, 30.86{]} & -12.50 & 53.91 {[}53.32, 54.69{]} \\
  CodeContests & P20$\rightarrow$S80 & 34.18 {[}31.84, 35.55{]} & -8.40 & 35.87 {[}31.64, 42.58{]} \\
  CodeContests & S80$\rightarrow$P20 & 43.10 {[}41.41, 45.70{]} & +0.52 & 6.45 {[}5.47, 7.42{]} \\
  \bottomrule
  \end{tabular*}
  \vspace{1mm}
  \caption*{\phantomsection\label{app:tab-f4c}\textbf{Table F4c:} Fixed-composition coding order control at Step 100. Each schedule contains 100 updates and three matched seeds. Every seed-level comparison favors the Purified-last order.}
\end{table*}

\section{RQ4: Expression}
\label{app:expression}
\raggedbottom

RQ4 uses a favorable correction-transfer stress test rather than tracing the Standard OPSD target end to end: after training the student on verified local corrections, we test on a question-disjoint held-out set whether better target fit yields generation without privileged support.

Section~\ref{app:expression-construction} documents correction construction, Section~\ref{app:expression-protocol} specifies training and evaluation, Sections~\ref{app:expression-longitudinal}-\ref{app:expression-endpoints} report target fit, generation, and endpoint diagnostics, and Section~\ref{app:expression-controls} gives the matched controls.

\subsection{Correction construction}
\label{app:expression-construction}

We construct model-specific correction sets because the error states and valid continuations differ between Qwen3-4B and Qwen3-8B. The source pool contains mixed-outcome Thinking-ON OpenThoughts questions, for which the frozen Base produced at least one correct and one incorrect trajectory. GPT-5.6-Sol receives the problem, privileged solution, and matched correct and incorrect trajectories, identifies a local actionable error, and writes a repair instruction that does not reveal the final answer. From the exact error-containing prefix, the corresponding frozen Qwen3 teacher receives the privileged solution and repair instruction and generates four candidate continuations with an 8,192-token completion cap. We retain verifier-correct, non-truncated candidates, then apply a condition-blind self-containment audit with GPT-5.6-Sol. Without seeing the source condition, the judge labels each candidate self-contained, source-dependent, or invalid. The shortest retained self-contained continuation per question defines the model-specific correction target. Table~\hyperref[app:tab-g1]{G1} reports the filtering counts and the question-disjoint split.

\begin{table*}[!t]
  \centering
  \scriptsize
  \setlength{\tabcolsep}{5pt}
  \renewcommand{\arraystretch}{1.0}
  \begin{tabular}{@{}lrrrrrr@{}}
  \toprule
  \textbf{Model} & \textbf{Candidates} & \textbf{Self-contained} & \textbf{Source-dependent} & \textbf{Invalid} & \textbf{Selected} & \textbf{Train / held out} \\
  \midrule
  Qwen3-4B & 499 & 383 & 36 & 80 & 125 & 100 / 25 \\
  Qwen3-8B & 415 & 320 & 28 & 67 & 102 & 82 / 20 \\
  \bottomrule
  \end{tabular}
  \vspace{1mm}
  \caption*{\phantomsection\label{app:tab-g1}\textbf{Table G1:} Correction-candidate filtering and question-disjoint split. All candidates are verifier-correct and non-truncated. Counts precede shortest-per-question selection.}
\end{table*}

This construction is deliberately favorable: every retained target begins at an audited local error, reaches a verifier-correct answer, is non-truncated, and passes the self-containment gate. The experiment therefore tests whether training on available valid corrections supports autonomous expression on held-out questions. It does not estimate how often arbitrary model errors admit such a correction.

\subsection{Protocol}
\label{app:expression-protocol}

Across the main grids and matched controls, 50,032 local generations are evaluated: 44,536 main-grid and frozen-Base readouts, 4,296 full-dose Matched-SFT readouts, and 1,200 same-prefix-control readouts. An answer-equivalence screen accepts clear matches, and GPT-5.6-Sol assigns task-level labels to the remaining 17,178 responses using only the problem, reference final answer, extracted answer, and final 500 whitespace-delimited response tokens. No worked solution or condition metadata is shown. Claude Opus 4.8 independently evaluates a deterministic 500-case sample stratified by model, readout, arm, seed, and checkpoint, reaching 93.6\% agreement (95\% Wilson CI: 91.1--95.4\%). A blinded third pass resolves all 32 disagreements. Ambiguous responses count as incorrect. Unless a caption states otherwise, 95\% seed-bootstrap intervals enumerate all $3^3$ ordered resamples of the three prespecified training seeds and are descriptive. Stated contrasts use unrounded seed means.

RQ4 otherwise uses the LoRA and optimizer configuration in Table~\hyperref[app:tab-b1]{B1}, with a Thinking-ON teacher and student. Each run uses 48 updates with a global batch size of 32, a linear learning-rate schedule from $5\times10^{-6}$, and checkpoints at Steps 12, 24, 36, and 48. Standard OPSD rollouts use a 1,024-token cap, temperature 1.0, top-p 1.0, and no top-k truncation. Correction updates replace Standard OPSD with continuation-only teacher-forced NLL on up to 1,024 verified-correction tokens after the audited error prefix, with no loss on the prefix. All arms use three prespecified matched seeds. Tables~\hyperref[app:tab-g2a]{G2a}--\hyperref[app:tab-g2b]{G2b} list the batch schedules and evaluation conditions.

\begin{table*}[!t]
  \centering
  \scriptsize
  \setlength{\tabcolsep}{6pt}
  \renewcommand{\arraystretch}{1.0}
  \begin{tabular}{@{}ccccc@{}}
  \toprule
  \textbf{Arm} & \textbf{Total updates} & \textbf{Correction updates} & \textbf{Correction share} & \textbf{Scheduled examples} \\
  \midrule
  Standard & 48 & 0 & 0\% & 1,536 \\
  Correction25 & 48 & 12 & 25\% & 1,536 \\
  Correction50 & 48 & 24 & 50\% & 1,536 \\
  Correction100 & 48 & 48 & 100\% & 1,536 \\
  \bottomrule
  \end{tabular}
  \vspace{1mm}
  \caption*{\phantomsection\label{app:tab-g2a}\textbf{Table G2a:} Fixed homogeneous-batch schedules. Non-correction updates use Standard OPSD. Correction25 replaces updates 4, 8, \ldots, 48. Correction50 replaces 2, 4, \ldots, 48. Correction100 replaces every update. These one-indexed positions are shared across arms, seeds, and scales.}
\end{table*}

\begin{table*}[!t]
  \centering
  \scriptsize
  \setlength{\tabcolsep}{3pt}
  \renewcommand{\arraystretch}{1.0}
  \resizebox{\textwidth}{!}{%
  \begin{tabular}{@{}lll@{}}
  \toprule
  \textbf{Readout} & \textbf{Input} & \textbf{Evaluation} \\
  \midrule
  Correction-target fit & Exact audited error prefix; teacher-forced verified correction & Question-averaged held-out per-token NLL; Steps 12/24/36/48 \\
  Audited-prefix generation & Exact audited error prefix; no privileged support & Avg@8 accuracy; 16,384-token cap \\
  Original-question restart & Original problem prompt; no privileged support & Avg@8 accuracy; 16,384-token cap \\
  Explicit repair feedback & Error prefix + answer-free repair instruction & Step 48 Avg@8 accuracy; 16,384-token cap \\
  Correction onset & Error prefix + first 32 non-answer-revealing correction tokens & Step 48 Avg@8 accuracy; 16,384-token cap \\
  Broad mathematics & Original benchmark prompt; no privileged support & Step 48 Avg@12; 38,912-token cap; five-benchmark macro \\
  \bottomrule
  \end{tabular}%
  }
  \vspace{1mm}
  \caption*{\phantomsection\label{app:tab-g2b}\textbf{Table G2b:} RQ4 evaluation conditions. Qwen3-4B uses 25 held-out correction questions and Qwen3-8B uses 20. The 8B correction-onset condition excludes one onset that reveals the answer and therefore uses 19 questions. All other local conditions use the full held-out split. All evaluation generation uses temperature 1.0, top-p 1.0, and no top-k truncation. Local frozen-Base readouts use one fixed evaluation. Broad-mathematics Base rows average decoding seeds 0, 1, and 2.}
\end{table*}

\subsection{Longitudinal results}
\label{app:expression-longitudinal}

Tables~\hyperref[app:tab-g3a]{G3a}--\hyperref[app:tab-g3b]{G3b} give the longitudinal target-fit and unsupported-generation values.

\noindent\textbf{Target fit improves without dose-ordered unsupported use.} Greater correction dose consistently lowers held-out NLL, but audited-prefix and question-restart accuracy do not improve monotonically with dose. At Step 48, Correction100 trails Standard in three of four start-state-by-scale cells and exceeds it only by 0.42pp on the 8B question restart.

\begin{table*}[!t]
  \centering
  \scriptsize
  \setlength{\tabcolsep}{2pt}
  \renewcommand{\arraystretch}{0.95}
  \begin{tabular*}{\textwidth}{@{\extracolsep{\fill}}cccccccc@{}}
  \toprule
  & & \multicolumn{2}{c}{\textbf{Held-out correction-target NLL $\downarrow$}} & \multicolumn{2}{c}{\textbf{Audited-prefix Avg@8 (\%) $\uparrow$}} & \multicolumn{2}{c}{\textbf{Question-restart Avg@8 (\%) $\uparrow$}} \\
  \cmidrule(lr){3-4}\cmidrule(lr){5-6}\cmidrule(l){7-8}
  \textbf{Arm} & \textbf{Step} & \textbf{Mean {[}95\% CI{]}} & \textbf{$\Delta$} & \textbf{Mean {[}95\% CI{]}} & \textbf{$\Delta$} & \textbf{Mean {[}95\% CI{]}} & \textbf{$\Delta$} \\
  \midrule
  Frozen Base & 0 & 0.414 &  & 91.50 &  & 88.00 &  \\
  \midrule[0.3pt]
  Standard & 12 & 0.420 {[}0.420, 0.420{]} & +0.006 & 90.67 {[}89.54, 91.50{]} & -0.83 & 88.50 {[}86.37, 90.28{]} & +0.50 \\
   & 24 & 0.430 {[}0.429, 0.430{]} & +0.016 & 89.50 {[}87.61, 91.92{]} & -2.00 & 90.17 {[}87.87, 92.17{]} & +2.17 \\
   & 36 & 0.438 {[}0.438, 0.438{]} & +0.024 & 92.33 {[}90.93, 93.67{]} & +0.83 & 92.50 {[}90.47, 94.00{]} & +4.50 \\
   & 48 & 0.441 {[}0.441, 0.441{]} & +0.027 & 93.83 {[}92.43, 95.17{]} & +2.33 & 89.50 {[}88.04, 90.78{]} & +1.50 \\
  \midrule[0.3pt]
  Correction25 & 12 & 0.412 {[}0.412, 0.413{]} & -0.002 & 89.67 {[}88.83, 90.39{]} & -1.83 & 87.67 {[}85.65, 89.57{]} & -0.33 \\
   & 24 & 0.408 {[}0.407, 0.408{]} & -0.006 & 89.83 {[}89.00, 90.96{]} & -1.67 & 90.00 {[}88.83, 91.17{]} & +2.00 \\
   & 36 & 0.405 {[}0.404, 0.405{]} & -0.009 & 90.83 {[}89.72, 92.07{]} & -0.67 & 89.83 {[}89.22, 90.39{]} & +1.83 \\
   & 48 & 0.404 {[}0.404, 0.404{]} & -0.010 & 92.00 {[}91.61, 92.39{]} & +0.50 & 90.00 {[}88.54, 91.28{]} & +2.00 \\
  \midrule[0.3pt]
  Correction50 & 12 & 0.406 {[}0.406, 0.406{]} & -0.008 & 90.33 {[}88.93, 91.67{]} & -1.17 & 88.00 {[}87.22, 88.78{]} & +0.00 \\
   & 24 & 0.395 {[}0.395, 0.395{]} & -0.019 & 88.00 {[}85.08, 90.57{]} & -3.50 & 88.33 {[}88.11, 88.50{]} & +0.33 \\
   & 36 & 0.387 {[}0.387, 0.387{]} & -0.027 & 91.00 {[}90.61, 91.39{]} & -0.50 & 88.17 {[}87.33, 88.89{]} & +0.17 \\
   & 48 & 0.384 {[}0.384, 0.385{]} & -0.030 & 89.33 {[}88.11, 90.85{]} & -2.17 & 91.00 {[}88.58, 92.89{]} & +3.00 \\
  \midrule[0.3pt]
  Correction100 & 12 & 0.394 {[}0.394, 0.394{]} & -0.020 & 87.50 {[}86.72, 88.28{]} & -4.00 & 88.50 {[}88.11, 88.89{]} & +0.50 \\
   & 24 & 0.374 {[}0.373, 0.374{]} & -0.040 & 86.00 {[}84.26, 87.39{]} & -5.50 & 89.17 {[}88.33, 89.89{]} & +1.17 \\
   & 36 & 0.364 {[}0.364, 0.364{]} & -0.050 & 86.33 {[}85.50, 87.46{]} & -5.17 & 87.83 {[}87.61, 88.00{]} & -0.17 \\
   & 48 & 0.361 {[}0.361, 0.361{]} & -0.053 & 87.67 {[}86.43, 88.78{]} & -3.83 & 87.17 {[}85.33, 89.24{]} & -0.83 \\
  \bottomrule
  \end{tabular*}
  \vspace{1mm}
  \caption*{\phantomsection\label{app:tab-g3a}\textbf{Table G3a:} Qwen3-4B longitudinal results. Trained cells are mean {[}95\% seed-bootstrap CI{]}. $\Delta$ is relative to frozen Base.}
\end{table*}

\begin{table*}[!t]
  \centering
  \scriptsize
  \setlength{\tabcolsep}{2pt}
  \renewcommand{\arraystretch}{0.92}
  \begin{tabular*}{\textwidth}{@{\extracolsep{\fill}}cccccccc@{}}
  \toprule
  & & \multicolumn{2}{c}{\textbf{Held-out correction-target NLL $\downarrow$}} & \multicolumn{2}{c}{\textbf{Audited-prefix Avg@8 (\%) $\uparrow$}} & \multicolumn{2}{c}{\textbf{Question-restart Avg@8 (\%) $\uparrow$}} \\
  \cmidrule(lr){3-4}\cmidrule(lr){5-6}\cmidrule(l){7-8}
  \textbf{Arm} & \textbf{Step} & \textbf{Mean {[}95\% CI{]}} & \textbf{$\Delta$} & \textbf{Mean {[}95\% CI{]}} & \textbf{$\Delta$} & \textbf{Mean {[}95\% CI{]}} & \textbf{$\Delta$} \\
  \midrule
  Frozen Base & 0 & 0.418 &  & 86.25 &  & 87.50 &  \\
  \midrule[0.3pt]
  Standard & 12 & 0.416 {[}0.415, 0.416{]} & -0.003 & 89.38 {[}88.40, 90.35{]} & +3.12 & 91.25 {[}89.51, 93.43{]} & +3.75 \\
   & 24 & 0.415 {[}0.414, 0.415{]} & -0.004 & 90.21 {[}89.51, 90.98{]} & +3.96 & 89.17 {[}88.40, 89.86{]} & +1.67 \\
   & 36 & 0.415 {[}0.414, 0.415{]} & -0.004 & 88.54 {[}87.50, 89.95{]} & +2.29 & 89.58 {[}89.02, 90.00{]} & +2.08 \\
   & 48 & 0.415 {[}0.415, 0.415{]} & -0.003 & 88.54 {[}86.66, 90.57{]} & +2.29 & 90.21 {[}89.02, 91.47{]} & +2.71 \\
  \midrule[0.3pt]
  Correction25 & 12 & 0.413 {[}0.413, 0.413{]} & -0.005 & 88.96 {[}88.40, 89.38{]} & +2.71 & 90.42 {[}89.38, 91.82{]} & +2.92 \\
   & 24 & 0.408 {[}0.408, 0.408{]} & -0.010 & 87.92 {[}87.50, 88.48{]} & +1.67 & 88.96 {[}87.55, 90.00{]} & +1.46 \\
   & 36 & 0.405 {[}0.405, 0.405{]} & -0.013 & 87.29 {[}84.78, 90.17{]} & +1.04 & 89.38 {[}88.40, 90.35{]} & +1.88 \\
   & 48 & 0.404 {[}0.404, 0.404{]} & -0.014 & 87.92 {[}87.64, 88.12{]} & +1.67 & 89.58 {[}88.89, 90.35{]} & +2.08 \\
  \midrule[0.3pt]
  Correction50 & 12 & 0.411 {[}0.410, 0.411{]} & -0.008 & 87.08 {[}86.52, 87.50{]} & +0.83 & 87.71 {[}87.15, 88.12{]} & +0.21 \\
   & 24 & 0.403 {[}0.403, 0.403{]} & -0.015 & 88.75 {[}87.42, 89.86{]} & +2.50 & 90.21 {[}88.31, 91.74{]} & +2.71 \\
   & 36 & 0.399 {[}0.398, 0.399{]} & -0.020 & 87.08 {[}85.76, 88.70{]} & +0.83 & 91.25 {[}90.76, 91.74{]} & +3.75 \\
   & 48 & 0.397 {[}0.397, 0.397{]} & -0.022 & 87.08 {[}86.52, 87.50{]} & +0.83 & 89.58 {[}88.89, 90.35{]} & +2.08 \\
  \midrule[0.3pt]
  Correction100 & 12 & 0.405 {[}0.405, 0.405{]} & -0.013 & 87.08 {[}85.90, 88.34{]} & +0.83 & 89.17 {[}88.89, 89.38{]} & +1.67 \\
   & 24 & 0.394 {[}0.394, 0.394{]} & -0.024 & 87.08 {[}86.03, 87.99{]} & +0.83 & 90.62 {[}89.78, 91.25{]} & +3.12 \\
   & 36 & 0.387 {[}0.387, 0.387{]} & -0.031 & 86.04 {[}84.78, 87.23{]} & -0.21 & 88.96 {[}87.15, 91.06{]} & +1.46 \\
   & 48 & 0.385 {[}0.385, 0.385{]} & -0.033 & 86.88 {[}86.39, 87.36{]} & +0.62 & 90.62 {[}89.02, 92.45{]} & +3.12 \\
  \bottomrule
  \end{tabular*}
  \vspace{1mm}
  \caption*{\phantomsection\label{app:tab-g3b}\textbf{Table G3b:} Qwen3-8B longitudinal results under the same convention as Table~\hyperref[app:tab-g3a]{G3a}.}
\end{table*}

\subsection{Step-48 generation endpoints}
\label{app:expression-endpoints}

Tables~\hyperref[app:tab-g4a]{G4a}--\hyperref[app:tab-g4b]{G4b} report the complete Step-48 endpoints plotted in Figure~\ref{fig:correction-expression}. Tables~\hyperref[app:tab-g5a]{G5a}--\hyperref[app:tab-g5b]{G5b} give the broad-mathematics results by benchmark.

\noindent\textbf{Truncation falls without broad accuracy gains.} Truncation is the only broad generation metric that improves monotonically with correction dose at both scales. Correction100 is 7.02pp below Standard at 4B and 3.39pp below Standard at 8B on the macro average. The per-benchmark breakdown shows that the broad decline is not an artifact of one benchmark. The size varies across datasets, but no correction dose establishes a consistent improvement over the matched frozen Base or Standard across all five benchmarks.

\begin{table*}[!t]
  \centering
  \scriptsize
  \setlength{\tabcolsep}{2pt}
  \renewcommand{\arraystretch}{1.0}
  \begin{tabular*}{\textwidth}{@{\extracolsep{\fill}}cccccc@{}}
  \toprule
  \textbf{Step-48 metric} & \textbf{Frozen Base} & \textbf{Standard} & \textbf{Correction25} & \textbf{Correction50} & \textbf{Correction100} \\
  \midrule
  Held-out correction-target NLL $\downarrow$ & 0.414 & 0.441 {[}0.441, 0.441{]} & 0.404 {[}0.404, 0.404{]} & 0.384 {[}0.384, 0.385{]} & 0.361 {[}0.361, 0.361{]} \\
  Exact audited prefix, Avg@8 $\uparrow$ & 91.50 & 93.83 {[}92.43, 95.17{]} & 92.00 {[}91.61, 92.39{]} & 89.33 {[}88.11, 90.85{]} & 87.67 {[}86.43, 88.78{]} \\
  Original-question restart, Avg@8 $\uparrow$ & 88.00 & 89.50 {[}88.04, 90.78{]} & 90.00 {[}88.54, 91.28{]} & 91.00 {[}88.58, 92.89{]} & 87.17 {[}85.33, 89.24{]} \\
  Explicit repair feedback, Avg@8 $\uparrow$ & 96.50 & 94.33 {[}94.11, 94.50{]} & 95.50 {[}95.11, 95.89{]} & 92.50 {[}90.47, 94.00{]} & 91.17 {[}88.37, 93.85{]} \\
  First 32 correction tokens, Avg@8 $\uparrow$ & 91.00 & 89.50 {[}88.72, 90.28{]} & 88.83 {[}88.22, 89.39{]} & 86.33 {[}86.11, 86.50{]} & 85.83 {[}84.54, 86.89{]} \\
  Five-benchmark Avg@12 $\uparrow$ & 60.42 & 60.19 {[}59.72, 60.68{]} & 58.89 {[}58.04, 59.58{]} & 56.53 {[}56.13, 56.83{]} & 53.16 {[}52.84, 53.43{]} \\
  Five-benchmark truncation (\%) $\downarrow$ & 2.58 & 7.81 {[}7.33, 8.35{]} & 2.78 {[}2.62, 2.92{]} & 1.18 {[}1.03, 1.33{]} & 0.20 {[}0.18, 0.22{]} \\
  \bottomrule
  \end{tabular*}
  \vspace{1mm}
  \caption*{\phantomsection\label{app:tab-g4a}\textbf{Table G4a:} Qwen3-4B Step-48 endpoints. Trained cells are three-seed means {[}95\% seed-bootstrap CI{]}.}
\end{table*}

\begin{table*}[!t]
  \centering
  \scriptsize
  \setlength{\tabcolsep}{2pt}
  \renewcommand{\arraystretch}{1.0}
  \begin{tabular*}{\textwidth}{@{\extracolsep{\fill}}cccccc@{}}
  \toprule
  \textbf{Step-48 metric} & \textbf{Frozen Base} & \textbf{Standard} & \textbf{Correction25} & \textbf{Correction50} & \textbf{Correction100} \\
  \midrule
  Held-out correction-target NLL $\downarrow$ & 0.418 & 0.415 {[}0.415, 0.415{]} & 0.404 {[}0.404, 0.404{]} & 0.397 {[}0.397, 0.397{]} & 0.385 {[}0.385, 0.385{]} \\
  Exact audited prefix, Avg@8 $\uparrow$ & 86.25 & 88.54 {[}86.66, 90.57{]} & 87.92 {[}87.64, 88.12{]} & 87.08 {[}86.52, 87.50{]} & 86.88 {[}86.39, 87.36{]} \\
  Original-question restart, Avg@8 $\uparrow$ & 87.50 & 90.21 {[}89.02, 91.47{]} & 89.58 {[}88.89, 90.35{]} & 89.58 {[}88.89, 90.35{]} & 90.62 {[}89.02, 92.45{]} \\
  Explicit repair feedback, Avg@8 $\uparrow$ & 93.12 & 93.54 {[}92.64, 94.59{]} & 92.29 {[}91.88, 92.85{]} & 93.33 {[}92.15, 94.59{]} & 92.29 {[}91.88, 92.85{]} \\
  First 32 correction tokens, Avg@8 $\uparrow$ & 88.16 & 89.47 {[}88.96, 89.99{]} & 87.72 {[}84.84, 89.99{]} & 87.50 {[}86.84, 88.39{]} & 87.94 {[}86.10, 89.70{]} \\
  Five-benchmark Avg@12 $\uparrow$ & 62.52 & 63.49 {[}62.92, 64.08{]} & 62.48 {[}62.37, 62.57{]} & 62.56 {[}61.58, 63.29{]} & 60.10 {[}59.56, 60.69{]} \\
  Five-benchmark truncation (\%) $\downarrow$ & 4.79 & 7.64 {[}7.54, 7.75{]} & 4.68 {[}4.60, 4.77{]} & 2.90 {[}2.61, 3.25{]} & 1.33 {[}1.22, 1.44{]} \\
  \bottomrule
  \end{tabular*}
  \vspace{1mm}
  \caption*{\phantomsection\label{app:tab-g4b}\textbf{Table G4b:} Qwen3-8B Step-48 endpoints under the same convention as Table~\hyperref[app:tab-g4a]{G4a}.}
\end{table*}

\begin{table*}[!t]
  \centering
  \scriptsize
  \setlength{\tabcolsep}{2pt}
  \renewcommand{\arraystretch}{1.0}
  \resizebox{\textwidth}{!}{%
  \begin{tabular}{@{}ccccccc@{}}
  \toprule
  \textbf{Arm} & \textbf{AIME24} & \textbf{AIME25} & \textbf{AIME26} & \textbf{HMMT Nov25} & \textbf{HMMT Feb26} & \textbf{Macro} \\
  \midrule
  Frozen Base (3-seed mean) & 73.33 & 65.74 & 66.48 & 55.37 & 41.16 & 60.42 \\
  Standard & 74.54 {[}73.26, 75.65{]} & 64.91 {[}64.23, 65.75{]} & 64.72 {[}64.13, 65.22{]} & 54.17 {[}52.74, 55.79{]} & 42.59 {[}41.58, 43.75{]} & 60.19 {[}59.72, 60.68{]} \\
  Correction25 & 73.61 {[}71.24, 75.59{]} & 63.06 {[}62.19, 63.93{]} & 63.52 {[}62.05, 64.66{]} & 51.76 {[}50.36, 53.19{]} & 42.51 {[}41.74, 43.13{]} & 58.89 {[}58.04, 59.58{]} \\
  Correction50 & 70.65 {[}68.56, 72.48{]} & 60.46 {[}60.12, 60.77{]} & 61.67 {[}60.54, 62.50{]} & 50.56 {[}48.66, 52.36{]} & 39.31 {[}38.44, 40.42{]} & 56.53 {[}56.13, 56.83{]} \\
  Correction100 & 67.96 {[}67.40, 68.49{]} & 56.57 {[}55.52, 57.69{]} & 58.15 {[}57.84, 58.49{]} & 46.94 {[}45.82, 47.78{]} & 36.20 {[}34.81, 37.78{]} & 53.16 {[}52.84, 53.43{]} \\
  \bottomrule
  \end{tabular}%
  }
  \vspace{1mm}
  \caption*{\phantomsection\label{app:tab-g5a}\textbf{Table G5a:} Qwen3-4B Step-48 results by benchmark. Trained cells are three-seed means {[}95\% seed-bootstrap CI{]}.}
  
  \resizebox{\textwidth}{!}{%
  \begin{tabular}{@{}ccccccc@{}}
  \toprule
  \textbf{Arm} & \textbf{AIME24} & \textbf{AIME25} & \textbf{AIME26} & \textbf{HMMT Nov25} & \textbf{HMMT Feb26} & \textbf{Macro} \\
  \midrule
  Frozen Base (3-seed mean) & 75.00 & 68.98 & 69.26 & 54.26 & 45.12 & 62.52 \\
  Standard & 74.91 {[}74.19, 75.50{]} & 68.52 {[}67.68, 69.42{]} & 69.44 {[}68.30, 70.69{]} & 58.61 {[}57.34, 60.17{]} & 45.96 {[}44.92, 47.09{]} & 63.49 {[}62.92, 64.08{]} \\
  Correction25 & 74.44 {[}73.79, 75.10{]} & 67.59 {[}66.57, 68.75{]} & 67.69 {[}66.97, 68.27{]} & 58.06 {[}56.63, 59.68{]} & 44.61 {[}43.51, 45.69{]} & 62.48 {[}62.37, 62.57{]} \\
  Correction50 & 75.74 {[}73.84, 77.32{]} & 66.48 {[}65.86, 66.94{]} & 67.59 {[}66.00, 68.83{]} & 57.59 {[}56.29, 59.12{]} & 45.37 {[}44.57, 45.96{]} & 62.56 {[}61.58, 63.29{]} \\
  Correction100 & 73.70 {[}73.33, 74.20{]} & 63.06 {[}62.68, 63.33{]} & 66.11 {[}65.89, 66.33{]} & 54.63 {[}53.95, 55.47{]} & 43.01 {[}41.94, 44.12{]} & 60.10 {[}59.56, 60.69{]} \\
  \bottomrule
  \end{tabular}%
  }
  \vspace{1mm}
  \caption*{\phantomsection\label{app:tab-g5b}\textbf{Table G5b:} Qwen3-8B Step-48 results by benchmark. Trained cells are three-seed means {[}95\% seed-bootstrap CI{]}.}
\end{table*}

\subsection{Matched controls}
\label{app:expression-controls}

We test whether correction-training degradation is a generic consequence of repeated SFT on a small dataset. Matched-SFT uses the same questions, batch size, update count, fixed schedule, and row-wise target-token cap as Correction100, but replaces each correction with the shortest verifier-correct frozen-Base trajectory from the original question. An additional Matched-SFT50 control replaces the same 24 batches as Correction50 with the corresponding Base-correct targets while preserving question order, batch position, and the row-wise target-token cap. All arms use three prespecified seeds. In correction training, the audited error prefix is prompt context and is excluded from the response mask, so its tokens receive no loss. We then isolate that conditioning context with a 4B same-prefix control. From the 100 Correction100 training prefixes, the frozen Base produces a verifier-correct, non-truncated recovery for 76 usable prefixes. We preserve the Correction100 order, repeat the 76-row sequence to the same 1,536 examples, and match the Base recovery and teacher correction target lengths row by row. The two arms therefore differ only in continuation content. The dose-matched and same-prefix controls in Tables~\hyperref[app:tab-g6c]{G6c}--\hyperref[app:tab-g6d]{G6d} are Qwen3-4B-only diagnostics. Table~\hyperref[app:tab-g6b]{G6b} reports the Qwen3-8B Matched-SFT control.

\begin{table*}[!t]
  \centering
  \scriptsize
  \setlength{\tabcolsep}{2pt}
  \renewcommand{\arraystretch}{1.0}
  \begin{tabular*}{\textwidth}{@{\extracolsep{\fill}}ccccc@{}}
  \toprule
  \textbf{Qwen3-4B metric} & \textbf{Standard} & \textbf{Correction100} & \textbf{Matched-SFT} & \textbf{C100 - Matched} \\
  \midrule
  Held-out correction-target NLL $\downarrow$ & 0.441 {[}0.441, 0.441{]} & 0.361 {[}0.361, 0.361{]} & 0.416 {[}0.416, 0.417{]} & -0.055 {[}-0.056, -0.055{]} \\
  Audited-prefix accuracy & 93.83 {[}92.43, 95.17{]} & 87.67 {[}86.43, 88.78{]} & 88.33 {[}87.50, 89.46{]} & -0.67 {[}-3.02, +1.28{]} \\
  Question-restart accuracy & 89.50 {[}88.04, 90.78{]} & 87.17 {[}85.33, 89.24{]} & 88.67 {[}85.30, 91.57{]} & -1.50 {[}-4.31, +0.78{]} \\
  Repair-instruction accuracy & 94.33 {[}94.11, 94.50{]} & 91.17 {[}88.37, 93.85{]} & 94.83 {[}94.22, 95.39{]} & -3.67 {[}-7.03, -0.37{]} \\
  32-token-onset accuracy & 89.50 {[}88.72, 90.28{]} & 85.83 {[}84.54, 86.89{]} & 86.33 {[}84.93, 87.67{]} & -0.50 {[}-3.13, +1.96{]} \\
  Five-benchmark Avg@12 & 60.19 {[}59.72, 60.68{]} & 53.16 {[}52.84, 53.43{]} & 59.71 {[}59.29, 60.27{]} & -6.55 {[}-7.06, -6.02{]} \\
  Truncation (\%) $\downarrow$ & 7.81 {[}7.33, 8.35{]} & 0.20 {[}0.18, 0.22{]} & 3.00 {[}2.79, 3.26{]} & -2.80 {[}-3.04, -2.61{]} \\
  \bottomrule
  \end{tabular*}
  \vspace{1mm}
  \caption*{\phantomsection\label{app:tab-g6a}\textbf{Table G6a:} Qwen3-4B Matched-SFT control. Entries are three-seed means {[}95\% seed-bootstrap CI{]}. The final column uses paired seeds.}
\end{table*}

\begin{table*}[!t]
  \centering
  \scriptsize
  \setlength{\tabcolsep}{2pt}
  \renewcommand{\arraystretch}{1.0}
  \begin{tabular*}{\textwidth}{@{\extracolsep{\fill}}ccccc@{}}
  \toprule
  \textbf{Qwen3-8B metric} & \textbf{Standard} & \textbf{Correction100} & \textbf{Matched-SFT} & \textbf{C100 - Matched} \\
  \midrule
  Held-out correction-target NLL $\downarrow$ & 0.415 {[}0.415, 0.415{]} & 0.385 {[}0.385, 0.385{]} & 0.419 {[}0.419, 0.419{]} & -0.034 {[}-0.034, -0.033{]} \\
  Audited-prefix accuracy & 88.54 {[}86.66, 90.57{]} & 86.88 {[}86.39, 87.36{]} & 86.88 {[}85.41, 88.34{]} & +0.00 {[}-0.98, +0.98{]} \\
  Question-restart accuracy & 90.21 {[}89.02, 91.47{]} & 90.62 {[}89.02, 92.45{]} & 88.12 {[}86.79, 89.24{]} & +2.50 {[}+1.03, +3.97{]} \\
  Repair-instruction accuracy & 93.54 {[}92.64, 94.59{]} & 92.29 {[}91.88, 92.85{]} & 93.33 {[}92.77, 93.75{]} & -1.04 {[}-1.88, +0.08{]} \\
  32-token-onset accuracy & 89.47 {[}88.96, 89.99{]} & 87.94 {[}86.10, 89.70{]} & 86.18 {[}85.30, 86.84{]} & +1.75 {[}+0.80, +2.86{]} \\
  Five-benchmark Avg@12 & 63.49 {[}62.92, 64.08{]} & 60.10 {[}59.56, 60.69{]} & 63.38 {[}62.98, 63.71{]} & -3.28 {[}-3.88, -2.83{]} \\
  Truncation (\%) $\downarrow$ & 7.64 {[}7.54, 7.75{]} & 1.33 {[}1.22, 1.44{]} & 3.94 {[}3.76, 4.10{]} & -2.61 {[}-2.76, -2.51{]} \\
  \bottomrule
  \end{tabular*}
  \vspace{1mm}
  \caption*{\phantomsection\label{app:tab-g6b}\textbf{Table G6b:} Qwen3-8B Matched-SFT control under the same convention as Table~\hyperref[app:tab-g6a]{G6a}.}
\end{table*}

\begin{table*}[!t]
  \centering
  \scriptsize
  \setlength{\tabcolsep}{2pt}
  \renewcommand{\arraystretch}{1.0}
  \begin{tabular*}{\textwidth}{@{\extracolsep{\fill}}cccccc@{}}
  \toprule
  \textbf{Qwen3-4B metric} & \textbf{Frozen Base} & \textbf{Standard} & \textbf{Correction50} & \textbf{Matched-SFT50} & \textbf{Matched - C50} \\
  \midrule
  Five-benchmark Avg@12 & 60.42 {[}60.33, 60.51{]} & 60.19 {[}59.62, 60.84{]} & 56.53 {[}55.95, 56.85{]} & 60.68 {[}60.28, 60.99{]} & +4.15 {[}+3.97, +4.32{]} \\
  Truncation (\%) $\downarrow$ & 2.58 {[}2.56, 2.61{]} & 7.81 {[}7.24, 8.55{]} & 1.18 {[}0.98, 1.36{]} & 3.49 {[}3.21, 3.76{]} & +2.31 {[}+2.23, +2.40{]} \\
  \bottomrule
  \end{tabular*}
  \phantomsection\label{app:tab-g6c}
  \vspace{1mm}
  \caption*{\textbf{Table G6c:} Qwen3-4B dose-matched 50\% control. Entries are three-seed means with seed ranges. The final column is the paired seed difference.}
\end{table*}

\begin{table*}[!t]
  \centering
  \scriptsize
  \setlength{\tabcolsep}{2pt}
  \renewcommand{\arraystretch}{1.0}
  \begin{tabular*}{\textwidth}{@{\extracolsep{\fill}}ccccc@{}}
  \toprule
  \textbf{Arm} & \textbf{Held-out correction-target NLL $\downarrow$} & \textbf{Audited-prefix accuracy} & \textbf{Five-benchmark Avg@12} & \textbf{Truncation $\downarrow$} \\
  \midrule
  Base-recovery SFT & 0.418 {[}0.418, 0.418{]} & 87.50 {[}86.22, 88.96{]} & 59.54 {[}58.95, 59.90{]} & 3.21 {[}2.94, 3.38{]} \\
  Privileged-correction subset & 0.363 {[}0.362, 0.363{]} & 85.17 {[}83.65, 86.39{]} & 53.17 {[}52.66, 53.77{]} & 0.22 {[}0.11, 0.38{]} \\
  \bottomrule
  \end{tabular*}
  \phantomsection\label{app:tab-g6d}
  \vspace{1mm}
  \caption*{\textbf{Table G6d:} Qwen3-4B same-prefix, length-matched control. The two arms share identical error-prefix conditioning and row-matched target lengths. Local accuracy uses the common 25-question held-out audited-prefix evaluator rather than the 76-prefix training subset. Entries are three-seed means. Audited-prefix brackets are 95\% seed-bootstrap intervals, and the other brackets are seed ranges.}
\end{table*}

\vspace{1mm}
\noindent\textbf{Matched controls localize the 4B decline to the tested continuation.} In the same-prefix control, Base-recovery SFT reaches 87.50\% local accuracy versus 85.17\% for the privileged-correction subset, a +2.33pp paired-seed difference {[}0.37, 3.89{]}. At the 50\% replacement dose, Matched-SFT50 remains near Base at 60.68 Avg@12 and 3.49\% truncation, whereas Correction50 reaches 56.53 and 1.18\%. Matching the replacement count and positions does not reproduce either the large accuracy decline or the strong termination shift. At full dose, Matched-SFT changes five-benchmark Avg@12 by -0.71pp {[}-1.18, +0.09{]} at 4B and +0.86 {[}+0.17, +1.55{]} at 8B. Correction100 instead changes it by -7.25 {[}-7.69, -6.94{]} and -2.42 {[}-3.28, -1.56{]}, showing that generic repeated small-data SFT is insufficient to explain the decline. The stricter same-prefix comparison in Table~\hyperref[app:tab-g6d]{G6d} holds both the error prefix and target length fixed and therefore localizes the additional broad degradation and stronger termination change to the tested correction continuation rather than to the error prefix alone.

\noindent\begin{minipage}{\columnwidth}
Overall, correction dose reliably improves held-out target fit and reduces truncation, but neither unsupported nor cue-supported generation improves in a dose-ordered manner.
\end{minipage}

\section{Discussion diagnostics}
\label{app:discussion}

\subsection{Protocol}

The diagnostics in Tables~\hyperref[app:tab-h1a]{H1a}--\hyperref[app:tab-h1f]{H1f} use reference-free Thinking-ON evaluation: Avg@12 over AIME 2024, AIME 2025, AIME 2026, HMMT November 2025, and HMMT February 2026, with a 38,912-token cap. Unless a diagnostic varies it, training uses a Qwen3-4B student, a frozen same-size Thinking-ON teacher with the privileged solution, Thinking-ON student rollouts, LoRA rank 64, global batch size 32, learning rate $5\times10^{-6}$, 1,024-token rollouts, and 100 updates. Tables H1a and H1e average three prespecified training seeds and report the seed range. Tables H1b, H1c, H1d, and H1f use one training seed. All six diagnostics use the corresponding three-decoding-seed Base mean as the common anchor, including the single-training-seed diagnostics. Only the parameterization diagnostic (Table H1a) appears in the main text, as Figure~\ref{fig:discussion-diagnostics}c. Figure~\ref{fig:interface-factorial} remains the primary factorial.

\noindent\textbf{Diagnostic-specific variants.} Table H1a compares Standard and Purified OPSD under LoRA and full-parameter training on the 3,200-example Hendrycks MATH set. The full-parameter BF16 FSDP runs use learning rate $10^{-6}$, making this a practical parameterization control. Table H1b varies the per-vocabulary-summand clip for Standard OPSD on the same set: 0 gives exact FKL, and 0.05 is the released default. Table H1c compares Standard, Purified, $\beta$-target, and $\beta$-target with return-to-go variants on the 3,200-example OpenThoughts set. Purified uses a centered PMI target with JSD loss. The $\beta$-OPSD variant uses geometric mixing, with $\gamma=0.99$ for return-to-go. Table H1d varies only the maximum training-response length among 1K, 4K, 8K, and 16K tokens on OpenThoughts. Evaluation retains the 38,912-token cap. Table H1e compares a privileged-solution 1.7B self-teacher with reference-free 4B and 8B teachers for a Qwen3-1.7B student on Hendrycks MATH. Because teacher context and scale change jointly, this is not a pure teacher-size ablation. Table H1f routes Hendrycks MATH Level 5 examples by Base competence (0/12, 1--5/12, 6--11/12, or 12/12 correct) and trains Standard or Purified OPSD for 48 updates. Comparisons are interpretable within each table because corpora, model scales, and checkpoint grids differ across diagnostics.

\noindent\textbf{Action diagnostics.} The action-selection audit uses a Qwen3-4B model and an 80-question OpenThoughts Frontier cohort, pooling Standard and Purified Steps 0, 24, and 36. It compares oracle, privileged-target, matched-random, and learned-critic selection over 135 branch events from 57 questions. Candidate actions are the policy and target top-1 tokens, and critic evaluation uses five-fold problem-grouped cross-validation. The single-seed sparse-replay pilot compares Oracle, FKL, and three matched-random replay sets using one selected token for each of 33 training events per arm. Fold 0 holds out 30 states from 12 questions, with four question-start samples per question and a 16K response cap.

\subsection{Diagnostic values}
\label{app:diagnostic-values}

\begin{table*}[!t]
  \centering
  \scriptsize
  \setlength{\tabcolsep}{2pt}
  \renewcommand{\arraystretch}{1.0}
  \resizebox{\textwidth}{!}{%
  \begin{tabular}{@{}ccccccc@{}}
  \toprule
  \textbf{Condition} & \textbf{Step 20} & \textbf{Step 40} & \textbf{Step 60} & \textbf{Step 80} & \textbf{Step 100} & \textbf{Final $\Delta$ Base} \\
  \midrule
  Frozen Base & 60.42 & 60.42 & 60.42 & 60.42 & 60.42 & +0.00 \\
  Standard, LoRA & 61.50 {[}60.48, 62.08{]} & 59.52 {[}59.23, 59.68{]} & 57.92 {[}57.55, 58.44{]} & 58.33 {[}57.96, 58.94{]} & 58.09 {[}57.29, 59.05{]} & -2.33 \\
  Standard, full parameter & 60.47 {[}60.23, 60.75{]} & 58.97 {[}58.57, 59.53{]} & 58.10 {[}57.14, 58.96{]} & 57.84 {[}57.35, 58.37{]} & 56.80 {[}56.23, 57.35{]} & -3.62 \\
  Purified, LoRA & 61.67 {[}60.97, 62.26{]} & 60.98 {[}60.16, 61.52{]} & 61.25 {[}60.47, 61.89{]} & 61.69 {[}60.83, 62.21{]} & 62.35 {[}61.99, 62.76{]} & +1.93 \\
  Purified, full parameter & 60.56 {[}59.85, 61.14{]} & 60.82 {[}60.56, 61.05{]} & 60.66 {[}60.22, 61.03{]} & 60.59 {[}59.52, 61.30{]} & 61.00 {[}60.85, 61.20{]} & +0.58 \\
  \bottomrule
  \end{tabular}%
  }
  \vspace{1mm}
  \caption*{\phantomsection\label{app:tab-h1a}\textbf{Table H1a:} Parameterization (Figure~\ref{fig:discussion-diagnostics}c). Values are three-seed mean five-benchmark Avg@12 percentages. Brackets give the seed range shaded in Figure~\ref{fig:discussion-diagnostics}c, and $\Delta$ is an absolute difference in pp from the three-decoding-seed Base mean.}
\vspace{2mm}
  \centering
  \scriptsize
  \setlength{\tabcolsep}{2pt}
  \renewcommand{\arraystretch}{1.0}
  \begin{tabular*}{\textwidth}{@{\extracolsep{\fill}}ccccccc@{}}
  \toprule
  \textbf{Condition} & \textbf{Step 20} & \textbf{Step 40} & \textbf{Step 60} & \textbf{Step 80} & \textbf{Step 100} & \textbf{Final $\Delta$ Base} \\
  \midrule
  Frozen Base & 60.42 & 60.42 & 60.42 & 60.42 & 60.42 & +0.00 \\
  Exact FKL (clip 0) & 58.21 & 54.71 & 55.94 & 51.72 & 45.20 & -15.22 \\
  Clip 0.01 & 60.97 & 55.81 & 54.65 & 55.68 & 56.43 & -3.99 \\
  Clip 0.02 & 61.29 & 56.31 & 52.71 & 52.04 & 52.76 & -7.66 \\
  Standard (clip 0.05) & 60.48 & 59.68 & 57.55 & 58.94 & 59.05 & -1.37 \\
  Clip 0.10 & 58.90 & 57.62 & 57.04 & 57.87 & 57.23 & -3.19 \\
  \bottomrule
  \end{tabular*}
  \vspace{1mm}
  \caption*{\phantomsection\label{app:tab-h1b}\textbf{Table H1b:} Clipping sweep. Values are single-training-seed five-benchmark Avg@12 percentages. $\Delta$ uses the common three-decoding-seed Base anchor of 60.42\%.}
\vspace{2mm}
  \centering
  \scriptsize
  \setlength{\tabcolsep}{2pt}
  \renewcommand{\arraystretch}{1.0}
  \begin{tabular*}{\textwidth}{@{\extracolsep{\fill}}ccccccc@{}}
  \toprule
  \textbf{Condition} & \textbf{Step 20} & \textbf{Step 40} & \textbf{Step 60} & \textbf{Step 80} & \textbf{Step 100} & \textbf{Final $\Delta$ Base} \\
  \midrule
  Frozen Base & 60.42 & 60.42 & 60.42 & 60.42 & 60.42 & +0.00 \\
  Standard & 60.05 & 58.88 & 57.62 & 57.50 & 56.63 & -3.79 \\
  Purified & 61.55 & 61.53 & 61.06 & 60.34 & 59.95 & -0.47 \\
  $\beta$ target, local credit & 57.96 & 55.59 & 56.94 & 57.16 & 56.03 & -4.39 \\
  $\beta$ target + return-to-go & 60.39 & 58.17 & 58.05 & 56.61 & 53.25 & -7.17 \\
  \bottomrule
  \end{tabular*}
  \vspace{1mm}
  \caption*{\phantomsection\label{app:tab-h1c}\textbf{Table H1c:} Target and credit variants. Values are single-training-seed five-benchmark Avg@12 percentages. $\Delta$ uses the common three-decoding-seed Base anchor of 60.42\%.}
\vspace{2mm}
  \centering
  \scriptsize
  \setlength{\tabcolsep}{2pt}
  \renewcommand{\arraystretch}{1.0}
  \begin{tabular*}{\textwidth}{@{\extracolsep{\fill}}cccccccc@{}}
  \toprule
  \textbf{Condition} & \textbf{Step 10} & \textbf{Step 20} & \textbf{Step 40} & \textbf{Step 60} & \textbf{Step 80} & \textbf{Step 100} & \textbf{Final $\Delta$ Base} \\
  \midrule
  Frozen Base & 60.42 & 60.42 & 60.42 & 60.42 & 60.42 & 60.42 & +0.00 \\
  Standard, 1K & --- & 60.05 & 58.88 & 57.62 & 57.50 & 56.63 & -3.79 \\
  Standard, 4K & 61.02 & 60.70 & 56.51 & 55.96 & 55.01 & 55.03 & -5.39 \\
  Standard, 8K & 59.24 & 60.37 & 57.90 & 56.56 & 54.80 & 53.08 & -7.34 \\
  Standard, 16K & 59.44 & 59.97 & 57.78 & 55.59 & 55.51 & 55.22 & -5.20 \\
  \bottomrule
  \end{tabular*}
  \vspace{1mm}
  \caption*{\phantomsection\label{app:tab-h1d}\textbf{Table H1d:} Training-rollout horizon. Values are single-training-seed five-benchmark Avg@12 percentages. $\Delta$ uses the common three-decoding-seed Base anchor of 60.42\%.}
\vspace{2mm}
  \centering
  \scriptsize
  \setlength{\tabcolsep}{2pt}
  \renewcommand{\arraystretch}{1.0}
  \resizebox{\textwidth}{!}{%
  \begin{tabular}{@{}ccccccc@{}}
  \toprule
  \textbf{Condition} & \textbf{Step 20} & \textbf{Step 40} & \textbf{Step 60} & \textbf{Step 80} & \textbf{Step 100} & \textbf{Final $\Delta$ Base} \\
  \midrule
  Frozen Base & 36.26 & 36.26 & 36.26 & 36.26 & 36.26 & +0.00 \\
  Privileged-solution self-teacher & 38.44 {[}37.82, 39.38{]} & 36.15 {[}35.86, 36.36{]} & 34.90 {[}33.98, 35.60{]} & 36.14 {[}35.66, 36.64{]} & 34.12 {[}33.54, 35.06{]} & -2.14 \\
  Reference-free Qwen3-4B teacher & 37.45 {[}36.51, 38.07{]} & 35.74 {[}35.09, 36.35{]} & 33.38 {[}32.36, 34.07{]} & 33.07 {[}32.85, 33.47{]} & 33.60 {[}32.78, 34.04{]} & -2.67 \\
  Reference-free Qwen3-8B teacher & 36.97 {[}36.50, 37.34{]} & 35.12 {[}34.84, 35.29{]} & 32.80 {[}32.07, 33.30{]} & 32.46 {[}31.81, 33.49{]} & 32.64 {[}32.17, 33.15{]} & -3.62 \\
  \bottomrule
  \end{tabular}%
  }
  \vspace{1mm}
  \caption*{\phantomsection\label{app:tab-h1e}\textbf{Table H1e:} Teacher context and scale. Values are three-seed mean five-benchmark Avg@12 percentages. Brackets give the seed range, and $\Delta$ is an absolute difference in pp from the three-decoding-seed Base mean.}

  \vspace{2mm}
  \begin{tabular*}{\textwidth}{@{\extracolsep{\fill}}ccccccc@{}}
  \toprule
  \textbf{Condition} & \textbf{Step 6} & \textbf{Step 12} & \textbf{Step 24} & \textbf{Step 36} & \textbf{Step 48} & \textbf{Final $\Delta$ Base} \\
  \midrule
  Frozen Base & 60.42 & 60.42 & 60.42 & 60.42 & 60.42 & +0.00 \\
  Unsolved (n=129), Standard & 59.75 & 59.97 & 62.11 & 58.64 & 58.71 & -1.71 \\
  Unsolved (n=129), Purified & 61.47 & 61.86 & 61.56 & 60.26 & 61.12 & +0.70 \\
  Frontier (n=80), Standard & 60.91 & 60.56 & 61.24 & 61.57 & 60.00 & -0.42 \\
  Frontier (n=80), Purified & 59.59 & 60.51 & 61.27 & 60.15 & 61.15 & +0.73 \\
  High non-saturated (n=243), Standard & 60.43 & 61.48 & 62.61 & 61.64 & 59.19 & -1.23 \\
  High non-saturated (n=243), Purified & 60.53 & 60.17 & 60.87 & 61.61 & 60.43 & +0.01 \\
  Saturated (n=1,790), Standard & 60.13 & 60.40 & 60.24 & 58.87 & 56.85 & -3.57 \\
  Saturated (n=1,790), Purified & 59.66 & 60.49 & 59.45 & 59.86 & 59.98 & -0.44 \\
  \bottomrule
  \end{tabular*}
  \vspace{1mm}
  \caption*{\phantomsection\label{app:tab-h1f}\textbf{Table H1f:} Training-data routing. Values are single-training-seed five-benchmark Avg@12 percentages. $\Delta$ uses the common three-decoding-seed Base anchor of 60.42\%.}
\end{table*}

The controls alter trajectories and sometimes improve early checkpoints, but every Standard or teacher-variation endpoint finishes below Base. Purified sometimes ends modestly above Base, with its most repeatable advantage being retention relative to Standard. Changing parameterization, clipping, rollout horizon, teacher context and scale jointly, or static competence routing does not prevent Standard\textquotesingle s late decline.

\subsection{Action selection}
\label{app:diagnostic-selection}

The action-selection audit pools six Qwen3-4B source cells from the 80-question OpenThoughts Frontier cohort: Standard and Purified targets at their shared Step 0 and matched Step-24 and Step-36 checkpoints. At 135 top-FKL branch events from 57 problems, we force either the current-policy top-1 token or the corresponding OPSD-target top-1 token, remove the reference, and draw four continuations per action at temperature 1.0, top-p 1.0, with top-k disabled and a 38,912-token response cap. Exact-answer success averaged over the four continuations defines each action\textquotesingle s downstream value.

The learned critic is a frozen Qwen3-4B sequence-classification backbone with only a binary value head trained. Its input contains only the problem, prefix, and candidate action. Five folds are grouped by problem, with separate held-out and validation folds. Training uses five epochs, batch size 2, gradient accumulation 8, learning rate $10^{-3}$, and maximum length 4,096, selecting the epoch by validation AUROC. Routing uses the target action only when its predicted advantage is positive.

For sparse replay, Oracle selects positive measured-advantage events, FKL selects the highest-FKL events, and three Random selectors match Oracle counts within source cell and fold. Fold 0 holds out 30 branch states from 12 questions, leaving 33 one-token training events per arm. Each fresh Base LoRA adapter uses rank 64, alpha 128, 24 updates, effective batch size 8, learning rate $5\times10^{-6}$ with cosine decay, and loss only on the selected token. Question-start evaluation draws four reference-free samples per held-out question at temperature 1.0 and top-p 1.0, with a 16K response cap. Table~\hyperref[app:tab-h2b]{H2b} reports this single-seed pilot. Table~\hyperref[app:tab-h2a]{H2a} gives the pooled action-selection results.

\begin{table*}[!t]
  \centering
  \scriptsize
  \setlength{\tabcolsep}{6pt}
  \renewcommand{\arraystretch}{1.0}
  \begin{tabular}{@{}ccc@{}}
  \toprule
  \textbf{Selector} & \textbf{Reference-free gain (pp)} & \textbf{Role} \\
  \midrule
  Hindsight oracle & +11.30 {[}+8.15, +14.63{]} & Non-deployable upper bound \\
  Always privileged target & +2.04 & No routing \\
  Matched random router & +1.12 & Matches 54.8\% target coverage \\
  Held-out learned critic & +0.56 {[}-2.96, +4.07{]} & Five-fold CV grouped by problem \\
  High-FKL matched router & -1.30 & Same target-action count as critic \\
  Low-FKL matched router & +3.15 & Same target-action count as critic \\
  \bottomrule
  \end{tabular}
  \vspace{1mm}
  \caption*{\phantomsection\label{app:tab-h2a}\textbf{Table H2a:} Pooled action-selection results over 135 paired events from 57 questions. Brackets give problem-bootstrap 95\% CIs, available for the oracle and critic rows.}
\end{table*}

\begin{table*}[!t]
  \centering
  \scriptsize
  \setlength{\tabcolsep}{2pt}
  \renewcommand{\arraystretch}{1.0}
  \begin{tabular*}{\textwidth}{@{\extracolsep{\fill}}cccccc@{}}
  \toprule
  \textbf{Arm} & \textbf{Accuracy {[}95\% CI{]}} & \textbf{$\Delta$ Base {[}95\% CI{]}} & \textbf{Pass-any (\%)} & \textbf{Trunc. (\%)} & \textbf{Mean tokens} \\
  \midrule
  Frozen Base & 6.25 {[}0.00, 12.50{]} & +0.00 {[}0.00, 0.00{]} & 25.00 & 39.58 & 12,928.5 \\
  Oracle replay & 8.33 {[}0.00, 18.75{]} & +2.08 {[}-6.25, +10.42{]} & 25.00 & 41.67 & 12,645.2 \\
  FKL replay & 2.08 {[}0.00, 6.25{]} & -4.17 {[}-10.42, +0.00{]} & 8.33 & 39.58 & 12,607.2 \\
  Random 0 & 8.33 {[}0.00, 18.75{]} & +2.08 {[}-4.17, +8.33{]} & 25.00 & 45.83 & 12,713.9 \\
  Random 1 & 2.08 {[}0.00, 6.25{]} & -4.17 {[}-10.42, +0.00{]} & 8.33 & 39.58 & 12,916.5 \\
  Random 2 & 4.17 {[}0.00, 10.42{]} & -2.08 {[}-8.33, +4.17{]} & 16.67 & 45.83 & 13,010.1 \\
  \bottomrule
  \end{tabular*}
  \vspace{1mm}
  \caption*{\phantomsection\label{app:tab-h2b}\textbf{Table H2b:} Question-start results for the single-seed sparse-replay pilot. Evaluation covers the 12 held-out fold-0 questions with four samples each. Intervals resample held-out records and do not represent training-seed replication.}
\end{table*}

The oracle gap shows that useful alternatives exist in the candidate set, but the held-out learned critic does not identify them: pooled AUROC is 0.452, nonzero-effect sign accuracy is 46.3\%, and critic routing captures only 4.9\% of oracle headroom. Sparse replay is similarly inconclusive: Oracle replay equals the best matched-random arm on question-start accuracy, while FKL replay is below Base.

\subsection{Interpolation}
\label{app:diagnostic-interpolation}

For each matched Qwen3-4B seed and checkpoint $s \in \{20, 40, 60, 80, 100\}$, let $\Delta_{\text{problem},s}$ and $\Delta_{\text{privileged},s}$ denote the effective LoRA weight updates from the problem-only and privileged-solution runs. We evaluate $\Delta_s(\lambda) = (1-\lambda)\Delta_{\text{problem},s} + \lambda\Delta_{\text{privileged},s}$ at $\lambda \in \{0, 0.5, 1, 1.5\}$.

We realize each interpolation exactly with a rank-$2r$ adapter. Given source factors $A_{\text{problem}}, B_{\text{problem}}$ and $A_{\text{privileged}}, B_{\text{privileged}}$, we concatenate $A = [A_{\text{problem}}; A_{\text{privileged}}]$ and $B = [(1-\lambda)B_{\text{problem}}, \lambda B_{\text{privileged}}]$, then double both rank and alpha to preserve alpha/$r$. Figure~\ref{fig:discussion-diagnostics}a--b reuses the matched Figure~\ref{fig:interface-factorial} evaluations at $\lambda=0$ and $\lambda=1$ and adds newly evaluated $\lambda=0.5$ and $\lambda=1.5$ adapters under the same decoding protocol. At Step 100, the reconstructed endpoint adapters match the original effective updates across all 252 target modules, and a fresh evaluation preserves their ordering.

\begin{figure*}[!t]
  \centering
  \includegraphics[width=0.6\textwidth,trim=0 892 0 0,clip]{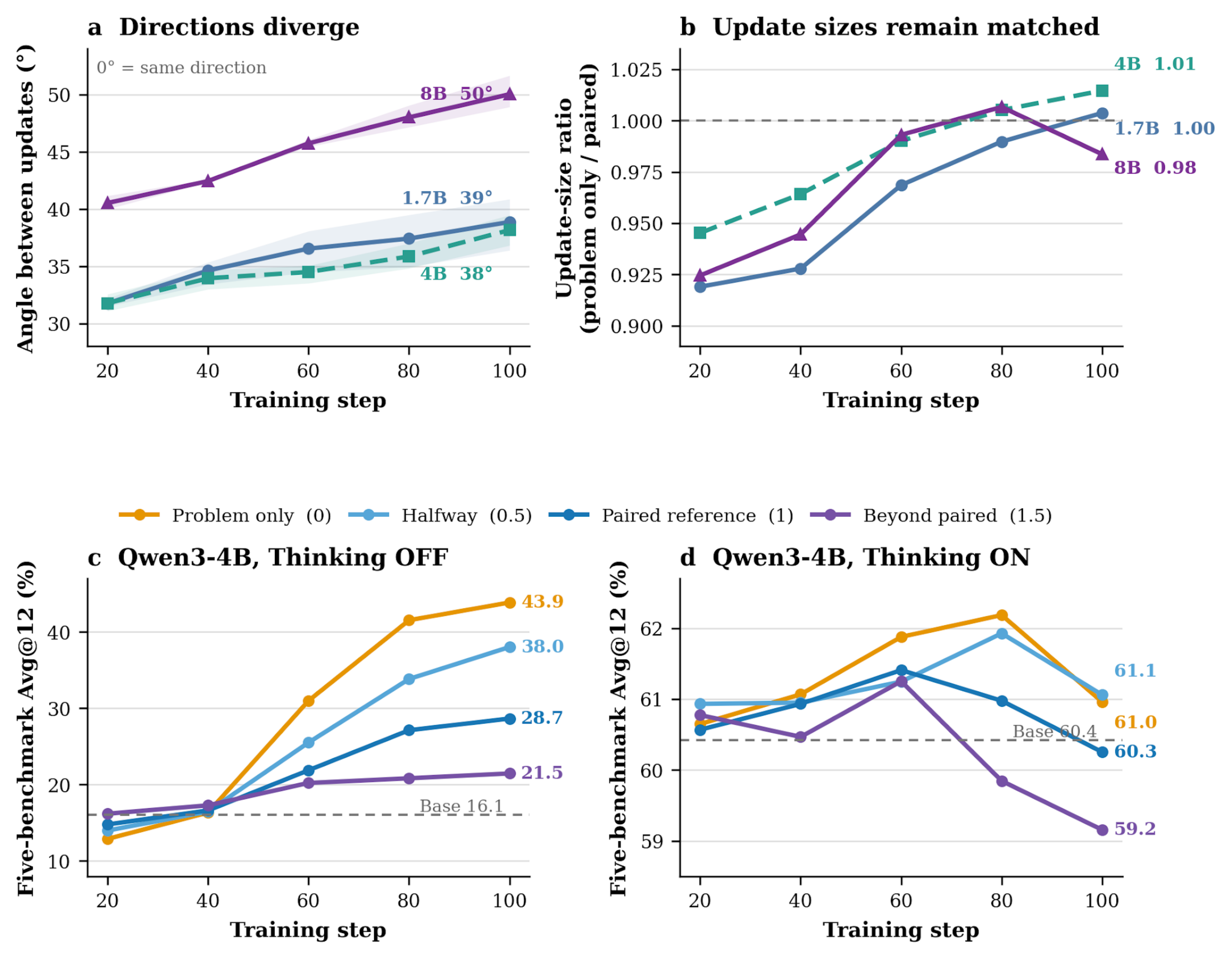}
  \caption*{\phantomsection\label{app:fig-h1}\textbf{Figure H1:} Privileged-solution versus problem-only LoRA updates. (a) Angle between the matched privileged-solution and problem-only effective updates. (b) Ratio of their update norms. Curves average three matched seeds. Endpoint labels give means, and shading in (a) shows the seed range.}
\end{figure*}

The privileged solution changes update direction rather than size: at Step 100, privileged-solution and problem-only updates differ by 39°, 38°, and 50° at 1.7B, 4B, and 8B, while their norm ratios remain between 0.98 and 1.01 (Figure~\hyperref[app:fig-h1]{H1}). Both Thinking-OFF and Thinking-ON evaluations use the five-benchmark Avg@12 suite, temperature 1.0, top-p 1.0, top-k disabled, and a 38,912-token response cap.

\begin{table*}[!t]
  \centering
  \scriptsize
  \setlength{\tabcolsep}{6pt}
  \renewcommand{\arraystretch}{1.0}
  \begin{tabular}{@{}ccccc@{}}
  \toprule
  $\boldsymbol{\lambda}$ & \textbf{Eval OFF: seed 0 / 1 / 2} & \textbf{OFF mean {[}95\% CI{]}} & \textbf{Eval ON: seed 0 / 1 / 2} & \textbf{ON mean {[}95\% CI{]}} \\
  \midrule
  0 & 38.94 / 41.79 / 52.37 & 44.37 {[}38.94, 52.37{]} & 60.89 / 61.09 / 61.06 & 61.01 {[}60.89, 61.09{]} \\
  0.5 & 33.77 / 37.30 / 43.04 & 38.04 {[}33.77, 43.04{]} & 60.12 / 60.95 / 62.11 & 61.06 {[}60.12, 62.11{]} \\
  1 & 27.66 / 30.49 / 30.17 & 29.44 {[}27.66, 30.49{]} & 59.75 / 59.98 / 60.88 & 60.21 {[}59.75, 60.88{]} \\
  1.5 & 21.44 / 22.25 / 20.73 & 21.47 {[}20.73, 22.25{]} & 58.93 / 59.25 / 59.29 & 59.15 {[}58.93, 59.29{]} \\
  \bottomrule
  \end{tabular}
  \vspace{1mm}
  \caption*{\phantomsection\label{app:tab-h3a}\textbf{Table H3a:} Fresh five-benchmark Avg@12 evaluation of interpolated adapters. Values are three-seed means with descriptive 95\% seed-bootstrap intervals. Seed-level values are shown for transparency.}
\end{table*}

\begin{table*}[!t]
  \centering
  \scriptsize
  \setlength{\tabcolsep}{6pt}
  \renewcommand{\arraystretch}{1.0}
  \begin{tabular}{@{}ccc@{}}
  \toprule
  \textbf{Evaluation mode} & \textbf{Contrast} & \textbf{$\Delta$ Avg@12 {[}95\% CI{]}} \\
  \midrule
  Thinking OFF & $\lambda=0.5$ - $\lambda=0$ & -6.33 {[}-9.34, -4.49{]} \\
  Thinking OFF & $\lambda=1$ - $\lambda=0$ & -14.93 {[}-22.20, -11.28{]} \\
  Thinking OFF & $\lambda=1.5$ - $\lambda=0$ & -22.90 {[}-31.64, -17.50{]} \\
  Thinking OFF & $\lambda=1.5$ - $\lambda=1$ & -7.97 {[}-9.44, -6.22{]} \\
  Thinking ON & $\lambda=0.5$ - $\lambda=0$ & +0.05 {[}-0.77, +1.05{]} \\
  Thinking ON & $\lambda=1$ - $\lambda=0$ & -0.80 {[}-1.14, -0.18{]} \\
  Thinking ON & $\lambda=1.5$ - $\lambda=0$ & -1.86 {[}-1.96, -1.77{]} \\
  Thinking ON & $\lambda=1.5$ - $\lambda=1$ & -1.05 {[}-1.59, -0.74{]} \\
  \bottomrule
  \end{tabular}
  \vspace{1mm}
  \caption*{\phantomsection\label{app:tab-h3b}\textbf{Table H3b:} Paired training-seed contrasts from the fresh Step-100 evaluation in Table~\hyperref[app:tab-h3a]{H3a}. Values are differences in pp with 95\% paired seed-bootstrap intervals.}
\end{table*}

\begin{table*}[!t]
  \centering
  \scriptsize
  \setlength{\tabcolsep}{6pt}
  \renewcommand{\arraystretch}{1.0}
  \begin{tabular}{@{}ccccccc@{}}
  \toprule
  \textbf{Eval mode} & $\boldsymbol{\lambda}$ & \textbf{Step 20} & \textbf{Step 40} & \textbf{Step 60} & \textbf{Step 80} & \textbf{Step 100} \\
  \midrule
  Thinking OFF & 0 & 12.89 & 16.35 & 31.00 & 41.55 & 43.87 \\
  Thinking OFF & 0.5 & 13.99 & 16.48 & 25.53 & 33.85 & 38.04 \\
  Thinking OFF & 1 & 14.80 & 16.62 & 21.89 & 27.13 & 28.66 \\
  Thinking OFF & 1.5 & 16.20 & 17.30 & 20.23 & 20.84 & 21.47 \\
  Thinking ON & 0 & 60.65 & 61.07 & 61.88 & 62.19 & 60.96 \\
  Thinking ON & 0.5 & 60.93 & 60.95 & 61.25 & 61.93 & 61.06 \\
  Thinking ON & 1 & 60.57 & 60.93 & 61.41 & 60.98 & 60.25 \\
  Thinking ON & 1.5 & 60.77 & 60.47 & 61.25 & 59.85 & 59.15 \\
  \bottomrule
  \end{tabular}
  \vspace{1mm}
  \caption*{\phantomsection\label{app:tab-h3c}\textbf{Table H3c:} Complete three-seed mean Avg@12 values plotted in Figure~\ref{fig:discussion-diagnostics}a--b. The $\lambda=0$ and $\lambda=1$ rows reuse the matched Figure~\ref{fig:interface-factorial} checkpoint evaluations. $\lambda=0.5$ and $\lambda=1.5$ are newly evaluated interpolated adapters. Table~\hyperref[app:tab-h3a]{H3a} reports the independent fresh Step-100 endpoint check.}
\end{table*}

At Step 100, Thinking-OFF performance decreases along the privileged-minus-problem-only direction in every seed, while Thinking-ON performance remains flat through $\lambda=0.5$ and then declines. The endpoint interpolation therefore erodes the cross-mode gain without revealing a compensating benefit in the thinking mode used by the privileged teacher. Because Table~\hyperref[app:tab-h3a]{H3a} re-evaluates the Step-100 endpoints with fresh samples, its $\lambda=0$ and $\lambda=1$ values differ slightly from the reused Figure~\ref{fig:interface-factorial} evaluations in Table~\hyperref[app:tab-h3c]{H3c} (44.37 versus 43.87 and 29.44 versus 28.66 under Thinking OFF; 61.01 versus 60.96 and 60.21 versus 60.25 under Thinking ON). These differences reflect sampling variation at temperature 1.0, and every reused value lies within the corresponding fresh seed range.

\subsection{Scope}

These experiments are falsification tests, not a joint method leaderboard. Panels (b), (c), (d), and (f) use one training seed. Panels (a) and (e) use three. Panels (g-h) average three matched seeds. Datasets, model scales, and training protocols differ across panels, while evaluation decoding is fixed within each matched comparison. The branch audit and replay pilot further separate the existence of useful local actions from the ability to select and retain them. Together, these tests narrow several simple implementation explanations without showing that every alternative OPSD formulation must fail.

\end{document}